\documentclass[runningheads]{llncs}

\usepackage{tikz}
\usepackage{comment}
\usepackage{amsmath,amssymb} % define this before the line numbering.
\usepackage{graphicx}
\usepackage{multirow}
\usepackage{booktabs} % for professional tables
\usepackage{amsfonts}       % blackboard math symbols
\usepackage{nicefrac}       % compact symbols for 1/2, etc.
\usepackage{bm}
\usepackage{algorithm}
\usepackage{algorithmic}
\usepackage{xcolor}
\usepackage{wrapfig}

\usepackage[pagebackref,breaklinks,colorlinks]{hyperref}
\usepackage{orcidlink}
\usepackage[accsupp]{axessibility}  % Improves PDF readability for those with disabilities.

\usepackage[capitalize]{cleveref}
\crefname{section}{Sec.}{Secs.}
\Crefname{section}{Section}{Sections}
\Crefname{table}{Table}{Tables}
\crefname{table}{Tab.}{Tabs.}

\newcommand{\supp}[1]{\textrm{Supp}({#1})}

\newcommand*{\APPENDIX}{}%

\begin{document}
% \linenumbers
\pagestyle{headings}
\mainmatter
\def\ECCVSubNumber{6532}  % Insert your submission number here

\title{Learning Energy-Based Models With Adversarial Training} % Replace with your title

\begin{comment}
\titlerunning{ECCV-22 submission ID 6532} 
\authorrunning{ECCV-22 submission ID 6532} 
\author{Anonymous ECCV submission}
\institute{Paper ID 6532}
\end{comment}
%******************

%\begin{comment}
\titlerunning{AT-EBMs}

%\author{Xuwang Yin\orcidID{0000-1111-2222-3333} \and
%Shiying Li\orcidID{1111-2222-3333-4444} \and
%Gustavo K. Rohde\orcidID{2222--3333-4444-5555}}

\author{Xuwang Yin\orcidlink{0000-0001-7572-9144} \and
	Shiying Li\orcidlink{0000-0001-6988-8229} \and
	Gustavo K. Rohde\orcidlink{0000-0003-1703-9035}}

\authorrunning{X. Yin et al.}

\institute{University of Virginia\\
\email{\{xy4cm, sl8jx, gustavo\}@virginia.edu}}
%\end{comment}
%******************
\maketitle

\begin{abstract}
   We study a new approach to learning energy-based models (EBMs) based on adversarial training (AT). We show that (binary) AT learns a special kind of energy function that models the support of the data distribution, and the learning process is closely related to MCMC-based maximum likelihood learning of EBMs. We further propose improved techniques for generative modeling with AT, and demonstrate that this new approach is capable of generating diverse and realistic images. Aside from having competitive image generation performance to explicit EBMs, the studied approach is stable to train, is well-suited for image translation tasks, and exhibits strong out-of-distribution adversarial robustness. Our results demonstrate the viability of the AT approach to generative modeling, suggesting that AT is a competitive alternative approach to learning EBMs.
\end{abstract}

\section{Introduction}
\label{sec:intro}
In unsupervised learning, energy-based models (EBMs)~\cite{lecun2006tutorial} are a class of generative model that uses an energy function to model the probability distribution of the observed data. Unlike explicit density models, EBMs model the unnormalized density function, which makes it difficult to evaluate the likelihood function. Maximum likelihood learning of EBMs hence makes use of the likelihood function's gradient which can be approximated using Monte Carlo methods. Each iteration of the learning process involves first generating synthesized data by sampling from the current model, and then updating the model to maximize the energy difference between synthesized data and observed data. This process leads to an energy function that outputs low energies on the data manifold and high energies on other regions. EBMs find applications in image restoration (denoising, inpainting, etc), out-of-distribution detection, and various sample generation tasks. The main difficulties of training EBMs lie in the computational challenges from the sampling procedure and some training stability issues~\cite{du2019implicit,nijkamp2019learning,nijkamp2020anatomy,Grathwohl2020Your,grathwohl2021no,Du2021ImprovedCD,zhao2021learning,xiao2021vaebm}.
% Energy based models is a compelling class of generative models. EBMs are typically trained with the maximum likelihood principle. The likelihood function typically cannot be directly evaluated, but its gradient can be approximated by Monte Carlo methods. 
% EBMs find applications in various inverse problems, including. EBM has many compelling properties. One is less limitation on the energy function architecture. Follow the synthesis by analysis, which facilitate solving certain computer vision tasks easier to.
% Within each learning iteration, we generate synthesized examples by sampling from the current model, and then
% update the model parameters based on the difference between the synthesized examples and the
% observed examples, so that eventually the synthesized examples match the observed examples in
% terms of some statistical properties defined by the model. To sample from the current EBM, we need
% to use Markov chain Monte Carlo (MCMC), such as the Gibbs sampler [12], Langevin dynamics,
% or Hamiltonian Monte Carlo [33]. Recent work that parametrizes the energy function by modern
% convolutional neural networks (ConvNets) [28, 26] suggests that the “analysis by synthesis” process
% can indeed generate highly realistic images [49, 11, 21, 10].

Another line of work on adversarial training (AT) show that adversarially robust classifiers learn high-level, interpretable features, and  can be utilized to solve various computer vision tasks including generation, inpainting, super-resolution, and image-to-image translation~\cite{tsipras2018robustness,ilyas2019adversarial,engstrom2019adversarial,santurkar2019image}.
% Although the idea of using a off-the-shelf robust classifier for image semantic manipulation is compelling, the synthesized samples as demonstrated by the authors are generally not as high quality of those obtained with other generative models.  
% Although AT provides a versatile tool for image semantic manipulation, its generative performance is not as competitive as other generative approaches.
% the synthesized samples as demonstrated by the authors are generally not as high quality of those obtained with other generative models.  
Compared to state-of-the-art generative models, this AT approach does not provide a competitive generation performance and is therefore of limited value in many of these tasks.
Nonetheless, the generative properties of the robust classifier suggest that the model has captured the distribution of the training data, although the underlying learning mechanism is not yet well understood.

At a high level, both EBMs training and AT are based on the idea of first using gradient-based optimization to generate samples that reach high activation under the current model, and then optimizing the model to minimize its activation on the generated samples. 
% These two methods also provide very similar utilities in computer vision tasks. 
In addition, both approaches synthesize new samples by performing gradient descent on the trained model.
These similarities suggest that there are some connections between these two approaches.
% Given that EBM and AT provide very similar utilities in computer vision tasks, one may wonder if there are any connections between these two methods. 
% At a high level, both methods use gradient-based optimization to generate samples that reach high activations under the current model, and then optimize the model to minimize its activations on the generated samples. However, due to the difference between AT's multiclass objective and EBM's binary discrimiantive objective (\cref{eq:dll}), the resemblance is only on the level of a general mechanism.
% To simply the discussion, we can consider a multiclass classifier as an ensemble of binary classifier with shared computation. 
% In AT, the binary classifier defined by a logit output is trained by maximizing the difference between the binary classifier's outputs on (perturbed) in-class data and adversarial samples computed from other classes. This is similar to EBM training where the model is trained by maximizing the energy difference between positive and negative samples. 
% Crucially, both the adversarial examples in AT and negative samples in EBM are compued by maximing the model's outputs using using a gradient-based procedure.
% This resemblance motivates us to investigate further about the connection between AT and EBM.

In this work we investigate the mechanism by which AT learns data distributions, and propose improved techniques for generative modeling with AT.
% Instead of working with AT's standard multiclass objective, we focus on the binary AT objective~\cite{Yin2020GAT} which is simpler to analyze and more relevant to the task of generative modeling.
We focus on binary AT~\cite{Yin2020GAT} which does not requires class labels and hence naturally fits the generative modeling task.
% In this work, to motivate a more formal and concrete discussion about the connection between AT and EMB, we consider a binary AT objective~\cite{Yin2020GAT} which is more relevant to the task of generative modeling.
We first analyze the binary AT objective and the corresponding training algorithm, and show that binary AT learns a special kind of energy function that models the support of the observed data.
We then draw a connection between AT and MCMC-base maximum likelihood learning of EBMs by showing that the binary AT objective can be interpreted as a gradient-scaled version of the likelihood objective in EBMs training, and the PGD attack can be viewed as an non-convergent sampler of the model distribution. 
This connection provides us with intuition of how AT learns data distributions from a maximum likelihood learning perspective, and suggests that binary AT can be viewed as an approximate maximum likelihood learning algorithm.
% This connection suggests that binary AT can be viewed as an approximate maximum likelihood learning algorithm.

We further propose improved techniques for generative modeling with AT based on the above analysis. Our empirical evaluation shows that this AT approach provides competitive generation performance to explicit EBMs, and at the same time is stable to train (just like regular adversarial training), is well-suited for image translation tasks, and exhibits strong out-of-distribution adversarial robustness.
% The AT approach also has better sampling efficiency than a recent work~\cite{xiao2021vaebm} that claims to have improved sampling speed (\cref{tab:generation-speed}).
% The AT approach makes use of real and diverse In addition, the AT model learns informative gradient for transforming arbitrary out-distribution samples into valid samples of the modeled data, which is a useful feature in image-to-image translation applications. Is robust to adversarial perturbation in OOD detection tasks.
The main limitation of the studied approach is that it cannot properly learn the underlying density function of the observed data.
% (although we did not find this to be an issue in many practical applications). 
However, this problem is not unique to the studied approach - most existing work on learning EBMs relies on short-run non-convergent sampler to improve the training efficiency, and the learned model typically does not have a valid steady-state that reflects the distribution of the observed data~\cite{nijkamp2019learning,nijkamp2020anatomy}.

In summary, the contributions of this paper are:
1) We show that binary AT learns a special kind of energy function that models the support of the data distribution, and  the learning process is closely related to MCMC-based maximum likelihood learning of EBMs.
% 2) We propose improved techniques for generative modeling with AT, and demonstrate that this AT approach is capable of generating samples rivaling the quality of state-of-the-art explicit EBMs. 
2) We propose improved techniques for generative modeling with AT, and demonstrate competitive image generation performance to state-of-the-art explicit EBMs. 
3) We show that the studied approach is stable to train, has competitive training and test time sampling efficiency, and can be applied to denoising, inpainting, image translation, and worst-case out-of-distribution detection.

\section{Related Work}
\label{sec:related_work}

%\linebreak
\subsubsection{Learning EBMs.}
Due to the intractability of the normalizing constant in the EBMs likelihood function, maximum likelihood learning of EBMs makes use of the gradient of the log-likelihood which can be approximated using MCMC sampling.
Recent work~\cite{xie2016theory,du2019implicit,nijkamp2019learning,nijkamp2020anatomy} scaling EBMs training to high-dimensional data performs sampling using SGLD~\cite{welling2011bayesian} and initialize the chain from a noise distribution.
% The sampling process typically incurs high computational cost due to the gradient estimation performed at each sampling step.
% \textcolor{red}{TODO: CD use gradient-free MCMC see~\cite{du2019implicit}, Persistent Contrastive Diver-
	% gence (Tieleman~\cite{tieleman2008training}, 2008) (PCD) training}
The sampling process involves estimating the model's gradient with respect to the current sample at each step and therefore has high computational cost.
% Contrastive Divergence (CD)~\cite{hinton2002training} mitigates this issue by initializing the Markov chain at each step with samples from the observed data.
To improve the sampling efficiency, many authors consider short-run non-convergent SGLD sampler in combination with a persistent sampling buffer~\cite{du2019implicit,nijkamp2020anatomy,Grathwohl2020Your,grathwohl2021no,xiao2021vaebm,Du2021ImprovedCD}.
% Recent work scaling EBMs training to high dimensional data relies on short-run non-convegent SGLD sampler initialized from random noise~\cite{nijkamp2019learning, nijkamp2020anatomy, du2019implicit}.
Although a short-run sampler is sufficient for learning a generation model, the resulting energy function typically does not have a valid steady-state~\cite{nijkamp2019learning,nijkamp2020anatomy}.
The mixing time of the sampling procedure also depends on how close the chain-initialization distribution is to the model distribution. A recent trend hence considers initializing the sampling chain from samples produced by a generator fitted on the target distribution~\cite{xie2018cooperative,kumar2019maximum,han2019divergence,pang2020learning,han2020joint,grathwohl2021no,xiao2021vaebm,xie2021learning,arbel2021generalized,nijkamp2022mcmc,xie2022a}.
% To further reduce the computational cost of MCMC sampling, many authors explore the idea of training auxiliary generator networks to facilitate the sampling process~\cite{xie2018cooperative, grathwohl2021no, xiao2021vaebm, xie2021learning}.
% \cite{Grathwohl2020Your, liu2020hybrid} combine standard discriminative training with EBM training to obtain a discriminative model that at the same time have generative capability.

Maximum likelihood learning of EBMs also has some training stability issues, and various techniques have been developed to address these issues. These techniques include 1) using 
%There are also works focusing on addressing the training stability issues of EBMs~\cite{zhao2021learning,Du2021ImprovedCD}. 
%VAEBM~\cite{xiao2021vaebm} 
weight normalization~\cite{salimans2016weight}, Swish activation~\cite{ramachandran2017swish}, gradient clipping, and weight decay (see~\cite{xiao2021vaebm}), 2) gradient norm clipping on model parameters and using a KL term in the training objective (see~\cite{Du2021ImprovedCD}), 3) adjusting learning rate and SGLD steps during training and adding Gaussian noise to input images (see \cite{Grathwohl2020Your}),  4) gradient clipping on SGLD and model parameters and spectral normalization (see~\cite{du2019implicit}), and 5) multiscale training and smooth activation functions (see~\cite{zhao2021learning}). 
Overall, there does not seem to have a consensus on how to stabilize EBMs training.
% Apart from the computational challenge, EMBs also has various training stability issues.
Due to the computational challenge of MCMC sampling and stability issues, the successful application of EBMs to modeling high-dimensional data such as $256\times 256$ images is only achieved in some very recent works~\cite{xiao2021vaebm,zhao2021learning}. 

Aside from MCMC-based maximum likelihood learning of EBMs, alternative approaches for learning EBMs exist.
Score matching~\cite{hyvarinen2005estimation} circumvents the difficulty of estimating the partition function by directly modeling the derivatives of the data distribution. Score matching has recently been successfully applied to modeling large natural images and achieves competitive performance to state-of-the-art generative models such as GANs~\cite{song2019generative,song2020improved,song2021scorebased,ho2020denoising}.
% output the derivatives of the density parameterize any explicit energy function
% Although score matching does not require sampling during training, it still relies on MCMC for sample generation.  The sampling process typically takes thousands of steps, leading to slow sample generation. Because score matching does not model the energy function explicitly, it is not straightforward to apply them in tasks which require explicit knowledge of the data distribution (\eg, out-of-distribution detection).
Noise contrastive estimation (NCE)~\cite{gutmann2010noise} learns data distributions by contrasting the observed data with data from a known noise distribution. Similar to our approach, NCE makes use of a logistic regression model. The main difference is that in NCE, the logit of the classifier is the difference in log probabilities of the model distribution and the noise distribution, whereas in our approach the logit directly defines the estimator (i.e., the energy function).
% NCE avoids the difficulty of estimating the partition function by including it as a learnable parameter of the model. Although both NCE and binary AT (formulate, cast) the learning problem as a binary classification problem, there is a fundamental difference between what these two model represent. In NCE, the model directly defines the probability density of the observed data, while in the context of logistic regression, binary AT, the model should be interpreted as the posterior probabilities of the classes given the data. NCE has not been succcesfull applied to modelling high dimentional data. Noise contrastive approaches, which learn energy functions through density ratio estimation,
Unlike other EBMs, NCE typically does not scale well to high-dimensional data~\cite{gutmann2010noise,ceylan2018conditional,rhodes2020telescoping}.
% ~\cite{grathwohl2021no}.
% The strategy used by score matching is to minimize the expected squared difference between the derivatives of the model’s log density with respect to the input and the derivatives of the data’s log density with respect to the input: Noise-contrastive estimation (NCE) ()

\subsubsection{Maximin interpretation of EBMs.}
When the noise term in the SGLD sampler is disabled, the learning process of EBMs can be interpreted as solving a \textit{maximin} game~\cite{xie2018learning,xie2020generative,xie2021cooperative}. This interpretation coincides with our formulation in \cref{eq:maxmin-problem}. The key differences lie in the value function, the setting of the sampler (SGLD vs. PGD attack), and the Markov chain initiation distribution.

\subsubsection{Understanding and Improving AT Generative Model.} Our work is related to~\cite{zhu2021towards,wang2022a} which are also attempts to understand and improve AT's generative capability. \cite{zhu2021towards,wang2022a}' focus is on the \textit{supervised setting}, where they use the concept of learning class-conditional energy functions to understand adversarially robust classifiers' generative capability. Our analysis is in an \textit{unsupervised setting}, where we attempt to understand \textit{binary} AT's generative property by connecting it with the standard EBMs formulation (\cref{eq:ebm}). Although~\cite{wang2022a} also considers the unsupervised scenario, their unsupervised generative model is a contrastive learning model~\cite{jiang2020robust} which requires training samples to perform test-time sampling. Our generative model is based on binary AT and follows the standard practice of MCMC sampling from the learned energy function. In addition to the theoretical analysis, we propose improve training techniques that allow us to obtain a significantly better FID on CIFAR-10 (\cref{tab:cifar10_scores}) and successfully scale the training to $ 256\times 256 $ datasets.

\section{Background}
\subsection{Energy-Based Models}

Energy-based models (EBMs)~\cite{lecun2006tutorial} represent probability distributions by converting the outputs of a scalar function $f_\theta$ into probabilities through a Gibbs distribution:
\begin{equation}
	\label{eq:ebm}
	p_\theta (x) = \frac{\exp(f_\theta (x))}{Z(\theta)},
\end{equation}
where the normalizing constant $Z(\theta)$, also known as the partition function, is an integral over the unnormalized probability of all states: $Z(\theta) = \int \exp(f_\theta(x)) dx$. The energy function is defined as $E_\theta(x) = -f_\theta(x)$, and thus has the property of attributing low energy outputs on the support of the target data distribution and high energy outputs in other regions.

% Fortunately, the graidnet of log of the partition function is tractable: $\nabla_\theta \log(Z) = -\mathbb{E}_{x\sim p_\theta(x)}\nabla_\theta E(x)$
% Hence give a dataset, , the gradient of can be computed as .
% \begin{equation}
	%     L(\theta) = \mathbb{E}_{x\sim p_\mathrm{data}} [\log p_\theta (x)]
	% \end{equation}

% \begin{align}
	%     -\nabla_\theta L & = \nabla_\theta \mathbb{E}_{p_\mathrm{data}} [-E_\theta (x) - \log (Z)] \\
	%                      & = \mathbb{E}_{x\sim p_\mathrm{data}} [\nabla_\theta E_\theta (x)] -\mathbb{E}_{x'\sim p_\theta(x)}\nabla_\theta E(x')
	% \end{align}
% \begin{equation}
	%     \nabla_\theta -L =  \mathbb{E}_{p_\mathrm{data}} [\nabla_\theta E_\theta (x)]-\mathbb{E}_{x\sim p_\theta(x)}\nabla_\theta E(x)
	% \end{equation}
For many interesting models, the partition function $Z(\theta)$ is intractable, and therefore maximum likelihood estimation (MLE) of the model parameters $\theta$ is not directly applicable. Standard maximum likelihood learning of EBMs makes use of the gradient of the log likelihood function. Denote the distribution of the observed data as $p_\mathrm{data}$, the gradient of the log likelihood takes the form
\begin{equation}
	\label{eq:dll}
		\nabla_\theta   \mathbb{E}_{\mathrm{x}\sim p_\mathrm{data}}  [\log p_\theta (x)] = 
		\mathbb{E}_{\mathrm{x}\sim p_\mathrm{data}} [\nabla_\theta f_\theta (x)] -\mathbb{E}_{\mathrm{x}\sim p_\theta(x)}[\nabla_\theta f_\theta(x)].
\end{equation}
Intuitively, maximizing log-likelihood with this gradient causes $f_\theta(x)$ to increase on $p_\mathrm{data}$ samples and  decrease on samples drawn from $p_\theta$; when $p_\theta$ matches $p_\mathrm{data}$, the gradient cancels out and the training terminates.

% By taking the The gradient of the log-likelihood with respect to the model parameters, we observe
Evaluating $\mathbb{E}_{\mathrm{x}\sim p_\theta(x)}\nabla_\theta f_\theta(x)$ requires sampling from the model distribution. This can be done with Markov chain Monte Carlo (MCMC) methods. Recent work scaling EBMs training to high-dimensional data~\cite{xie2016theory,du2019implicit,nijkamp2019learning,nijkamp2020anatomy} makes use of the SGLD method~\cite{welling2011bayesian} which samples the model distribution by 
%Stochastic Gradient Langevin Dynamics (SGLD)~\cite{welling2011bayesian} is a popular sampler in recent work scaling EBMs training to high-dimensional data~\cite{xie2016theory,du2019implicit,nijkamp2019learning,nijkamp2020anatomy}.
% Recent work~\cite{xie2016theory, du2019implicit, nijkamp2019learning, nijkamp2020anatomy} scaling EBMs training to high-dimensional data uses a sampler based on Stochastic Gradient Langevin Dynamics (SGLD)~\cite{welling2011bayesian}
% A widely adopted sampler is based on Stochastic Gradient Langevin Dynamics (SGLD)which performs sampling by first drawing a sample from a random noise distribution $p_0$, and then updating the sample through
%The sampling process of SGLD follows
\begin{equation}
	\label{eq:sgld}
	x_0\sim p_0, \quad x_{i+1} = x_i + \frac{\lambda}{2} \nabla_x f_\theta(x_i) + \epsilon, \quad \epsilon\sim \mathcal{N} (0,\lambda),
\end{equation}
where $p_0$ is some random noise distribution.
A proper SGLD sampler requires a large number of update steps in order for the distribution of sampled data to match $p_\theta$.
Due to the high computational cost of this sampling process, many authors resort to short-run non-convergent MCMC to improve the sampling efficiency\cite{nijkamp2019learning,nijkamp2020anatomy,du2019implicit,xie2016theory,Grathwohl2020Your}. The resulting model typically does not have a valid steady-state that reflects the distribution of the observed data, but  is still capable of generating realistic and diverse samples~\cite{nijkamp2019learning,nijkamp2020anatomy}.
% \textcolor{red}{TODO: Check SGLD definition \cite{gao2021learning}}
% \textcolor{red}{TODO: talk about steps sizes, convergence, more about sampling~\cite{Grathwohl2020Your} and non-convergent nature~\cite{nijkamp2019learning} }
% In practice the step-size, α, and the standard deviation
% of ϵ is often chosen separately leading to a biased sampler which allows for faster training. 
% leading to non-convegent MCMC, which is sufficient for synthesis tasks.
% \cite{zhao2021learning} Section 3.1, 
% \cite{pang2020learning}
% \cite{nijkamp2019learning}
% \cite{du2019implicit} Section 3
% \cite{du2020improved}
% \cite{song2019generative} Section 2.2
% \cite{gao2021learning}
% \cite{Grathwohl2020Your} H.1
% While we are no longer working with a valid MCMC sampler,
% \cite{nijkamp2020anatomy}
% \cite{xiao2021vaebm} Although they do not require
% iterative MCMC during training, they need very long sampling chains to anneal the noise when
% sampling from the model (& 1000 steps). Therefore, sample generation is extremely slow.

\subsection{Binary Adversarial Training}
\label{sec:gat}
Binary adversarial training \cite{Yin2020GAT} is a method for detecting adversarial examples.
% The proposed method uses an ensemble of adversarially trained binary classifiers to detect adversarial examples.
In a $K$ class classification problem, the detection method consists of $K$ binary classifiers, with the $k$-th binary classifier trained to distinguish clean data of class $k$ from adversarially perturbed data of other classes. A committee of $K$ binary classifiers then provides a complete solution for detecting adversarially perturbed samples of any classes.

Denote the data distribution of class $k$ as $p_\mathrm{data}$, the mixture distribution of other classes as $p_0 =\frac{1}{K-1}\sum_{i=1,...,K, i\neq k} p_i$, the $k$-th binary classifier is trained by maximizing the objective
\begin{equation}
	\label{eq:gat-obj}
		J({D}) = \mathbb{E}_{\mathrm{x} \sim p_\mathrm{data}}[\log {D}({x})]  +  \mathbb{E}_{\mathrm{x} \sim p_0}[\min_{x'\in\mathbb{B}(x,\epsilon)}\log (1 - {D}(x')))],
\end{equation}
where $D: \mathcal{X} \subseteq \mathbb{R}^d \rightarrow [0,1]$ is the classification function, and $\mathbb{B}(x,\epsilon)$ is a neighborhood of $x$: $\mathbb{B}(x,\epsilon)=\{x' \in \mathcal{X}: \| x'-x\|_2 \leq \epsilon \}$. 
In practice, $D$ is defined by applying a logistic sigmoid function to the output of a neural network: 
\begin{equation}
	\label{eq:D_define}
	D(x) = \sigma(f_\theta(x)),
\end{equation} 
where $f_\theta$ is a neural network with a single output node and parameters $\theta$.
%\cref{eq:gat-obj} is characterized by an \textit{inner minimization problem} and an \textit{outer maximization problem}; when the inner minimization is perfectly solved and the training achieves a vanishing loss,
%$D$ becomes a perfectly robust model capable of separating $p_\textrm{data}$ samples from any $\epsilon$-bounded adversarial examples perturbed from  $p_{0}$ data.  

The inner minimization in \cref{eq:gat-obj} is solved using the PGD attack~\cite{madry2017towards,kurakin2016adversarial}, a first-order method that employs an iterative update rule of ($l^2$-based attack):
\begin{equation}
	\label{eq:pgd_d}
	x_0\sim p_0, \quad x_{i+1} = \text{Proj}(x_i - \lambda \frac{\nabla_x \log (1-D(x_i))}{\|\nabla_x \log (1-D(x_i))\|_2}),
\end{equation}
where $\lambda$ is some step size, and \text{Proj} is the operation of projecting onto the feasible set $\mathbb{B}(x,\epsilon)$. 
%The \textit{normalized steepest descent} rule inside the \text{Proj} function, was introduced for dealing with vanishing gradient when optimizing the cross-entropy loss~\cite{advtutorials}.
Because the gradient vector in \cref{eq:pgd_d} is normalized to have unit norm, we can equivalently implement the attack by directly performing gradient ascent on $f_\theta$:
\begin{equation}
	\label{eq:pgd_f}
	x_0\sim p_0, \quad x_{i+1} = \text{Proj}(x_i +\lambda \frac{\nabla_x f_\theta(x_i)}{\|\nabla_x f_\theta(x_i)\|_2}).
\end{equation}
%Similar to standard AT, a model trained with binary AT has strong interpretability --- an unbounded attack that maximizes the model's output results in samples that resemble data from $p_\textrm{data}$, suggesting that the model has captured $p_\textrm{data}$. 
% Compared to the multiclass objective employed in standard adversarial training, the binary objective~\cref{eq:gat-obj} used by GAT is  much easier to analyze.

\section{Binary AT Generative Model}
In this section we develop a generative model based on binary AT. We first analyze the optimal solution to the binary AT problem, and then investigate the mechanism by which binary AT learns the data distribution, and finally interpret the learning process from the maximum likelihood learning perspective. Our main result is that under a proper configuration of perturbation limit and $p_0$ data, binary AT learns a special kind of energy function that models the support of $p_\textrm{data}$. Based on these theoretical insights,  we proposed improved training techniques.

%\label{sec:AT-properties}
\subsection{Optimal Solution to the Binary AT Problem}
We consider the optimal solution of \cref{eq:gat-obj} under the scenario of unbounded perturbation: $\mathbb{B}(x,\epsilon)=\mathcal{X}$. This allows us to further simplify the PGD attack by removing the \text{Proj} operator:
% We can understand the gradient-based search as a sampling process.
\begin{equation}
	\label{eq:pgd_noproj}
	x_0\sim p_0, \quad x_{i+1} = x_i +\lambda \frac{\nabla_x f_\theta(x_i)}{\|\nabla_x f_\theta(x_i)\|_2}.
\end{equation}
% Perturbing  $p_{0}$ data can then be thought of as moving mass of $p_{0}$ to locations in $\mathcal{X}$ via a
% transformation function $T:\mathcal{X}\rightarrow\mathcal{X}$ where $T(x)=x+\Delta_x$, with $\Delta_x$ being the perturbation computed on sample $x$.
Perturbing  $p_{0}$ samples can be thought of as moving $p_{0}$ samples via a
translation function $T(x)=x+\Delta_x$, with $\Delta_x$ being the perturbation computed on sample $x$.
We can write the density function of the perturbed distribution $p_T$ using  random
variable transformation:
\begin{equation}
	p_T(z) = \int_\mathcal{X} p_{0}(x)\delta (z-T(x))dx.
\end{equation}
% Let $\mathcal{M}_{+}^1(\mathcal{X})$ be the set of density functions in $\mathcal{X}$, 
The inner problem in \cref{eq:gat-obj} can then be interpreted as determining the distribution  which has the lowest expected value of $\log(1-D(x))$:
\begin{equation}
	\label{eq:inner_min}
	p_T^* = \arg\min_{p_T} \mathbb{E}_{\mathrm{x} \sim p_T} [ \log(1-{D}(x)) ].
\end{equation}
The objective of the outer problem is then the log-likelihood in a logistic regression model which discriminates $p_\mathrm{data}$ samples from $p_T^*$ samples:
\begin{equation}
	\label{eq:D_objective}
	J (D) = \mathbb{E}_{\mathrm{x} \sim p_\mathrm{data}} [\log D(x)] + \mathbb{E}_{\mathrm{x}\sim p_T^*} [\log (1-D(x))].
\end{equation}
We can equivalently formulate \cref{eq:gat-obj} as a \textit{maximin} problem
\begin{equation}
	\label{eq:maxmin-problem}
		\max_{{D}}\min_{p_T} U({D}, p_T) = \mathbb{E}_{\mathrm{x} \sim p_\mathrm{data}}[\log {D}({x})] +  \mathbb{E}_{\mathrm{x} \sim p_T} [ \log(1-{D}(x)) ],
\end{equation}
and obtain its optimal solution by following the standard approach to solving maximin problems:

\begin{proposition}
	\label{pro:optimalD}
	% 	In the maximin game $\max_{D}\min_{p_T} U(D, p_T)$, the best strategy for player $D$ is to choose a $D$ that outputs $\frac{1}{2}$ in $\supp {p_k}$ and $\leq \frac{1}{2}$ in $\mathcal{X}\setminus \supp{p_k}$, the best strategy for player $p_T$ is to choose a $p_T$ with its mass distributed in locations where $D$ outputs $\frac{1}{2}$, and the maximum payoff is $-\log(4)$.
	The optimal solution of $\max_{D}\min_{p_T} U(D, p_T)$ is $U(D^*, p_T^*)=-\log(4)$, where $D^*$ outputs $\frac{1}{2}$ on $\supp {p_\textrm{data}}$ and $\leq \frac{1}{2}$ outside $\supp{p_\textrm{data}}$, and $p_T^*$ is supported in the contour set  $\{D=\frac{1}{2}\}$.
\end{proposition}
\begin{proof}
	See the supplementary materials.
\end{proof}
The above maximin problem can also be interpreted as a two-player zero-sum game, and is closely related to GANs~\cite{goodfellow2014generative}'s \textit{minimax} game which has the form
\begin{equation}
	\label{eq:gan-minmax}
	\min_{G} \max_D V(D, G) = \mathbb{E}_{\mathrm{x} \sim p_{\text{data}}}[\log D({x})] + \mathbb{E}_{\mathrm{z} \sim p_{{z}}}[\log (1 - D(G({z})))].
\end{equation}
The game-theory point of view provides a convenient way to understand their differences. We include a game theory-based analysis of $\max_{D}\min_{p_T} U(D, p_T)$ and a comparative analysis of GANs in the supplementary materials.

\subsection{Learning Mechanism}
\label{sec:mechanism}
\cref{pro:optimalD} states that by solving $\max_{D}\min_{p_T} U(D, p_T)$
we can obtain a $D$ that outputs $\frac{1}{2}$ on the support of $p_\mathrm{data}$ and $\le \frac{1}{2}$ on other regions. 
This result is obtained by assuming that for any $D$, the inner minimization \cref{eq:inner_min} is always perfectly solved. 
In practice, when $D$ is randomly initialized, it has many local maxima outside the support of $p_\textrm{data}$.  Because the inner minimization is solved by taking $p_0$ samples and then performing gradient ascent on $D$ with \cref{eq:pgd_noproj}, this process  can  get trapped in different local maxima of $D$. Hence we can think of this process as searching for these local maxima and then put the perturbed $p_0$ data in these regions. 
%low probability under $p_\textrm{data}$. 
Then in the model update stage (outer maximization), $D$ is updated by increasing its outputs on $p_\textrm{data}$ samples and decreasing its outputs on the perturbed $p_0$ data.  
%This process causes local maxima values to decrease.
% Hence by alternating between these two optimization procedures, local maxima of $D$ will be constantly suppressed.
By repeating this process, local maxima get suppressed and the model learns to correctly model  $\supp {p_\textrm{data}}$.

%In practice, the inner problem is solved by performing PGD attacks on $p_0$ samples; when $D$ is a non-concave function, this process can  get trapped in different local maxima of $D$. Given that an optimal $p_T^*$ is not always attainable, the actual $D$ solution obtained from an algorithm can be different from the one predicted by \cref{pro:optimalD}.

\begin{figure}[h!]
	\centering
	\scriptsize
	\begin{minipage}{.25\textwidth}
		\centering
		\includegraphics[width=\linewidth]{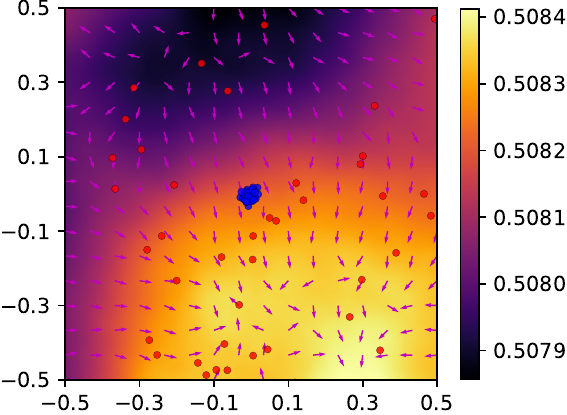}
		\\ (1a)
		%		\caption*{wo}
	\end{minipage}%
	\begin{minipage}{0.25\textwidth}
		\centering
		\includegraphics[width=\linewidth]{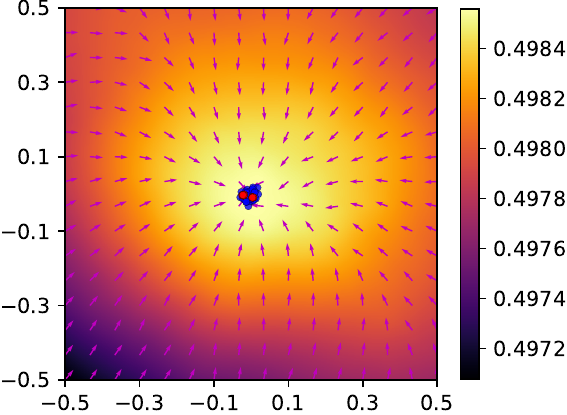}
		\\ (1b)
	\end{minipage}
	\vrule
	\hfill
	\begin{minipage}{.24\textwidth}
		\centering
		\includegraphics[width=\linewidth]{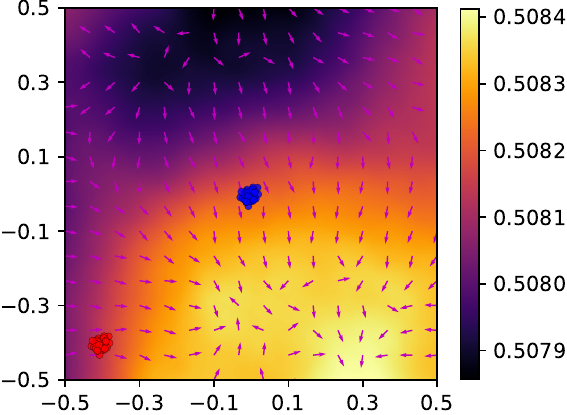}
		\\ (2a)
	\end{minipage}%
	\begin{minipage}{0.24\textwidth}
		\centering
		\includegraphics[width=\linewidth]{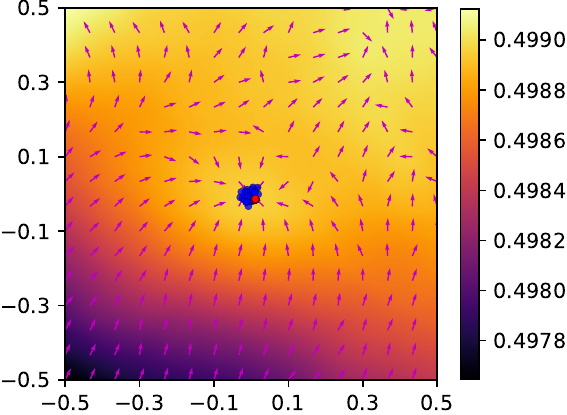}
		\\ (2b)
	\end{minipage}
	\caption{Plots of contours and (normalized) gradient vector fields of the $D$ functions learned with different $p_0$ data. Left and right panel respectively show the initial state (1a and 2a) and final state (1b and 2b) of $D$ when $p_0$ data is respectively uniformly distributed (red points in 1a) and concentrated in the lower left corner (red points in 2a). $p_\textrm{data}$ is a Gaussian distribution centered at $(0,0)$ (blue points).  
	}
	\label{fig:2D}
\end{figure}

The algorithm for solving the maximin problem is described in \cref{alg:maximin-solver}. \cref{fig:2D}  left panel  shows the 2D simulation result of the algorithm when the $p_0$ dataset contains random samples from the uniform distribution. 
It can be seen that when the algorithm converges, local maxima outside $\supp {p_\mathrm{data}}$ are suppressed, and $D$ (approximately) outputs $\frac{1}{2}$ on  $\supp {p_\mathrm{data}}$ as predicted by \cref{pro:optimalD}. Meanwhile, $D$  retains the gradient information for translating out-distribution samples to $\supp {p_\mathrm{data}}$. 

Because the PGD attack is deterministic gradient ascent, its ability to discover different local maxima depends on the diversity of $p_0$ samples.
%Whether can learn well depends on how well the inner minimization can find different local maxima.
\cref{fig:2D} right panel shows that when $p_0$ data is concentrated in the bottom left corner, the final $D$ still has local maxima outside the support of $p_\mathrm{data}$. These local maxima are not suppressed because they were never discovered by the perturbed $p_0$ data.
% Inspection of the gradient vector field reveals that these spurious local maxima cannot be visited by the gradient-based search initialized at $p_0$ data (hence cannot be suppressed). 
%This result suggests that in order to learn a valid energy function that is well defined in the entire data space, the support of $p_0$ should span as much space as possible.

The above analysis reveals how binary AT learns data distributions: \textit{the learning starts with a randomly-initialized $D$ solution, and then iteratively refine the solution by suppressing local maxima outside the support of the target distribution}. This process is similar to EBMs training where the model distribution's spurious modes are constantly discovered by MCMC sampling and subsequently suppressed in the model update stage. 
However, unlike the EBMs likelihood objective \cref{eq:dll}, the AT objective \cref{eq:maxmin-problem} cannot properly learn the density function, but can only capture its support. This is corroborated by the 2D experiment where $D$ outputs $\frac{1}{2}$ uniformly on the support of $p_\textrm{data}$ (blue points).

\begin{algorithm}[h]
	\caption{\small Binary Adversarial Training}
	\label{alg:maximin-solver}
	%	\scalebox{0.95}{%
		{\small
			\begin{algorithmic}[1]
				\REPEAT
				%			\STATE Sample minibatch of $m$ samples $\{ {x}_1^k, \dots, {x}_m^k \}$  from $p_k$, and $m$ samples $\{ {x}_1^{-k}, \dots, {x}_m^{-k} \}$  from $p_{{-k}}$.
				\STATE Draw samples $\{ {x}_i \}_{i=1}^m$  from $p_\mathrm{data}$, and samples $\{ {x}_i^{0}\}_{i=1}^m$  from $p_0$.
				% 			\STATE Compute perturbed samples $\{ {x}_i^*  \}_{i=1}^m$ by solving $\min_{x'\in\mathbb{B} (x,\epsilon)} \log(1- {D}(x'))$ for each $ {x}\in \{ {x}_i^{-k}\}_{i=1}^m $.
				\STATE Update $\{ {x}_i^{0}\}_{i=1}^m$ by performing $K$ steps PGD attack \cref{eq:pgd_noproj} on each sample. Denote the resulting samples as $\{ {x}_i^*\}_{i=1}^m$.
				\STATE Update $D$ by maximizing $\frac{1}{m} \sum_{i=1}^m \log {D}({x_i}) + \frac{1}{m} \sum_{i=1}^m \log (1 - D({x_i^*}))$ (single step).
				%			\STATE Compute the perturbed distribution $\supp{p_t}:=\{\arg\min_{x'\in\mathbb{B}(x,\epsilon)} \log(1-D(x')), {x} \in \supp{p_{-k}}\}$ \}
				%			\STATE Update $D$ by maximizing $\mathbb{E}_{\mathrm{x} \sim p_k}[\log D({x})] +  \mathbb{E}_{\mathrm{x} \sim p_t}[ \log(1-D(x)) ]$ (single step).
				\UNTIL {$D$ convergences}
			\end{algorithmic}
		}
		%	}%
\end{algorithm}

%
%In the binary AT  \cref{alg:maximin-solver}, after drawing sampling from $p_\mathrm{data}$ and $p_0$, step 3  solves the inner problem~\cref{eq:inner_min} by performing PGD attacks on $p_0$ samples. When $D$ is a non-concave function, the resulting samples $\{ {x}_i^*  \}_{i=1}^m$ are in local maxima of $D$. In step 4, $D$ is updated by maximizing its outputs on $\{ {x}_i \}_{i=1}^m$ and minimizing its outputs on $\{ {x}_i^*\}_{i=1}^m$. A key observation is that when $\{ {x}_i^*  \}_{i=1}^m$ are in local maxima, the update in step 4 causes these local maxima values to decrease. Hence by alternating between these two optimization procedures, local maxima of $D$ will be constantly suppressed.

\subsection{Maximum Likelihood Learning Interpretation}
\label{sec:connection}

We next consider the learning process of binary AT from a maximum likelihood learning point of view.
% The learning process of binary AT is closely related to MCMC-based EBMs training.
Both binary AT and MCMC-based EBMs learning 
% Both approaches 
employ an iterative optimization algorithm, where in each iteration the contrastive data is computed by performing gradient ascent on the current model, and then the model is updated  by maximizing its outputs on the observed data and minimizing its outputs on the contrastive data.
The following analysis shows that the PGD attack can be viewed as a non-convergent sampler of the model distribution, and the binary AT objective~\cref{eq:D_objective} can be interpreted as a gradient-scaled version of the EBMs objective~\cref{eq:dll}. 
\cref{tab:compare} summaries their key differences. 
\setlength{\tabcolsep}{4pt}
\begin{table}[h!]
	\begin{center}
		\caption{Key differences between binary AT and maximum likelihood EBMs.
		}
		\label{tab:compare}
		\resizebox{0.85\textwidth}{!}{%
			\begin{tabular}{@{}ll@{}}
				\toprule
				\multirow{2}{*}{\textbf{Objective gradient}} & EBMs: $\mathbb{E}_{\mathrm{x}\sim p_\mathrm{data}} [\nabla_\theta f_\theta (x)] -\mathbb{E}_{\mathrm{x}\sim p_\theta(x)}[\nabla_\theta f_\theta(x)]$                                                                                    \\ \cmidrule(l){2-2} 
				& Binary AT: $\mathbb{E}_{\mathrm{x} \sim p_\mathrm{data}}[(1-\sigma(f_\theta(x)))\nabla_\theta f_\theta(x)] - \mathbb{E}_{\mathrm{x}\sim p_T^*}[\sigma(f_\theta(x))\nabla_\theta f_\theta(x)]$                    \\ \midrule
				\multirow{2}{*}{\textbf{Contrastive data}}   & EBMs: $x_0\sim p_0,x_{i+1} = x_i + \frac{\lambda}{2} \nabla_x f_\theta(x_i) + \epsilon, \epsilon\sim \mathcal{N}(0,\lambda)$                                                                                                            \\ \cmidrule(l){2-2} 
				& Binary AT: $x_0\sim p_0,x_{i+1} = x_i +\lambda \frac{\nabla_x f_\theta(x_i)}{\|\nabla_x f_\theta(x_i)\|_2}$                                                                                                                             \\ \midrule
				\multirow{2}{*}{\textbf{$p_0$ data}}         & EBMs: A noise distribution or a distribution close to $p_\textrm{data}$                                                                                                                                                                 \\ \cmidrule(l){2-2} 
				& \begin{tabular}[c]{@{}l@{}}Binary AT: A real and diverse out-distribution dataset \\ (80 million tiny images for CIFAR-10 and ImageNet for 256x256 datasets).\end{tabular} \\ \bottomrule
			\end{tabular}%
		} 
	\end{center}
\end{table}
\setlength{\tabcolsep}{1.4pt}

\noindent\textbf{Contrastive Data Computation.}
In EBMs training, the contrastive data is computed by MCMC-sampling, typically with Langevin dynamics \cref{eq:sgld}. In binary AT, the contrastive data is computed using the PGD attack \cref{eq:pgd_noproj}.
Comparing \cref{eq:pgd_noproj} with \cref{eq:sgld}, we find that both approaches compute the contrastive data by first initializing from some out-distribution data, and then performing gradient ascent on $f_\theta$. 
The main differences are that the PGD attack does not have the noise term, and makes use of normalized gradient.
%We first note that although Langevin dynamics requires the noise term to be a valid sampler, the noise term is not absolutely necessary when learning short-run EBMs~\cite{nijkamp2020anatomy}.
%Although the noise term in the SGLD sampler is not absolutely necessary when learning short-run EBMs~\cite{nijkamp2020anatomy}, it can encourage the exploration of different modes of the model distribution.
Intuitively, the noise term enables the sampler to explore different modes by helping gradient ascent escape local maxima. Although the PGD attack does not have the noise term, its ability to explore different modes can be enhanced by using a diverse $p_0$ dataset (\cref{fig:p0-ablation}).

% The normalized gradient in the PGD attack has several benefits. One is that it solves the problem of vanishing gradient when performing adversarial attacks using the cross-entropy loss, and the other is that 
% Normalized gradient in the PGD attack was originally introduced to deal with the problem of vanishing gradient when training with the cross-entropy loss. 
% As a side effect, 
In the PGD attack, as the normalized gradient vector has unit norm, the perturbation imposed on $x_i$ is $\lambda$; in a $K$ iterations of the update, the overall perturbation $\|x_i^* - x_i\|_2$ is always $\le \lambda K$. Hence with the PGD attack we can more easily control the distribution of the contrastive data.
% it provides a means for constraining the distribution of the perturbed data:
In contrast, Langevin dynamics adjusts $x_i$ in a scale that corresponds to the magnitude of the gradient of $f_\theta$ at $x_i$; when $f_\theta$ is updated during training, the overall perturbation may undergo a large change. This behavior of Langevin dynamics can be a source of some training stability issues~\cite{nijkamp2020anatomy}.
% An additional benefit of the normalized gradient in that it solves the vanishing gradient problem when performing adversarial attacks using the cross-entropy loss.
%Although the PGD attack may not correspond  exactly to a valid sampler, in practice we find it capable of producing samples that follow the distribution of the modeled data. On the other hand, in MCMC-based EBMs learning, due to the high computational cost of MCMC sampling,  it is not uncommon to use an invalid sampler which is short-run and non-convergent~\cite{nijkamp2019learning,nijkamp2020anatomy,du2019implicit,xie2016theory,Grathwohl2020Your}. 

\subsubsection{Gradient of the Training Objective.}
By definition \cref{eq:D_define}, the gradient of $D$'s training objective \cref{eq:D_objective} takes the form
\begin{equation}
	\label{eq:grad}
		\nabla_\theta J(D)  =  \mathbb{E}_{\mathrm{x} \sim p_\mathrm{data}}[(1-\sigma(f_\theta(x)))\nabla_\theta f_\theta(x)] - \mathbb{E}_{\mathrm{x}\sim p_T^*}[\sigma(f_\theta(x))\nabla_\theta f_\theta(x)].
\end{equation}
Comparing the above equation with \cref{eq:dll} we find both equations consisting of gradient terms that yield similar effects: the first term causes $f_\theta$ outputs on $p_\mathrm{data}$ samples to increase, and the second causes $f_\theta$ outputs on the contrastive samples to decrease.
% \\ &= -\mathbb{E}_{\mathrm{x} \sim p_\mathrm{data}} [(1-\sigma(-E(x)))\nabla_\theta E(x)] + \mathbb{E}_{\mathrm{x}\sim p_T} [\sigma(-E(x))\nabla_\theta E(x)]
Specifically, as $(1-\sigma(f_\theta(x)))$ and $\sigma(f_\theta(x))$ are scalars in the range 0 to 1, the two gradient terms in \cref{eq:grad} are respectively the scaled versions of the gradient terms in \cref{eq:dll}. 
%Although these scalars do not change the gradient update direction in the parameter space, they could cause \cref{eq:grad} to converge to different points than the maximum likelihood estimator \cref{eq:dll}.
It should be noted that although these scalars do not change the gradient update direction of individual terms in the model parameter space, the overall gradient update directions of \cref{eq:grad} and \cref{eq:dll} can be different. 
%to converge to different points than the maximum
% likelihood estimator \cref{eq:dll}.

\cref{eq:grad} also helps to understand why binary AT can only learn the support of the observed data. In \cref{eq:dll}, when $p_\theta(x)$ matches $p_\mathrm{data}$, the gradient cancels out and training terminates, whereas in \cref{eq:grad}, when $p_T^*$ matches $p_\mathrm{data}$ the gradient becomes $\mathbb{E}_{\mathrm{x} \sim p_\mathrm{data}} [(1-2\sigma(f_\theta(x)))\nabla_\theta f_\theta(x)]$ and only vanishes when $\sigma(f_\theta(x))=\frac{1}{2}$ everywhere on the support of $p_\mathrm{data}$. This result is consistent with \cref{pro:optimalD} and the 2D experiment result.

\subsection{Improved Training of Binary AT Generative Model}
\label{sec:improved}
%In this section we propose three improvements over the original binary AT algorithm.
%Building on insights from the analysis in \cref{sec:AT-properties}, we propose improved techniques for generative modeling with AT. Specifically, we propose to use a realistic and diverse out-distribution dataset as the $p_0$ dataset, and train with an unconstrained objective. We also address a failure mode of the training algorithm that we observe on CelebA-HQ~\cite{karras2017progressive}.
% \paragraph{Training stability}
% Grathwohl~\etal~\cite{Grathwohl2020Your} discussed thoroughly about the stability issue of MCMC-based EBM learning. In summary, . AT has been observed to be stable to train. However, we do observe one stability issue.
\subsubsection{Diverse $p_0$ Data.}
%The benefit of a diverse $p_0$ dataset can be confirmed from the 2D experiment (\cref{fig:2D} left panel versus right panel):  a diverse $p_0$ dataset promotes the learning of an energy function that is well-behaved (no spurious modes) in the entire data space.
%When $p_0$ is at the same time a real dataset, the model achieves out-of-distribution adversarial robustness (\cref{sec:applications} OOD detection) and learns informative gradient for transforming arbitrary out-distribution samples (not just noise samples) into valid samples of the modeled data (\cref{sec:applications} image-to-image translation).
As discussed in \cref{sec:mechanism},
a diverse $p_0$ dataset improve the PGD attack's ability to explore different local maxima of $D$. 
%a diverse $p_0$ dataset promotes the learning of an energy function that is well-behaved (no spurious modes) in the entire data space. 
To validate this 2D intuition generalizes to high dimensions, we evaluate the image generation performances of models trained with different $p_0$ datasets. \cref{fig:p0-ablation} shows that the best FID is obtained when $p_0$ is the most diverse dataset among the considered $p_0$ datasets. 
(Note that we have used the same setting of the PGD attack and source images dataset to  perform test-time generation in these three experiments.)

We follow existing work on adversarial training and use a $p_0$ dataset that contains real data samples to train the model. Using a real dataset (as opposed to a noise distribution) helps the model achieve out-of-distribution adversarial robustness (\cref{sec:applications} OOD detection) and learn informative gradient for transforming real out-distribution samples (not just noise samples) into valid samples of $p_\textrm{data}$. The latter can be a useful feature in image translation applications (\cref{sec:applications}).
% Based on the 2D simulation results  we propose to use a real and diverse out-distribution dataset as the $p_0$ dataset.
% Specifically, for CIFAR-10~\cite{krizhevsky2009learning}, we use the 80 million tiny images dataset~\cite{torralba200880} as $p_0$, and for $256\times 256$ resolution datasets we use the ImageNet~\cite{deng2009imagenet}.
The setting of $p_0$ in our experiments can be found in \cref{tab:compare}.
% We note that this setting is different from the original binary AT method~\cite{Yin2020GAT} where $p_0$ data is a mixture of data from a few number of classes.
% As we will demonstrate in the experiments, this setting of $p_0$ data causes the model to learn informative gradient for transforming arbitrary out-distribution samples (not just noise samples) into valid samples of the modeled data, which is a useful feature in image-to-image translation applications.
% The benefit of such a $p_0$ dataset is already clear in the 2D experiment.
% Another important difference between adversarial training and MCMC-based EBM training lies in $p_0$. 
% In adversarial training, $p_0$ is typically some out-of-distribution data (\eg, the mixture distribution in~\cite{Yin2020GAT}), while 
%We note that this setting of $p_0$ data is different from existing work on learning EBMs where $p_0$ is typically some noise distribution or a distribution very close to $p_\mathrm{data}$ (e.g., a distribution defined by a generator fitted on $p_\mathrm{data}$, see \cref{sec:related_work}).
%  a downsampled version of $p_\mathrm{data}$ or 

\subsubsection{Training With Unconstrained Perturbations.}
\begin{figure}
	\centering
	\begin{minipage}{.32\textwidth}
	\centering
	\includegraphics[width=\textwidth]{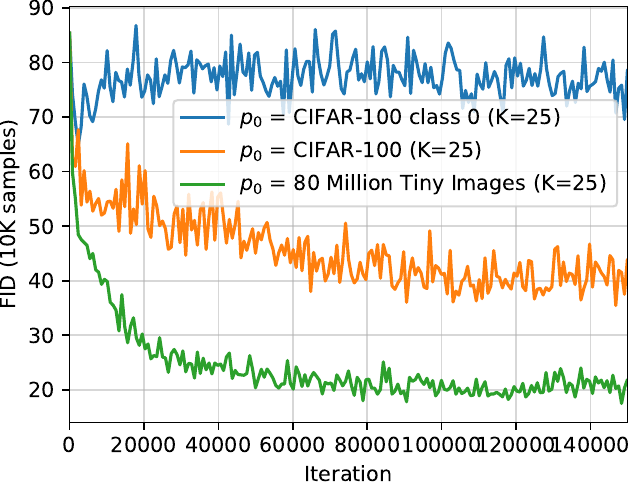}
	\caption{FID scores obtained with models trained with different $p_0$s on CIFAR-10.}
	\label{fig:p0-ablation}
	\end{minipage}%
	\hfill\hfill
	\begin{minipage}{.32\textwidth}
		\centering
		\includegraphics[width=\textwidth]{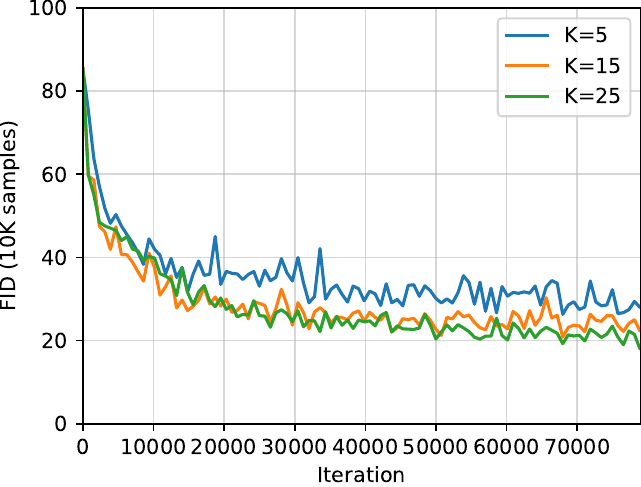}
		\caption{FID scores obtained with different $K$s in \cref{alg:maximin-solver} on CIFAR-10.}
		\label{fig:steps-ablation}
	\end{minipage}%
	\hfill
	\begin{minipage}{.32\textwidth}
		\centering
		\includegraphics[width=\textwidth]{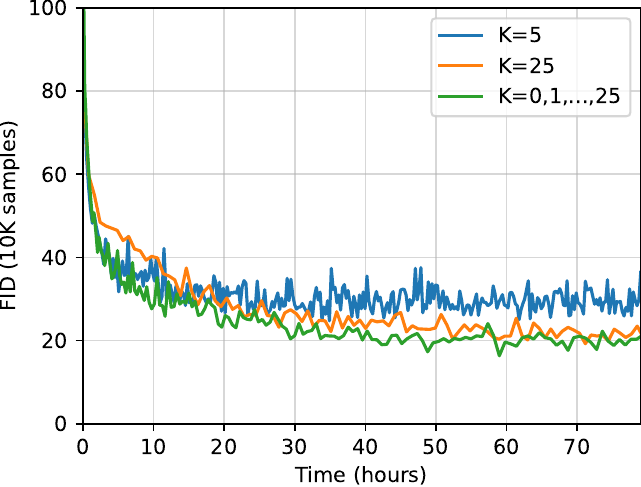}
		\caption{Progressive training vs. training with fixed-$K$ on CIFAR-10.}
		\label{fig:progressive-ablation}
	\end{minipage}
\end{figure}

Existing work on using adversarial training to train robust classifiers uses a small, fixed perturbation limit~\cite{madry2017towards}. In the generative modeling task, we would like the perturbed  $p_0$ data to travel in a larger space to find more local maxima. This can be achieved by taking more PGD attack steps ($K$) in step 3 of \cref{alg:maximin-solver}.
% unnecessarily constrains the distribution of the contrastive data.
%In this work we consider training with the unconstrained objective \cref{eq:maxmin-problem} instead of the $\epsilon$-constrained objective \cref{eq:gat-obj}.
\cref{fig:steps-ablation} shows that a larger $K$ results in better FID scores. 
%(We use step size 0.1 and therefore $K=5$ corresponds to $\epsilon=0.5$ which is the perturbation limit used in~\cite{santurkar2019image,wang2022a}.) 

The downside of a large $K$ is that it converges slower because more gradient steps are taken in each iteration (\cref{fig:progressive-ablation} $K=25$ vs. $K=5$). To improve the training efficiency we propose a mixed scenario in which we progressively increase the $K$ value during training.
We observe that this progressive training scenario converges faster than training with fixed-$K$  (\cref{fig:progressive-ablation} $K=0, 1, ..., 25$ vs. $K=25$). The pseudo code for progressive training is in \cref{alg:progressive-training}.
\begin{algorithm}[h!]
	\caption{\small Progressive Binary Adversarial Training}
	\label{alg:progressive-training}
	{\small
		\begin{algorithmic}[1]
			\FOR {$K$ in $[0, 1,\dots, N]$}
			\FOR {number of training iterations}
			\STATE Draw samples $\{ {x}_i\}_{i=1}^m$ from $p_\textrm{data}$, and samples $\{ x_i^{0} \}_{i=1}^m$ from $p_0$.
			%			\STATE For each $x\in \{x_i^{-k} \}_{i=1}^m$, compute the perturbed sample $x'$ by performing $K$ times update
			%			$\tilde{x}_i^{k+1}\leftarrow \tilde{x}_i^k - \gamma \frac{\nabla \log (1-D\left(\tilde{x}_i^k\right))}{\|\nabla \log \left(1-D\left(\tilde{x}_i^k\right)\right)\|_2}
			%			$ (at initialization $\tilde{x}_i^0 \leftarrow \tilde{x}_i$).
			\STATE Update $\{ x_i^{0} \}_{i=1}^m$ by performing $K$ steps unconstrained PGD attack~\cref{eq:pgd_noproj} on each sample. Denote the resulting samples as $\{ x_i^{*} \}_{i=1}^m$.
			\STATE  Update $D$ by maximizing
			$\frac{1}{m} \sum_{i=1}^m \log D({x_i}) + \frac{1}{m} \sum_{i=1}^m \log(1-D(x_i^*))$.
			\ENDFOR
			\ENDFOR
		\end{algorithmic}
	}
\end{algorithm}

\begin{wrapfigure}{r}{0.4\textwidth}
	\centering
	\includegraphics[width=\linewidth]{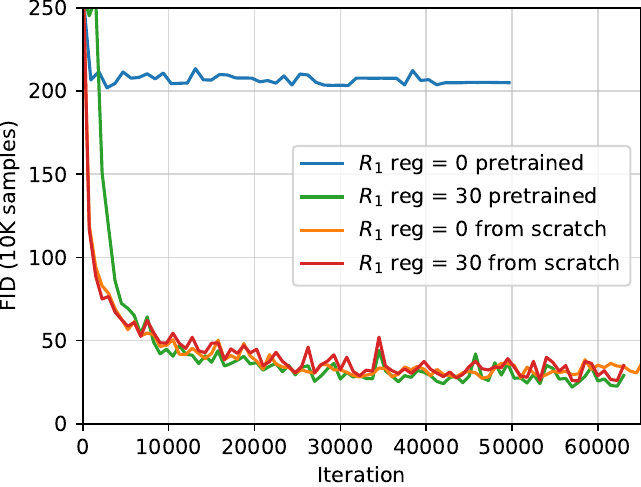}
	\caption{The effect of $R_1$ regularization on CelebA-HQ.}
	\label{fig:r1-ablation}
\end{wrapfigure}

%\begin{figure}
%	\centering
%	\begin{minipage}{.25\textwidth}
%		\centering
%		\includegraphics[width=\textwidth]{figs/ablation/p0}
%		\caption{FID scores obtained with different settings of $p_0$.}
%		\label{fig:p0_ablation}
%	\end{minipage}%
%	\hfill
%	\begin{minipage}{.25\textwidth}
%		\centering
%		\includegraphics[width=\textwidth]{figs/ablation/r1reg}
%		\caption{Progressive training vs fixed-$K$ training on CIFAR-10.}
%		\label{fig:progressive-ablation}
%	\end{minipage}
%\end{figure}

\noindent\textbf{Regularization.}
While other generative models typically require some forms of regularization, the proposed model can be trained successfully without using any regularization. One trick that we find beneficial for achieving better FID (\cref{fig:r1-ablation}, $R_1$ reg = 30, pretrained vs. from scratch) is to pretrain the $D$ model on the ImageNet classification task. (This requires adding auxiliary output nodes which are ignored when later training the $D$ model.) When using the pretrained model, we find it necessary to use $R_1$ regularization~\cite{mescheder2018training}, otherwise the FID stops improving after a few hundred iterations (\cref{fig:r1-ablation}, $R_1$ reg = 0 pretrained). Note that when $D$ is trained from scratch, $R_1$ regularization is not strictly  required, but adding the regularizer does not hurt the performance (\cref{fig:r1-ablation}, $R_1$ reg = 0 from scratch vs. $R_1$ reg = 30 from scratch).
%When we use \cref{alg:maximin-solver} to train on CelebA-HQ 256~\cite{karras2017progressive}, we observe that the binary classification accuracy quickly reaches 100\% after a few hundred iterations. Meanwhile the mean $l^2$ distance between the original $p_0$ samples and perturbed $p_0$ samples is only a small fraction of $\lambda K$, and the perturbed $p_0$ samples  show no meaningful features of human faces. 
%Although the symptoms are different, we believe this failure has the same root cause as mode collapse in GANs.
%In the adversarial game \cref{eq:maxmin-problem}, the $D$ model can learn to pick up a handful of low-level features which correlate well with the 0-1 labels to solve the binary classification task. With these features the $D$ model cannot provide sufficient gradient for the adversarial $p_0$ data to move towards the manifold of $p_\textrm{data}$. Without the accurate information, the $p_0$ adversary fails to compete with $D$, and the $D$ model starts to dominate the adversarial game.
%We follow GANs literature~\cite{mescheder2018training,brock2018large,karras2020training} and address this issue by using $R_1$ regularization~\cite{mescheder2018training}  to regularize the sensitivity of $D$'s output to the input features (\cref{fig:r1-ablation} shows the effect of this regularizer).
% We do not observe other stability issues other than this failure mode.

\section{Experiments}
In this section we provide an empirical evaluation of the proposed AT generative model.
We first evaluate the approach's image generation performance and then demonstrate its applications to image-to-image translation and worst-case out-of-distribution detection.  We further provide an analysis of the proposed approach's training stability in \cref{sec:stability}. 

In the supplementary materials we provide experiment setup details including model architectures, training hyperparameters, sample generation settings, and evaluation protocols. We also include additional results including sampling efficiency analysis, uncurated image generation and image translation results, and demonstration of applications to denoising, inpainting, and compositional visual generation~\cite{du2020compositional}.  The interpolation results and nearest neighbor analysis in the supplementary materials suggest that our model captures the manifold structure of the observed data, as opposed to simply memorizes the data samples.

\subsection{Image Generation}
\cref{tab:cifar10_scores} shows that on CIFAR-10~\cite{krizhevsky2009learning}  our approach achieves the best Inception Score (IS)~\cite{salimans2016improved} and FID~\cite{heusel2017gans} among AT generative models. Our approach also improves over state-of-the-art explicit EBMs in terms of IS, and at the same time has a slightly worse FID. 
%(Generated samples can be found in the supplementary materials.) 
Compared to VAEBM~\cite{xiao2021vaebm}, our method does not require an auxiliary model to train, and  has better test time sampling efficiency (see supplementary materials). 
% The energy-based model of Diffusion Recovery~\cite{gao2021learning} is essentially a conditional model that defines the conditional distribution of a noisy sample given the same sample at a higher noise level.  
Diffusion Recovery~\cite{gao2021learning} trains a sequence of conditional EBMs, with each one defining the conditional distribution of a noisy sample given the same sample at a higher noise level. 
Similar to score-based approaches, these conditional EBMs do not directly model the data distribution of the observed data, so it is unclear 
how these models can be applied to 
% what utilities they can provide to
tasks which require explicit knowledge of the data distribution (e.g., OOD detection).

\begin{figure}[h!]
	\centering
	\includegraphics[width=\textwidth]{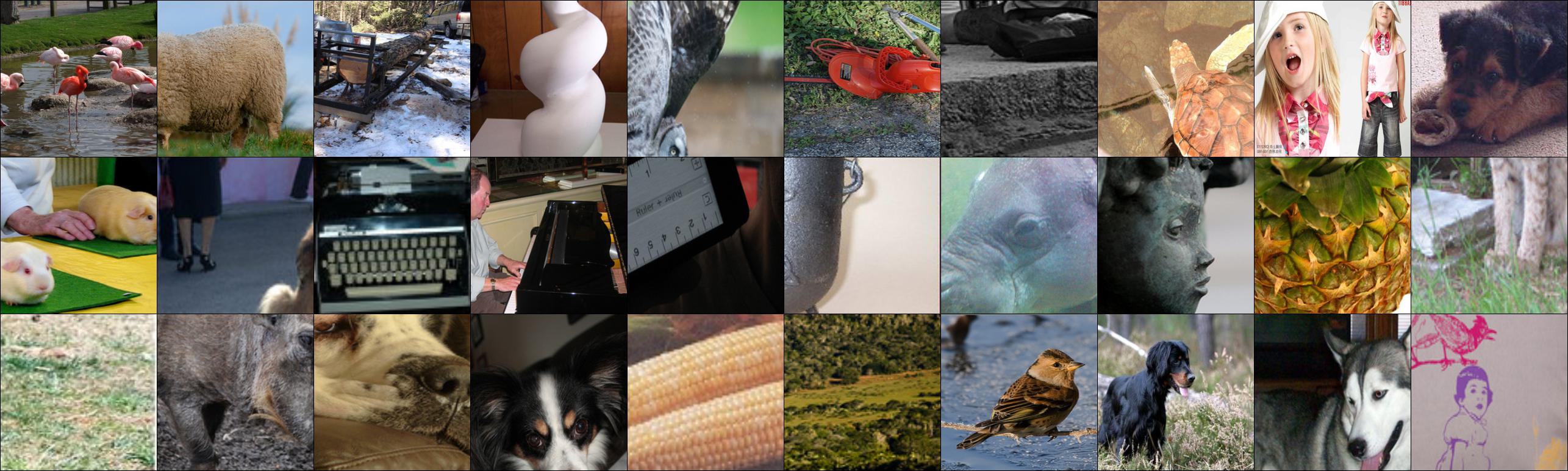}
	\includegraphics[width=0.1\textwidth]{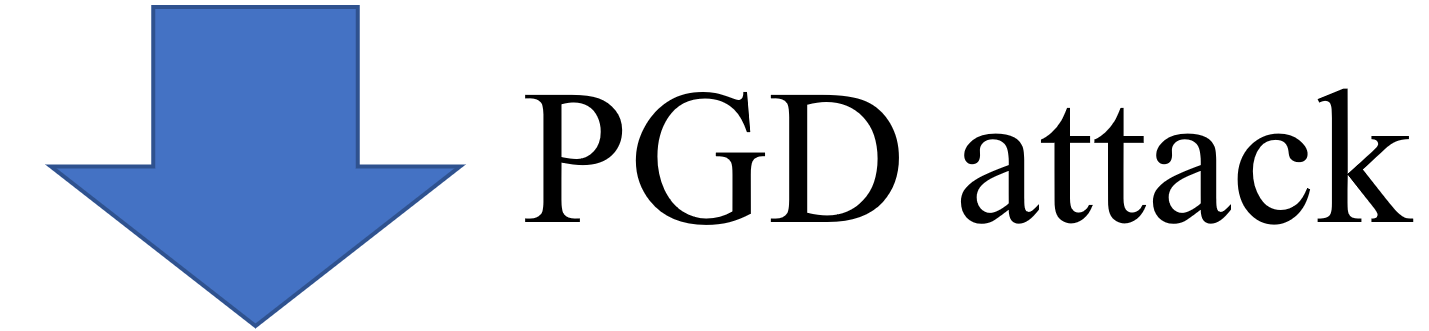}
%	{\scriptsize PGD attack}
	\includegraphics[width=\textwidth]{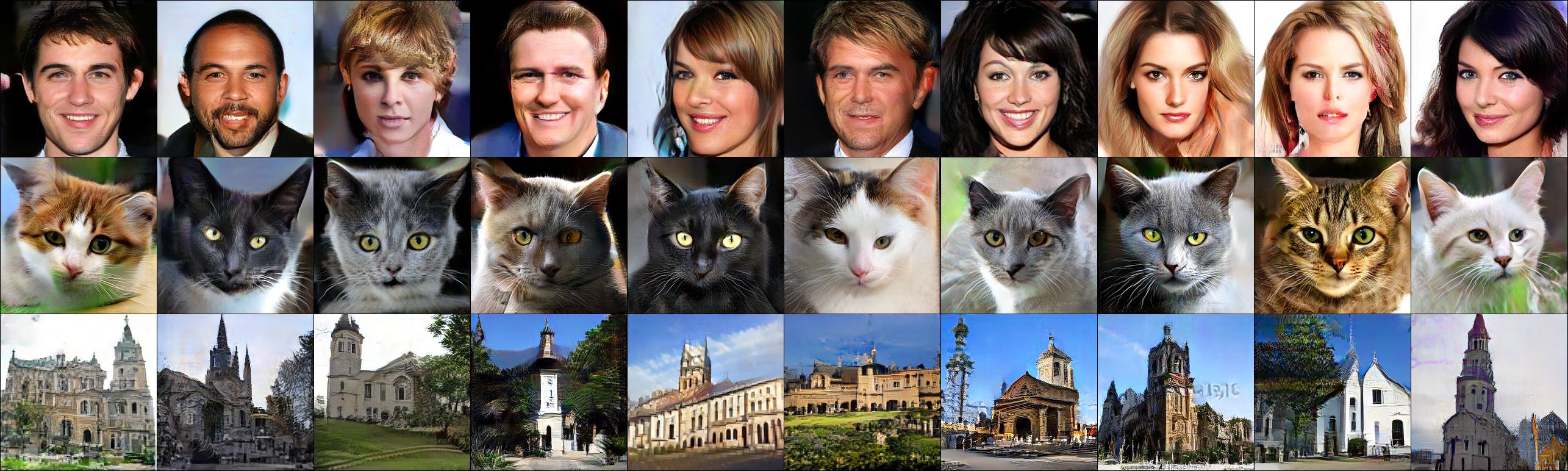}
	\caption{Source images (top panel) and generated images (bottom panel, $256\times 256$ resolution) on CelebA-HQ, AFHQ-CAT, and LSUN Church.}
	\label{fig:face_samples}
\end{figure}

\cref{tab:celebahq-scores} shows that on CelebA-HQ 256~\cite{karras2017progressive} our method 
% performs on par with 
outperforms or is on par with
state-of-the-art generative models except GANs. 
%Our method similarly falls below GANs on AFHQ-CAT 256~\cite{choi2020stargan}. 
On LSUN Church~\cite{yu2015lsun} our method outperforms a latest energy-based model VAEBM~\cite{xiao2021vaebm} (the authors only provided the $64\times64$ result), but falls bellow DDPM and GANs.
% \cref{fig:celebahq-samples} and \cref{fig:afhq-cat-samples} 
\cref{fig:face_samples}
shows sample image generation results.
%although the generated samples contain artifacts and hence have a large room for improvement.
% hence it is unclear how these models can be applied to cenrtain tasks considered in this work (\eg, out-of-distribution detection).

\setlength{\tabcolsep}{4pt}
\begin{table}
	\begin{minipage}{0.48\linewidth}
	\caption{IS and FID scores on CIFAR-10}
	\label{tab:cifar10_scores}
	\resizebox{\textwidth}{!}{%
		\begin{tabular}{llll} 
			\toprule
			& Approach                                           & IS$\uparrow$ & FID$\downarrow$  \\ 
			\midrule
\multirow{3}{*}{AT-based} & Ours                                                      & 9.10 & 13.21 \\
& CEM~\cite{wang2022a}                                      & 8.68 & 36.4  \\
& JEAT~\cite{zhu2021towards}                                      & 8.80 & 38.2  \\
& Adv. Robust Classifier~\cite{santurkar2019image} & 7.5  & -     \\
			\midrule
			\multirow{9}{*}{\begin{tabular}[c]{@{}l@{}}Explicit\\ EBMs\end{tabular}} 
			& Diffusion Recovery~\cite{gao2021learning} & 8.30   &  9.58    \\
			& VAEBM~\cite{xiao2021vaebm}                      & 8.43 & 12.19 \\
			& CoopFlow~\cite{xie2022a}                      & - & 15.80 \\
			& CF-EBM~\cite{zhao2021learning} &  -  &   16.71   \\ 
			& ImprovedCD~\cite{Du2021ImprovedCD} & 7.85 & 25.1 \\
			& VERA~\cite{grathwohl2021no} &  -  &   27.5   \\ 
			& EBMs-VAE~\cite{xie2021learning} &  6.65  &   36.2   \\ 
			& JEM~\cite{Grathwohl2020Your} &  8.76  &   38.4   \\ 
			& IGEBM (Ensemble)~\cite{du2019implicit} &  6.78  &   38.2   \\
			& Short-Run EBMs~\cite{nijkamp2019learning} &  6.21 &  44.16 \\
			\midrule
			\multirow{4}{*}{GANs} 
			& StyleGAN2 w/o ADA~\cite{karras2020training}  &  8.99  &  9.9    \\
			& BigGAN~\cite{brock2018large}  & 9.22   & 14.73     \\
			& SNGAN~\cite{miyato2018spectral}  & 8.22   &  21.7    \\
			& WGAN-GP~\cite{gulrajani2017improved} & 7.86 & 36.4 \\
			\midrule
			\multirow{4}{*}{Score-based} 
			& SDE~\cite{song2021scorebased}  & 9.89   & 2.20     \\
			& DDPM~\cite{ho2020denoising}  &  9.46  &  3.17    \\
			& NCSNv2~\cite{song2020improved}  & 8.4   &  10.87    \\
			& NCSN~\cite{song2019generative} & 8.87 & 25.32 \\
			\bottomrule
		\end{tabular}%
	}
	\end{minipage}\hfill
	\begin{minipage}{0.48\linewidth}
	\caption{FID scores on CelebA-HQ 256, AFHQ-CAT~\cite{choi2020stargan}, and LSUN Church 256}
	\label{tab:celebahq-scores}
	\resizebox{\textwidth}{!}{%
		\begin{tabular}{@{}lll@{}}
			\toprule
			Dataset & Approach & FID$\downarrow$ \\ \midrule
			\multirow{7}{*}{CelebA-HQ 256} & Ours & 17.31 \\
			& VAEBM~\cite{xiao2021vaebm} & 20.38 \\
			& CF-EBM~\cite{zhao2021learning} ($128\times 128$) & 23.50 \\ \cmidrule(l){2-3} 
			& NVAE~\cite{vahdat2020nvae} & 45.11 \\
			& GLOW~\cite{kingma2018glow} & 68.93 \\ \cmidrule(l){2-3} 
			& Adversarial Latent AE~\cite{pidhorskyi2020adversarial} & 19.21 \\
			& ProgressiveGAN~\cite{karras2017progressive} & 8.03 \\ \midrule
			\multirow{2}{*}{AFHQ-CAT} & Our ($256\times 256$) & 13.35 \\
			& StyleGAN2 ($512\times 512$)~\cite{karras2020training} & 5.13 \\ \midrule
			\multirow{5}{*}{LSUN Church}
			& VAEBM ($64\times 64$)~\cite{xiao2021vaebm} & 13.51 \\
			& Ours ($64\times 64$) & 10.84 \\
			& Ours ($256\times 256$) & 14.87 \\
			& DDPM ($256\times 256$)~\cite{ho2020denoising} & 7.89 \\  	
			& ProgressiveGAN ($256\times 256$)~\cite{karras2017progressive} & 6.42 \\  					
			 \bottomrule
		\end{tabular}
		}
	\end{minipage}
\end{table}
\setlength{\tabcolsep}{1.4pt}

\begin{figure}[h!]
	\scriptsize
	\centering
	\begin{minipage}{.48\textwidth}
		\centering
		\includegraphics[width=\linewidth]{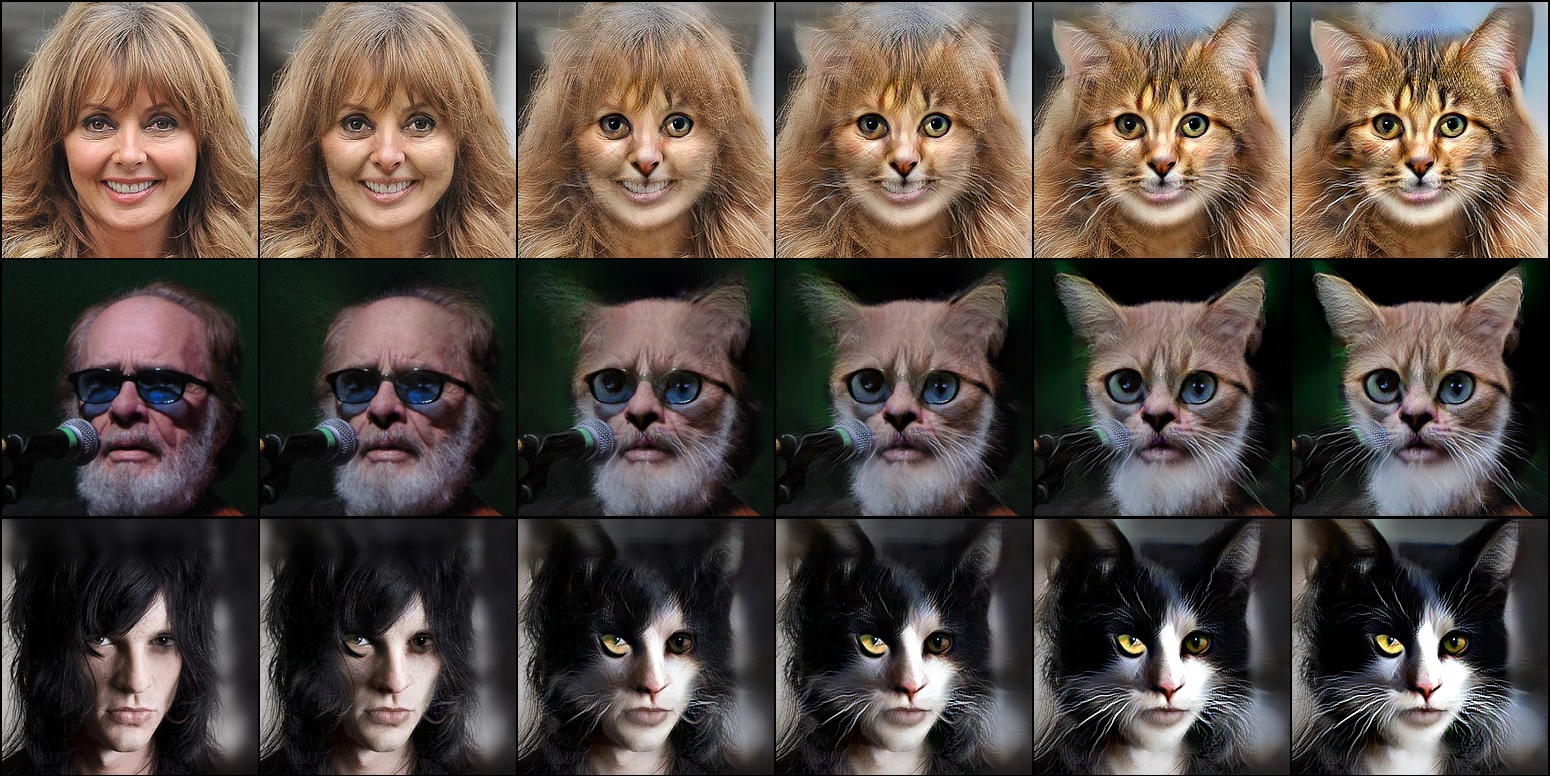} 		
		\includegraphics[width=0.4\linewidth]{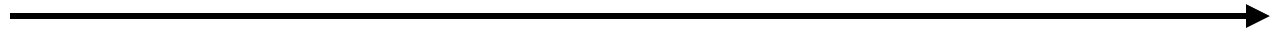} 
		\\ PGD attack
%		\\ $\overrightarrow{\text{PGD attack}}$
%				\caption*{jfkdsf}
	\end{minipage}%
%		\vrule
	\hfill
%	\hspace
	\begin{minipage}{.48\textwidth}
		\centering
		\includegraphics[width=\linewidth]{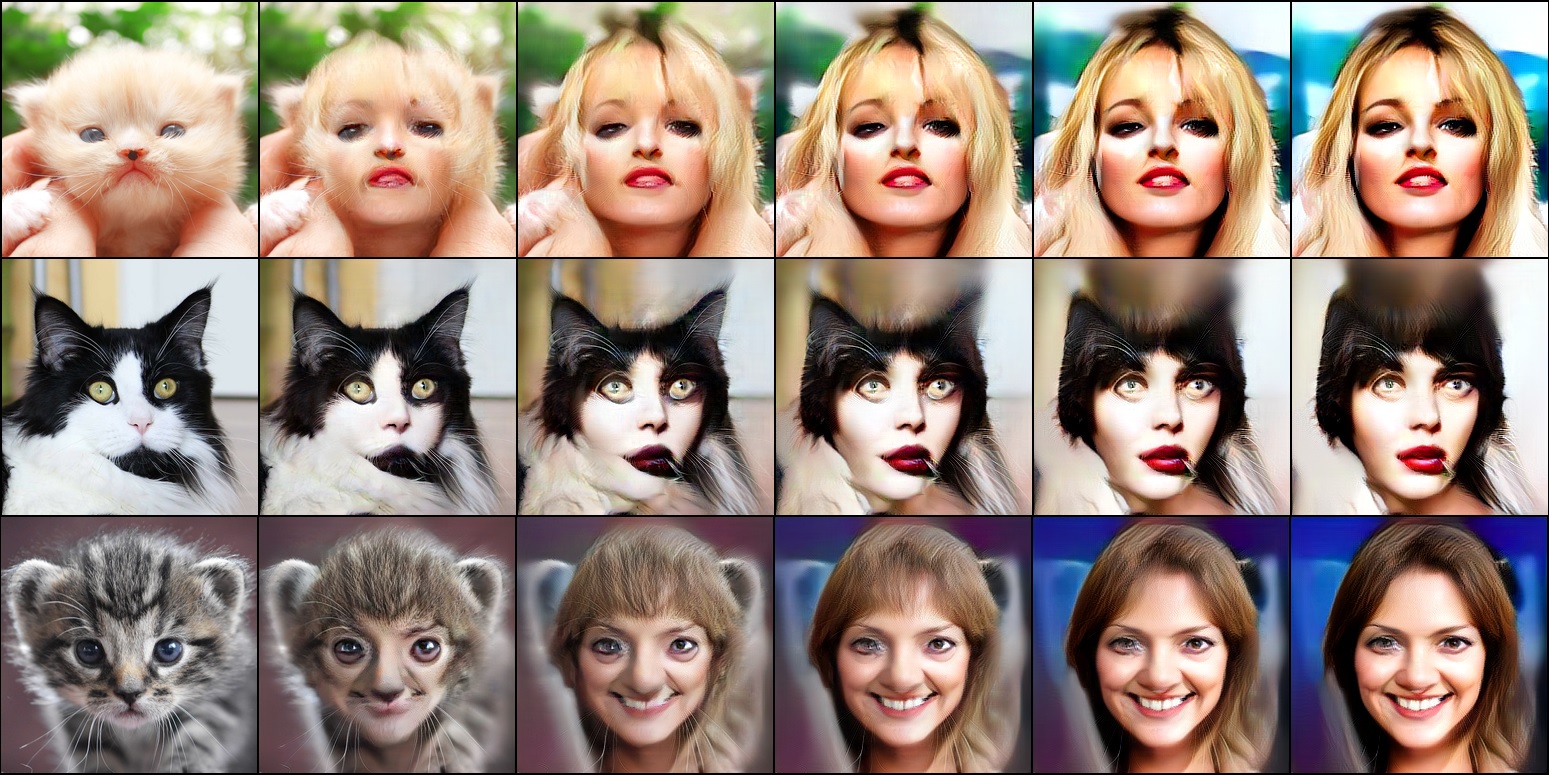}
		\includegraphics[width=0.4\linewidth]{figs/rightarrow} 
		\\ PGD attack
	\end{minipage}%
%\\  $\overrightarrow{\text{PGD attack}}$
	\caption{Image-to-image translation demonstration
	}
	\label{fig:translation}
\end{figure}

%\begin{figure}[h!]
%	\centering
%	\begin{subfigure}[t]{0.48\textwidth}
%		\centering
%		\includegraphics[width=\textwidth]{figs/celeba_gen.png}
%		%			\caption{Original samples}
%	\end{subfigure}
%	\begin{subfigure}[t]{0.48\textwidth}
%		\centering
%		\includegraphics[width=\textwidth]{figs/cat_gen_updated.png}
%	\end{subfigure}
%	
%	\caption{Generated samples on CelebA-HQ 256 and AFHQ-CAT 256.}
%	%		We perform the targeted attack by maximizing the logit output of the targeted class, using $L_\infty\ \epsilon=0.4$ constrained PGD attack of steps 100 and step size 0.01. Both classifiers are trained with $L_\infty\ \epsilon=0.3$ constraint.
%	\label{fig:face_samples}
%\end{figure}

%\begin{figure}[h!]
%	\centering
%	\includegraphics[width=\textwidth]{figs/celeba_gen3.jpg}
%		\includegraphics[width=\textwidth]{figs/cat_gen2.jpg}
%			\includegraphics[width=\textwidth]{figs/church_gen.jpg}
%	\caption{Generated samples ($256\times 256$) on CelebA-HQ, AFHQ-CAT, and LSUN Church.}
%	\label{fig:face_samples}
%\end{figure}

%\begin{figure}[h!]
%	\centering
%	\includegraphics[width=\textwidth]{figs/trans.png}
%	\caption{Sample image translation results (note that no source domain images are used during the training of the models).}
%	\label{fig:translation}
%\end{figure}

\subsection{Applications}
\label{sec:applications}
\subsubsection{Image-To-Image Translation.}
\cref{fig:translation} shows that  the  AFHQ-CAT model can be used to transform CelebA-HQ images into cat images, and vice-versa.
% We note that the models are not specially trained  to transform between these two domains.
Note that these two models are trained independently without knowledge of the source domain, indicating that our approach may generalize better to unseen data than approaches (e.g., pix2pix~\cite{isola2017image}, CycleGAN~\cite{zhu2017unpaired}, and StarGAN~\cite{choi2020stargan}) that explicitly use the source domain dataset to train the model.
% This separates our approach from existing solutions such as pix2pix~\cite{isola2017image}, CycleGAN~\cite{zhu2017unpaired}, and StarGAN~\cite{choi2020stargan} where the source domain must be known at training time.
% In addition, our approach allows the user to choose the strength of transformation, which is more flexible than approaches that use a fixed generator.
The translation results may be further improved by finetuning the trained model on the source domain dataset, or including the source domain data in the $p_0$ dataset during training.
%Synthesizing content with a dynamic process of gradient descent is also more flexible than using a fixed generator: it allows the user to choose how much transformation to apply, or create cinematic effect from intermediate results.
The proposed approach is also more flexible than approaches that employs a fixed generator, as it allows the user to choose how much transformation to apply, and/or create cinematic effect from intermediate transformation results.
% In terms of application, our approach does not have a fixed generator but rather uses a dynamic process of gradient descent to synthesize content. This allows the user to choose how much transformation to apply to the content, or create cinematic effect from intermediate results.
%Demonstration of applications to denosing and inpainting, and more translation results are provided in the supplementary materials.

\subsubsection{Worst-Case Out-Of-Distribution Detection.}
\label{para:ood_detection}
Out-of-distribution (OOD) detection is a classic application of EBMs. Some recent works~\cite{sehwag2019better,meinke2019towards,bitterwolf2020certifiably,augustin2020adversarial} find that EBMs and some other OOD detection approaches such as OE~\cite{hendrycks2018deep} are vulnerable to adversarial OOD inputs. Given the challenge of adversarial inputs, many authors attempt to address OOD detection in an adversarial setting (also known as worst-case, or adversarial OOD detection)~\cite{sehwag2019better,hein2019relu,meinke2019towards,bitterwolf2020certifiably,augustin2020adversarial}. Among these works, RATIO~\cite{augustin2020adversarial} is a state-of-the-art method that combines in- and out-distribution adversarial training to obtain a robust classifier that has uniform outputs in a neighborhood around OOD samples. Similar to OE, RATIO employs 80 million tiny images~\cite{torralba200880} as the out-distribution dataset to train the model.

\cref{tab:cifar10_ood} shows that our model achieves comparable OOD detection performance to the state-of-the-art method of RATIO~\cite{augustin2020adversarial}.
OE, RATIO, and JEM all perform OOD detection by utilizing a classifier that has low confidence predictions on the out-distribution data (clean or worst-case).
In RATIO, the worst-case out-distribution data is computed by performing the PGD attack on 80 million tiny images, whereas in JEM it is computed via Langevin dynamics initialized from uniform random noise.
% It can be seen that JEM's out-distribution adversarial robustness is limited.
RATIO and our method's strong out-distribution adversarial robustness demonstrates the benefit of using a real and diverse out-distribution dataset to train the model. Our method does not make use of class labels and therefore can be considered as a binary variant of RATIO.
On CelebA-HQ 256, AFHQ-CAT, and LSUN-Church our model similarly achieves strong out-distribution adversarial robustness (supplementary materials).
These results suggest that our generative model can be applied to detect both naturally occurring OOD data and adversarially created malicious content.  
%(A review of related work on out-of-distribution detection is provided in the supplementary materials.)
% trains a binary classifier to separate in-distribution data from worst-case out-distribution data.
% Apart from standard datasets, we follow \cite{augustin2020adversarial} and consider smoothed noise \cite{hein2019relu} as a type of OOD data; as the code of \cite{augustin2020adversarial} is not publicly available, we use the implementation from \cite{meinke2019towards}
% \footnote{\url{https://github.com/AlexMeinke/certified-certain-uncertainty/blob/master/utils/dataloaders.py}}
%  to compute the noise.

\setlength{\tabcolsep}{4pt}
\begin{table}[h!]
	\centering
	\caption{CIFAR-10 standard and worst-case OOD detection results (AUC scores).
		We use the same settings of AutoAttack~\cite{Croce2020ReliableEO}, number of OOD samples, and perturbation limit as in \cite{augustin2020adversarial} to compute adversarial OOD samples.
		% Following Augustin~\etal~\cite{augustin2020adversarial}, we compute adversarial OOD samples by  maximizing the model output in an $l^2$-ball of radius 1.0 around OOD samples via Auto-PGD~\cite{Croce2020ReliableEO} with 100 steps and 5 random restarts.
		Results of OE, JEM, and RATIO are from \cite{augustin2020adversarial}.}
	\label{tab:cifar10_ood}
%	\resizebox{0.6\columnwidth}{!}{%
%		\begin{tabular}{lllll}
%			\toprule
%			\multirow{2}{*}{OOD dataset} &
%			\multicolumn{3}{c}{Classifier-based approach} &
%			\multicolumn{1}{c}{\multirow{2}{*}{Ours}} \\ \cmidrule(lr){2-4}
%			&
%			OE~\cite{hendrycks2018deep} &
%			JEM~\cite{Grathwohl2020Your} &
%			RATIO~\cite{augustin2020adversarial} &
%			\multicolumn{1}{c}{} \\ \hline
%			SVHN          & {99.4} / 0.6 & 89.3 / 7.3  & 96.5 / {81.3}   & 93.0 / 81.6   \\
%			CIFAR-100     & 91.4 / 2.7   & 87.6 / 19.2 & {91.6} / {73.0} & 88.3 / {70.3} \\
%			ImageNet      & 89.8 / 1.5   & 86.7 / 21.2 & {91.3} / 73.5   & 89.7 / 71.2   \\
%			Uniform Noise & 99.5 / 43.1  & 11.8 / 2.5  & 99.9 / 99.8     & 100  / {100}  \\ \bottomrule
%		\end{tabular}%
\begin{tabular}{@{}lllll@{}}
	\toprule
	\multirow{2}{*}{OOD dataset} &
	\multicolumn{3}{c}{Classifier-based approach} &
	\multicolumn{1}{c}{\multirow{2}{*}{Ours}} \\ \cmidrule(lr){2-4}
	&
	OE~\cite{hendrycks2018deep} &
	JEM~\cite{Grathwohl2020Your} &
	RATIO~\cite{augustin2020adversarial} &
	\multicolumn{1}{c}{} \\ \midrule
	\multicolumn{5}{l}{Standard OOD detection}                \\ \midrule
	SVHN          & {99.4} & 89.3 & 96.5   & 93.5   \\
	CIFAR-100     & 91.4   & 87.6 & {91.6} & 88.7   \\
	ImageNet      & 89.8   & 86.7 & {91.3} & 89.7   \\
	Uniform Noise & 99.5   & 11.8 & 99.9   & 100    \\ \midrule
	\multicolumn{5}{l}{Worst-case OOD detection}              \\ \midrule
	SVHN          & 0.6    & 7.3  & {81.3} & 83.0   \\
	CIFAR-100     & 2.7    & 19.2 & {73.0} & {70.6} \\
	ImageNet      & 1.5    & 21.2 & 73.5   & 72.5   \\
	Uniform Noise & 43.1   & 2.5  & 99.8   & {100}  \\ \bottomrule
\end{tabular}%
%	}
	
\end{table}
\setlength{\tabcolsep}{1.4pt}

\subsection{Training Stability Analysis}
\label{sec:stability}
% \paragraph{Using the PGD attack in standard EBMs training}
% In \cref{sec:connection} we discuss how the PGD attack can be viewed as a sampler of the model distribution.
% In this ablation we study whether the PGD attack can be used with the EBMs training objective~\cref{eq:dll}.
To gain some insight into the training stability of our approach we investigate whether the PGD attack can be used with the EBMs training objective~\cref{eq:dll}.
Specifically, in \cref{alg:progressive-training}, we perform step 5's update on $\theta$ using the  gradient  $\nabla_\theta (\frac{1}{m} \sum_{i=1}^m f_\theta(x_i) - \frac{1}{m} \sum_{i=1}^m f_\theta(x_i^*)).$
%\begin{equation}
%	\nabla_\theta (\frac{1}{m} \sum_{i=1}^m f_\theta(x_i) - \frac{1}{m} \sum_{i=1}^m f_\theta(x_i^*)).
%\end{equation}
We observe that even under a small learning rate of $1e-6$, $\frac{1}{m} \sum_{i=1}^m f_\theta(x_i) - \frac{1}{m} \sum_{i=1}^m f_\theta(x_i^*)$ quickly increases and eventually overflows.
% This training divergence phenomenon is also recorded in other EBMs works.
This suggests that the stability of the AT approach can be largely attributed to the log-likelihood objective~\cref{eq:D_objective}. We argue that the stability is due to the gradient cancelling effect of this objective:
% when $f_\theta$ has large positive outputs on $p_\mathrm{data}$ samples, or large negative outputs on $p_T^*$ samples, the corresponding scalar ($1-\sigma(f_\theta(x))$ or $\sigma(f_\theta(x))$) approaches 0, and thus the scaled gradient in \cref{eq:grad} vanishes.
when $f_\theta$ has a large positive output on a sample $x\sim p_\mathrm{data}$, $1-\sigma(f_\theta(x))$ approaches 0 and therefore the corresponding scaled gradient in \cref{eq:grad} vanishes, and similarly $\sigma(f_\theta(x^*))\nabla_\theta f_\theta(x^*)$ vanishes when $f_\theta$ has a large negative output on a sample $x^*\sim p_T^*$.
In contrast, the EBMs objective~\cref{eq:dll} does not have constraints on $f_\theta$'s outputs and is therefore prone to divergence.

\section{Conclusion}
We have studied an AT-based approach to learning EBMs. Our analysis shows that binary AT learns a special kind of energy function that models the support of the observed data, and the training procedure can be viewed as an approximate maximum likelihood learning algorithm.
% We proposed improved techniques for generative modeling with AT, and demonstrated that the proposed method provides competitive generation performance to explicit EBMs, has competitive sampling efficiency, and is stable to train. We have also demonstrated the proposed approach's applications to denosing, inpainting, image translation, and worst-case out-of-distribution detection.
We proposed improved techniques for generative modeling with AT, and demonstrated that the proposed method provides competitive generation performance to explicit EBMs, has competitive sampling efficiency, is stable to train, and is well-suited for image translation tasks.
The proposed approach's strong out-distribution adversarial robustness suggests its potential application to detecting abnormal inputs and/or adversarially created fake content.

\clearpage
\bibliographystyle{splncs04}

\bibliography{ref}

\begin{thebibliography}{10}
\providecommand{\url}[1]{\texttt{#1}}
\providecommand{\urlprefix}{URL }
\providecommand{\doi}[1]{https://doi.org/#1}

\bibitem{arbel2021generalized}
Arbel, M., Zhou, L., Gretton, A.: Generalized energy based models. In:
  International Conference on Learning Representations (2021),
  \url{https://openreview.net/forum?id=0PtUPB9z6qK}

\bibitem{augustin2020adversarial}
Augustin, M., Meinke, A., Hein, M.: Adversarial robustness on in-and
  out-distribution improves explainability. In: European Conference on Computer
  Vision. pp. 228--245. Springer (2020)

\bibitem{bitterwolf2020certifiably}
Bitterwolf, J., Meinke, A., Hein, M.: Certifiably adversarially robust
  detection of out-of-distribution data. Advances in Neural Information
  Processing Systems  \textbf{33} (2020)

\bibitem{brock2018large}
Brock, A., Donahue, J., Simonyan, K.: Large scale {GAN} training for high
  fidelity natural image synthesis. In: International Conference on Learning
  Representations (2019), \url{https://openreview.net/forum?id=B1xsqj09Fm}

\bibitem{ceylan2018conditional}
Ceylan, C., Gutmann, M.U.: Conditional noise-contrastive estimation of
  unnormalised models. In: International Conference on Machine Learning. pp.
  726--734. PMLR (2018)

\bibitem{choi2020stargan}
Choi, Y., Uh, Y., Yoo, J., Ha, J.W.: Stargan v2: Diverse image synthesis for
  multiple domains. In: Proceedings of the IEEE/CVF Conference on Computer
  Vision and Pattern Recognition. pp. 8188--8197 (2020)

\bibitem{Croce2020ReliableEO}
Croce, F., Hein, M.: Reliable evaluation of adversarial robustness with an
  ensemble of diverse parameter-free attacks. In: ICML (2020)

\bibitem{cubuk2019autoaugment}
Cubuk, E.D., Zoph, B., Mane, D., Vasudevan, V., Le, Q.V.: Autoaugment: Learning
  augmentation strategies from data. In: Proceedings of the IEEE/CVF Conference
  on Computer Vision and Pattern Recognition. pp. 113--123 (2019)

\bibitem{deng2009imagenet}
Deng, J., Dong, W., Socher, R., Li, L.J., Li, K., Fei-Fei, L.: Imagenet: A
  large-scale hierarchical image database. In: 2009 IEEE conference on computer
  vision and pattern recognition. pp. 248--255. Ieee (2009)

\bibitem{du2020compositional}
Du, Y., Li, S., Mordatch, I.: Compositional visual generation with energy based
  models. Advances in Neural Information Processing Systems  \textbf{33},
  6637--6647 (2020)

\bibitem{Du2021ImprovedCD}
Du, Y., Li, S., Tenenbaum, J.B., Mordatch, I.: Improved contrastive divergence
  training of energy based models. In: ICML (2021)

\bibitem{du2019implicit}
Du, Y., Mordatch, I.: Implicit generation and modeling with energy based
  models. In: Advances in Neural Information Processing Systems. vol.~32
  (2019),
  \url{https://proceedings.neurips.cc/paper/2019/file/378a063b8fdb1db941e34f4bde584c7d-Paper.pdf}

\bibitem{engstrom2019adversarial}
Engstrom, L., Ilyas, A., Santurkar, S., Tsipras, D., Tran, B., Madry, A.:
  Adversarial robustness as a prior for learned representations. arXiv preprint
  arXiv:1906.00945  (2019)

\bibitem{gao2021learning}
Gao, R., Song, Y., Poole, B., Wu, Y.N., Kingma, D.P.: Learning energy-based
  models by diffusion recovery likelihood. In: International Conference on
  Learning Representations (2021),
  \url{https://openreview.net/forum?id=v_1Soh8QUNc}

\bibitem{goodfellow2014generative}
Goodfellow, I., Pouget-Abadie, J., Mirza, M., Xu, B., Warde-Farley, D., Ozair,
  S., Courville, A., Bengio, Y.: Generative adversarial nets. In: Advances in
  neural information processing systems. pp. 2672--2680 (2014)

\bibitem{Grathwohl2020Your}
Grathwohl, W., Wang, K.C., Jacobsen, J.H., Duvenaud, D., Norouzi, M., Swersky,
  K.: Your classifier is secretly an energy based model and you should treat it
  like one. In: International Conference on Learning Representations (2020),
  \url{https://openreview.net/forum?id=Hkxzx0NtDB}

\bibitem{grathwohl2021no}
Grathwohl, W.S., Kelly, J.J., Hashemi, M., Norouzi, M., Swersky, K., Duvenaud,
  D.: No mcmc for me: Amortized sampling for fast and stable training of
  energy-based models. In: International Conference on Learning Representations
  (2021), \url{https://openreview.net/forum?id=ixpSxO9flk3}

\bibitem{gulrajani2017improved}
Gulrajani, I., Ahmed, F., Arjovsky, M., Dumoulin, V., Courville, A.C.: Improved
  training of wasserstein gans. In: Advances in Neural Information Processing
  Systems. vol.~30 (2017),
  \url{https://proceedings.neurips.cc/paper/2017/file/892c3b1c6dccd52936e27cbd0ff683d6-Paper.pdf}

\bibitem{gutmann2010noise}
Gutmann, M., Hyv{\"a}rinen, A.: Noise-contrastive estimation: A new estimation
  principle for unnormalized statistical models. In: Proceedings of the
  thirteenth international conference on artificial intelligence and
  statistics. pp. 297--304. JMLR Workshop and Conference Proceedings (2010)

\bibitem{han2019divergence}
Han, T., Nijkamp, E., Fang, X., Hill, M., Zhu, S.C., Wu, Y.N.: Divergence
  triangle for joint training of generator model, energy-based model, and
  inferential model. In: Proceedings of the IEEE/CVF Conference on Computer
  Vision and Pattern Recognition. pp. 8670--8679 (2019)

\bibitem{han2020joint}
Han, T., Nijkamp, E., Zhou, L., Pang, B., Zhu, S.C., Wu, Y.N.: Joint training
  of variational auto-encoder and latent energy-based model. In: Proceedings of
  the IEEE/CVF Conference on Computer Vision and Pattern Recognition. pp.
  7978--7987 (2020)

\bibitem{he2016deep}
He, K., Zhang, X., Ren, S., Sun, J.: Deep residual learning for image
  recognition. In: Proceedings of the IEEE conference on computer vision and
  pattern recognition. pp. 770--778 (2016)

\bibitem{hein2019relu}
Hein, M., Andriushchenko, M., Bitterwolf, J.: Why relu networks yield
  high-confidence predictions far away from the training data and how to
  mitigate the problem. In: Proceedings of the IEEE Conference on Computer
  Vision and Pattern Recognition. pp. 41--50 (2019)

\bibitem{hendrycks2018deep}
Hendrycks, D., Mazeika, M., Dietterich, T.: Deep anomaly detection with outlier
  exposure. arXiv preprint arXiv:1812.04606  (2018)

\bibitem{heusel2017gans}
Heusel, M., Ramsauer, H., Unterthiner, T., Nessler, B., Hochreiter, S.: Gans
  trained by a two time-scale update rule converge to a local nash equilibrium.
  In: Advances in neural information processing systems. pp. 6626--6637 (2017)

\bibitem{ho2020denoising}
Ho, J., Jain, A., Abbeel, P.: Denoising diffusion probabilistic models. In:
  Advances in Neural Information Processing Systems. vol.~33 (2020),
  \url{https://proceedings.neurips.cc/paper/2020/file/4c5bcfec8584af0d967f1ab10179ca4b-Paper.pdf}

\bibitem{hyvarinen2005estimation}
Hyv{\"a}rinen, A., Dayan, P.: Estimation of non-normalized statistical models
  by score matching. Journal of Machine Learning Research  \textbf{6}(4) (2005)

\bibitem{ilyas2019adversarial}
Ilyas, A., Santurkar, S., Tsipras, D., Engstrom, L., Tran, B., Madry, A.:
  Adversarial examples are not bugs, they are features. arXiv preprint
  arXiv:1905.02175  (2019)

\bibitem{isola2017image}
Isola, P., Zhu, J.Y., Zhou, T., Efros, A.A.: Image-to-image translation with
  conditional adversarial networks. In: Proceedings of the IEEE conference on
  computer vision and pattern recognition. pp. 1125--1134 (2017)

\bibitem{jiang2020robust}
Jiang, Z., Chen, T., Chen, T., Wang, Z.: Robust pre-training by adversarial
  contrastive learning. Advances in Neural Information Processing Systems
  \textbf{33},  16199--16210 (2020)

\bibitem{karras2017progressive}
Karras, T., Aila, T., Laine, S., Lehtinen, J.: Progressive growing of gans for
  improved quality, stability, and variation. arXiv preprint arXiv:1710.10196
  (2017)

\bibitem{karras2020training}
Karras, T., Aittala, M., Hellsten, J., Laine, S., Lehtinen, J., Aila, T.:
  Training generative adversarial networks with limited data. In: Advances in
  Neural Information Processing Systems. vol.~33, pp. 12104--12114 (2020),
  \url{https://proceedings.neurips.cc/paper/2020/file/8d30aa96e72440759f74bd2306c1fa3d-Paper.pdf}

\bibitem{kingma2018glow}
Kingma, D.P., Dhariwal, P.: Glow: Generative flow with invertible 1x1
  convolutions. In: Advances in neural information processing systems. pp.
  10215--10224 (2018)

\bibitem{krizhevsky2009learning}
Krizhevsky, A., Hinton, G., et~al.: Learning multiple layers of features from
  tiny images  (2009)

\bibitem{kumar2019maximum}
Kumar, R., Ozair, S., Goyal, A., Courville, A., Bengio, Y.: Maximum entropy
  generators for energy-based models. arXiv preprint arXiv:1901.08508  (2019)

\bibitem{kurakin2016adversarial}
Kurakin, A., Goodfellow, I., Bengio, S.: Adversarial machine learning at scale.
  arXiv preprint arXiv:1611.01236  (2016)

\bibitem{lecun2006tutorial}
LeCun, Y., Chopra, S., Hadsell, R., Ranzato, M., Huang, F.: A tutorial on
  energy-based learning. Predicting structured data  \textbf{1}(0) (2006)

\bibitem{madry2017towards}
Madry, A., Makelov, A., Schmidt, L., Tsipras, D., Vladu, A.: Towards deep
  learning models resistant to adversarial attacks. arXiv preprint
  arXiv:1706.06083  (2017)

\bibitem{meinke2019towards}
Meinke, A., Hein, M.: Towards neural networks that provably know when they
  don't know. arXiv preprint arXiv:1909.12180  (2019)

\bibitem{mescheder2018training}
Mescheder, L., Geiger, A., Nowozin, S.: Which training methods for gans do
  actually converge? In: International conference on machine learning. pp.
  3481--3490. PMLR (2018)

\bibitem{miyato2018spectral}
Miyato, T., Kataoka, T., Koyama, M., Yoshida, Y.: Spectral normalization for
  generative adversarial networks. In: International Conference on Learning
  Representations (2018), \url{https://openreview.net/forum?id=B1QRgziT-}

\bibitem{nijkamp2022mcmc}
Nijkamp, E., Gao, R., Sountsov, P., Vasudevan, S., Pang, B., Zhu, S.C., Wu,
  Y.N.: {MCMC} should mix: Learning energy-based model with flow-based
  backbone. In: International Conference on Learning Representations (2022),
  \url{https://openreview.net/forum?id=4C93Qvn-tz}

\bibitem{nijkamp2020anatomy}
Nijkamp, E., Hill, M., Han, T., Zhu, S.C., Wu, Y.N.: On the anatomy of
  mcmc-based maximum likelihood learning of energy-based models. In:
  Proceedings of the AAAI Conference on Artificial Intelligence. vol.~34, pp.
  5272--5280 (2020)

\bibitem{nijkamp2019learning}
Nijkamp, E., Hill, M., Zhu, S.C., Wu, Y.N.: Learning non-convergent
  non-persistent short-run mcmc toward energy-based model. In: NeurIPS (2019)

\bibitem{pang2020learning}
Pang, B., Han, T., Nijkamp, E., Zhu, S.C., Wu, Y.N.: Learning latent space
  energy-based prior model. In: Advances in Neural Information Processing
  Systems. vol.~33 (2020),
  \url{https://proceedings.neurips.cc/paper/2020/file/fa3060edb66e6ff4507886f9912e1ab9-Paper.pdf}

\bibitem{pidhorskyi2020adversarial}
Pidhorskyi, S., Adjeroh, D.A., Doretto, G.: Adversarial latent autoencoders.
  In: Proceedings of the IEEE/CVF Conference on Computer Vision and Pattern
  Recognition. pp. 14104--14113 (2020)

\bibitem{ramachandran2017swish}
Ramachandran, P., Zoph, B., Le, Q.V.: Swish: a self-gated activation function.
  arXiv preprint arXiv:1710.05941  \textbf{7}, ~1 (2017)

\bibitem{rhodes2020telescoping}
Rhodes, B., Xu, K., Gutmann, M.U.: Telescoping density-ratio estimation. arXiv
  preprint arXiv:2006.12204  (2020)

\bibitem{salimans2016improved}
Salimans, T., Goodfellow, I., Zaremba, W., Cheung, V., Radford, A., Chen, X.:
  Improved techniques for training gans. Advances in neural information
  processing systems  \textbf{29},  2234--2242 (2016)

\bibitem{salimans2016weight}
Salimans, T., Kingma, D.P.: Weight normalization: A simple reparameterization
  to accelerate training of deep neural networks. Advances in neural
  information processing systems  \textbf{29},  901--909 (2016)

\bibitem{santurkar2019image}
Santurkar, S., Ilyas, A., Tsipras, D., Engstrom, L., Tran, B., Madry, A.: Image
  synthesis with a single (robust) classifier. In: Advances in Neural
  Information Processing Systems. pp. 1260--1271 (2019)

\bibitem{sehwag2019better}
Sehwag, V., Bhagoji, A.N., Song, L., Sitawarin, C., Cullina, D., Chiang, M.,
  Mittal, P.: Better the devil you know: An analysis of evasion attacks using
  out-of-distribution adversarial examples. arXiv preprint arXiv:1905.01726
  (2019)

\bibitem{song2019generative}
Song, Y., Ermon, S.: Generative modeling by estimating gradients of the data
  distribution. In: Advances in Neural Information Processing Systems. vol.~32
  (2019),
  \url{https://proceedings.neurips.cc/paper/2019/file/3001ef257407d5a371a96dcd947c7d93-Paper.pdf}

\bibitem{song2020improved}
Song, Y., Ermon, S.: Improved techniques for training score-based generative
  models. arXiv preprint arXiv:2006.09011  (2020)

\bibitem{song2021scorebased}
Song, Y., Sohl-Dickstein, J., Kingma, D.P., Kumar, A., Ermon, S., Poole, B.:
  Score-based generative modeling through stochastic differential equations.
  In: International Conference on Learning Representations (2021),
  \url{https://openreview.net/forum?id=PxTIG12RRHS}

\bibitem{torralba200880}
Torralba, A., Fergus, R., Freeman, W.T.: 80 million tiny images: A large data
  set for nonparametric object and scene recognition. IEEE transactions on
  pattern analysis and machine intelligence  \textbf{30}(11),  1958--1970
  (2008)

\bibitem{tsipras2018robustness}
Tsipras, D., Santurkar, S., Engstrom, L., Turner, A., Madry, A.: Robustness may
  be at odds with accuracy. arXiv preprint arXiv:1805.12152  (2018)

\bibitem{vahdat2020nvae}
Vahdat, A., Kautz, J.: Nvae: A deep hierarchical variational autoencoder. In:
  Advances in Neural Information Processing Systems. vol.~33 (2020),
  \url{https://proceedings.neurips.cc/paper/2020/file/e3b21256183cf7c2c7a66be163579d37-Paper.pdf}

\bibitem{wang2022a}
Wang, Y., Wang, Y., Yang, J., Lin, Z.: A unified contrastive energy-based model
  for understanding the generative ability of adversarial training. In:
  International Conference on Learning Representations (2022),
  \url{https://openreview.net/forum?id=XhF2VOMRHS}

\bibitem{welling2011bayesian}
Welling, M., Teh, Y.W.: Bayesian learning via stochastic gradient langevin
  dynamics. In: Proceedings of the 28th international conference on machine
  learning (ICML-11). pp. 681--688. Citeseer (2011)

\bibitem{xiao2021vaebm}
Xiao, Z., Kreis, K., Kautz, J., Vahdat, A.: Vaebm: A symbiosis between
  variational autoencoders and energy-based models. In: International
  Conference on Learning Representations (2021),
  \url{https://openreview.net/forum?id=5m3SEczOV8L}

\bibitem{xie2018cooperative}
Xie, J., Lu, Y., Gao, R., Zhu, S.C., Wu, Y.N.: Cooperative training of
  descriptor and generator networks. IEEE transactions on pattern analysis and
  machine intelligence  \textbf{42}(1),  27--45 (2018)

\bibitem{xie2016theory}
Xie, J., Lu, Y., Zhu, S.C., Wu, Y.: A theory of generative convnet. In:
  International Conference on Machine Learning. pp. 2635--2644. PMLR (2016)

\bibitem{xie2021cooperative}
Xie, J., Zheng, Z., Fang, X., Zhu, S.C., Wu, Y.N.: Cooperative training of fast
  thinking initializer and slow thinking solver for conditional learning. IEEE
  Transactions on Pattern Analysis and Machine Intelligence  (2021)

\bibitem{xie2018learning}
Xie, J., Zheng, Z., Gao, R., Wang, W., Zhu, S.C., Wu, Y.N.: Learning descriptor
  networks for 3d shape synthesis and analysis. In: Proceedings of the IEEE
  conference on computer vision and pattern recognition. pp. 8629--8638 (2018)

\bibitem{xie2020generative}
Xie, J., Zheng, Z., Gao, R., Wang, W., Zhu, S.C., Wu, Y.N.: Generative
  voxelnet: learning energy-based models for 3d shape synthesis and analysis.
  IEEE Transactions on Pattern Analysis and Machine Intelligence  (2020)

\bibitem{xie2021learning}
Xie, J., Zheng, Z., Li, P.: Learning energybased model with variational
  auto-encoder as amortized sampler. In: The Thirty-Fifth AAAI Conference on
  Artificial Intelligence (AAAI). vol.~2 (2021)

\bibitem{xie2022a}
Xie, J., Zhu, Y., Li, J., Li, P.: A tale of two flows: Cooperative learning of
  langevin flow and normalizing flow toward energy-based model. In:
  International Conference on Learning Representations (2022),
  \url{https://openreview.net/forum?id=31d5RLCUuXC}

\bibitem{Yin2020GAT}
Yin, X., Kolouri, S., Rohde, G.K.: Gat: Generative adversarial training for
  adversarial example detection and robust classification. In: International
  Conference on Learning Representations (2020),
  \url{https://openreview.net/forum?id=SJeQEp4YDH}

\bibitem{yu2015lsun}
Yu, F., Seff, A., Zhang, Y., Song, S., Funkhouser, T., Xiao, J.: Lsun:
  Construction of a large-scale image dataset using deep learning with humans
  in the loop. arXiv preprint arXiv:1506.03365  (2015)

\bibitem{zhao2021learning}
Zhao, Y., Xie, J., Li, P.: Learning energy-based generative models via
  coarse-to-fine expanding and sampling. In: International Conference on
  Learning Representations (2021),
  \url{https://openreview.net/forum?id=aD1_5zowqV}

\bibitem{zhu2017unpaired}
Zhu, J.Y., Park, T., Isola, P., Efros, A.A.: Unpaired image-to-image
  translation using cycle-consistent adversarial networks. In: Proceedings of
  the IEEE international conference on computer vision. pp. 2223--2232 (2017)

\bibitem{zhu2021towards}
Zhu, Y., Ma, J., Sun, J., Chen, Z., Jiang, R., Chen, Y., Li, Z.: Towards
  understanding the generative capability of adversarially robust classifiers.
  In: Proceedings of the IEEE/CVF International Conference on Computer Vision.
  pp. 7728--7737 (2021)

\end{thebibliography}

\ifdefined\APPENDIX
\clearpage
%\appendix
\title{Supplementary Materials}
\author{}
\institute{}

\maketitle

\section{Proof of Proposition~\eqref{pro:optimalD}}
\begin{proposition}
	\label{pro:optimalD}
	The optimal solution of $\max_{D}\min_{p_T} U(D, p_T)$ is $U(D^*, p_T^*)=-\log(4)$, where $D^*$ outputs $\frac{1}{2}$ on $\supp {p_\textrm{data}}$ and $\leq \frac{1}{2}$ outside $\supp{p_\textrm{data}}$, and $p_T^*$ is supported in the contour set  $\{D=\frac{1}{2}\}$.
\end{proposition}
\begin{proof}
	Let 
	\begin{equation}
		p_T^* = \arg\min_{p_T}\mathbb{E}_{\mathrm{x} \sim p_T} [ \log(1-{D}(x)) ],
	\end{equation}
	then
	\begin{equation}
		\max_{{D}}\min_{p_T} U({D}, p_T) = \max_{{D}} U({D}, p_T^*).
	\end{equation}
	We solve $\max_{{D}} U({D}, p_T^*)$ by first deriving its upper bound. 
	Let  $\alpha=\max_{\mathcal{X}}D$, 
	then  $\mathbb{E}_{\mathrm{x} \sim p_T^*} [ \log(1-{D}(x)) ]$ is minimized when $p_T^*$ is supported in $\{x: D(x)=\alpha\}$
	With this result, we can derive an upper bound of $ U({D}, p_T^*)$:
	%     \begin{equation}
		%         \begin{aligned}
			% 		&U({D}, p_T^*) \\
			% 		&= \int_{\mathcal{X}} p_\mathrm{data}(x)\log {D}({x})dx +  \int_{\mathcal{X}}p_T^*(x)  \log(1-{D}(x))dx \\
			% 		                     &= \int_{\mathcal{X}} p_\mathrm{data}(x)\log {D}({x})dx +  \int_{\mathcal{X}}p_T^*(x)  \log(1-\alpha)dx \\
			% 		                     &\le \int_{\mathcal{X}} p_\mathrm{data}(x)\log (\alpha) dx + \int_{\mathcal{X}}p_T^*(x)  \log(1-\alpha)dx \\
			% 		                     &= \log(\alpha) + \log(1-\alpha) \\
			% 		                     & \le -\log(4),
			%         \end{aligned}
		%     \end{equation}
	\begin{align}
		&U({D}, p_T^*)  \nonumber\\
		&= \int_{\mathcal{X}} p_\mathrm{data}(x)\log {D}({x})dx +  \int_{\mathcal{X}}p_T^*(x)  \log(1-{D}(x))dx \nonumber\\
		&= \int_{\mathcal{X}} p_\mathrm{data}(x)\log {D}({x})dx +  \int_{\mathcal{X}}p_T^*(x)  \log(1-\alpha)dx  \nonumber\\
		&\le \int_{\mathcal{X}} p_\mathrm{data}(x)\log (\alpha) dx + \int_{\mathcal{X}}p_T^*(x)  \log(1-\alpha)dx  \nonumber\\
		&= \log(\alpha) + \log(1-\alpha)  \nonumber\\
		& \le -\log(4),
	\end{align}
	where the last inequality follows from the fact that the function $f(\alpha)= \log (\alpha) +  \log(1-\alpha)$ achieves its maximum value of $-\log(4)$ at $\alpha = \frac{1}{2}$.
	It is not hard to see that equality holds if and only if i) $\max_{\mathcal{X}}D=\frac{1}{2}$, ii) $D = \frac{1}{2}$ on $\supp {p_\mathrm{data}}$, and iii) $\supp{p_T^*}\subseteq \{x: D(x)=\frac{1}{2}\}$. In summary,  $\max_{{D}}\min_{p_T} U({D}, p_T)$ achieves its optimal value of  $-\log (4)$ at $(D^*, p_T^*)$ where 
	\begin{equation}
		D^*(x) =\begin{cases}
			\frac{1}{2} \quad x\in \supp {p_\mathrm{data}}\\
			\leq \frac{1}{2} \quad x\in \mathcal{X}\setminus \supp {p_\mathrm{data}}
		\end{cases},
	\end{equation}
	and $p_T^*$ is supported in the contour set  $\{D=\frac{1}{2}\}$.
\end{proof}

\section{Connection to GANs}
%\subsection{Game theory background}
% Overleaf project icml2021 appendix.tex
In this section we provide a comparative analysis of the proposed AT generative model and GANs~\cite{goodfellow2014generative}.
The proposed approach learns data distribution by solving the maximin problem
\begin{equation}
	\label{eq:maxmin-problem}
	\max_{{D}}\min_{p_T} U({D}, p_T) = \mathbb{E}_{\mathrm{x} \sim p_\mathrm{data}}[\log {D}({x})] +  \mathbb{E}_{\mathrm{x} \sim p_T} [ \log(1-{D}(x)) ],
\end{equation}
while GANs learn a generator function $G$ by solving the minimax problem
\begin{equation}
	\label{eq:gan-minmax}
	\min_{G} \max_D V(D, G) = \mathbb{E}_{\mathrm{x} \sim p_{\text{data}}}[\log D({x})] + \mathbb{E}_{\mathrm{z} \sim p_{{z}}}[\log (1 - D(G({z})))].
\end{equation}
The generator $G$ implicitly defines a distribution $p_{g}$  by mapping a prior distribution $p_z$ from a low-dimensional latent space $\mathcal{Z}\subseteq \mathbb{R}^z$ to the high-dimensional data space $\mathcal{X}\subseteq \mathbb{R}^d$.
%$D: \mathcal{X}\rightarrow [0,1]$ is a function that distinguish the target distribution $p_{\text{data}}$ samples  from $p_g$ samples. It can be shown that in the optimal solution to this minimax game, $p_g=p_\text{data}$.
Plugging $p_g$ into \cref{eq:gan-minmax}, we get:
\begin{equation}
	\label{eq:gan-minmax-pg}
	\min_{p_g}\max_{D} U({D}, p_g) = \mathbb{E}_{\mathrm{x} \sim p_\mathrm{data}}[\log {D}({x})] +  \mathbb{E}_{\mathrm{x} \sim p_g} [ \log(1-{D}(x)) ]
\end{equation}
Comparing \cref{eq:maxmin-problem} with \cref{eq:gan-minmax-pg} we find both problems making use of the standard log-likelihood objective for binary classification, but have a reversed order of minimization and maximization. 
In fact, both formulations solve a two-player zero-sum game, a mathematical representation of a situation in which one player's gain is balanced by another player's loss. This game can be described by the \textit{payoff function} $f:\mathbb{R}^{p+q}\rightarrow \mathbb{R}$, which represents the amount of payment that one player (player 1) makes to the other player (player 2). The goal of player 1 is to choose a strategy $u\in\mathbb{R}^p$ such that the payoff is minimized, while the goal of player 2 is to choose a strategy $u\in\mathbb{R}^q$ such that the payoff is maximized.  Depending on the order of maximization and minimization, the best strategies for both players, and the optimal payoff, can be solved via $\min_{u}\max_{v}f(u,v)$ or $\max_{v}\min_u f(u, v)$.

%In the minimax game $\min_{u}\max_{v}$, player 1 makes the first move. Player 2, after learning that player 1 has made the move $u$, will choose a $v$ to maximize $f(u,v)$, which results in a payoff of $\max_v f(u,v)$.
%Player 1, who is informed of player 2's strategy, will choose a $u$ such that the worse case payoff $\max_v f(u,v)$  is minimized, which results in a payoff of $\min_{u}\max_{v}f(u,v)$.

%In the maximin game $\max_{v}\min_u$, the order of play is reversed. Player 2 makes the first move, and then player 1 minimizes the payoff by choosing $u=\arg\min_u f(u, v)$.  Player 2 knows that player 1 will follow this strategy and will choose a $v$ such that the worse case payoff $\min_u f(u,v)$ is maximized, which results in a payoff of $\max_{v}\min_u f(u, v)$.

%The payoff $\min_{u}\max_{v}f(u,v)$ is always greater or equal to $\max_{v}\min_u f(u, v)$. 
%This difference can be intuitively understood as the result of player 2's extra knowledge gained by taking the second move.
%According to the minimax theorem~\cite{neumann1928theorie}, when $f$  is a continuous function that is concave-convex (i.e., for each $v$, $f(u,v)$ is a convex function of $u$, and for each $u$, $f(u,v)$ is a concave function of $v$)), these two quantities are equal. We refer the reader to \cite{boyd2004convex} (\S 5.4.3, \S 10.3.4) for more details on this topic.

%\subsubsection{Comparative analysis with GANs}
In \cref{eq:maxmin-problem}, $U (D, p_T)$ is the {payoff function}, and
the goal of player  $p_T$  is to choose a strategy $p_T^*$ such that the payoff
is minimized, whereas the goal of player $D$ is to choose a strategy $D^*$ such that the payoff is maximized.
This \textit{maximin} game is played by following such a rule: player $D$ makes the first move by choosing a $D$; player $p_T$, after learning that player $D$ has made the move, will choose a $p_T$ to minimize its payment, which results in a payoff of $\min_{p_T}U(D, p_T)$; player $D$, who is informed of player $p_T$'s strategy, will chooses a $D$ such that the worse case payoff $\min_{p_T}U(D, p_T)$ is maximized, which results in an overall payoff of $\max_{D}\min_{p_T}U(D, p_T)$.
The best strategies of both players and the maximum payoff can be derive from \cref{pro:optimalD} : 	In the maximin game $\max_{D}\min_{p_T} U(D, p_T)$, the best strategy for player $D$ is to choose a $D^*$ that outputs $\frac{1}{2}$ on $\supp {p_\textrm{data}}$ and $\leq \frac{1}{2}$ outside $ \supp{p_\textrm{data}}$, the best strategy for player $p_T$ is to choose a $p_T^*$ which is supported in  $\{x:D(x)=\frac{1}{2}\}$, and the maximum payoff is $-\log(4)$.

In \cref{eq:gan-minmax-pg}, $U(D, p_g)$ is the payoff function. Similar  to \cref{eq:maxmin-problem}, the goal of player  $p_g$  is to minimize the payoff, and the goal of player $D$ is to maximize the payoff. In contrast to \cref{eq:maxmin-problem}, player $p_g$ makes the first move. 
%Compared to the maximin game, the minimax game $\min_{p_T}\max_D U(D, p_T)$ has a reversed rule: player $p_T$ makes the first move by choosing a $p_T$; player $D$ then chooses a $D$ to maximize the payoff, which results in a payoff of $\max_D U(D, p_T)$; player $p_T$ knows player $D$'s strategy and will choose a $p_T$ such that the worst case payoff $\max_D U(D, p_T)$ is minimized, which results in a overall payoff of $\min_{p_T}\max_D U(D, p_T)$.
The solution to this minimax game is analyzed in~\cite{goodfellow2014generative}: 
the best strategy of player $p_g$ is to choose a  $p_g^*$ which minimizes the Jensen-Shannon divergence (JSD) between $p_g$ and $p_\textrm{data}$: $p_g^*=\arg\min_{p_g} {\rm {JSD}}(p_g\parallel p_\textrm{data})=p_\textrm{data}$,
and the best strategy of player $D$ is to choose ${D}^* (x) = \frac{p_\textrm{data} (x)}{p_\textrm{data}(x)+p_g^*(x)}=\frac{1}{2}$. Under these strategies, the payoff function $U$ measures the JSD between $p_g$ and $p_\textrm{data}$: $U({D}^*, p_g^*) = -\log(4)+2\cdot {\rm {JSD}}(p_g^*\parallel p_\textrm{data})=-\log(4)$, which coincides with the $U$ solution in the maximin game. Note that in the minimax game, $D^*$ does not need to be defined outside $\supp{p_g} \cup \supp{p_\textrm{data}}$~\cite{goodfellow2014generative}.

The optimal solutions to these two formulations are summarized in \cref{tab:diff}.

The pseudo code for solving the minimax problem is outlined in \cref{alg:gat-minimax}.  \cref{fig:minimax2D} shows the simulation results in two settings where $p_0$ data is respectively uniformly distributed (left panel) and concentrated in the lower left corner (right panel). In can be seen that in both cases $p_T^*$ matches $p_\textrm{data}$ when the algorithm converges. 
%The $D$ solution is also affected by the setting of the $p_0$ data.
The right panel shows that when $p_0$ data is concentrated in the lower left corner, the $D$ solution has undefined outputs outside  $\supp {p_\textrm{data}}$.

We find these two formulations giving rise to different applications. The minimax formulation is ideal for learning a generator model that can produce a distribution that matches $p_\textrm{data}$. The discriminator, because of its undefined behavior outside $\supp {p_\textrm{data}}$, may not be very useful for some downstream tasks such as out-of-distribution detection.
In the maximin formulation, as we have discussed in the main text, can be used for sample generation, image-to-image translation, image restoration such as denoising and inpainting, and (worst-case) out-of-distribution detection.

\setlength{\tabcolsep}{4pt}
\begin{table}[h!]
	\begin{center}
		\caption{Optimal solutions to the minimax problem and maximin problem}
		\label{tab:diff}
		\resizebox{\textwidth}{!}{%
			\begin{tabular}{@{}lll@{}}
				\toprule
				& Minimax (GANs)~\cref{eq:gan-minmax-pg}                                                                                                           & Maximin (ours)~\cref{eq:maxmin-problem}                                                                                                                      \\ \midrule
				$p_T^*$/$p_g^*$                 & $p_g^* = p_\textrm{data}$                                                                                                & $p_T^*$ is supported in $\{x:D(x)=\frac{1}{2}\}$                                                                                       \\ \midrule
				$D^*$                           & \begin{tabular}[c]{@{}l@{}}$D^*(x)=\frac{1}{2}$ on $\supp {p_\textrm{data}}$, \\ undefined outside $\supp {p_\textrm{data}}$\end{tabular} & \begin{tabular}[c]{@{}l@{}}$D^*(x)=\frac{1}{2}$ on $\supp {p_\textrm{data}}$, \\ $D^*(x)\le \frac{1}{2}$ outside $\supp {p_\textrm{data}}$\end{tabular} \\ \midrule
				$U(D^*, p_g^*)$/$U(D^*, p_T^*)$ & $-\log(4)$                                                                                                               & $-\log(4)$                                                                                                                             \\ \bottomrule
			\end{tabular}%
		}
	\end{center}	
	
\end{table}
\setlength{\tabcolsep}{1.4pt}

\begin{algorithm}[h]
	\caption{\small Solving the minimax problem}
	\label{alg:gat-minimax}
	%	\scalebox{0.95}{%
		{\small
			\begin{algorithmic}[1]
				\STATE Draw samples $\{ {x}_i \}_{i=1}^m$  from $p_\mathrm{data}$, and samples $\{ {x}_i^*\}_{i=1}^m$  from $p_0$.
				\REPEAT
				\STATE Update $D$ by maximizing  $\frac{1}{m} \sum_{i=1}^m \log {D}({x_i}) + \frac{1}{m} \sum_{i=1}^m \log (1 - D({x_i^*}))$ (until converge).
				\STATE For each  $x\in \{ {x}_i^*\}_{i=1}^m$, update its value by \\$x\leftarrow x + \lambda \frac{\nabla D(x)}{\|\nabla D(x)\|_2}$ (single step).
				\UNTIL {$\{ {x}_i^*\}_{i=1}^m = \{ {x}_i \}_{i=1}^m$}
			\end{algorithmic}
		}
		%	}%
\end{algorithm}

%
%\begin{figure}[ht]
%	\centering
%	\includegraphics[width=0.3\columnwidth]{figs/appendix/minimax2d_uniform_init.pdf}
%	\includegraphics[width=0.3\columnwidth]{figs/appendix/minimax2d_uniform_end.pdf}
%	% \noindent\rule{\columnwidth}{0.5pt}
%	\vspace{6pt}
%	\hrule
%	\vspace{6pt}
%	\includegraphics[width=0.3\columnwidth]{figs/appendix/minimax2d_concentrated_init.pdf}
%	\includegraphics[width=0.3\columnwidth]{figs/appendix/minimax2d_concentrated_end.pdf}
%	\caption{Plots of contours and (normalized) gradient vector fields of the $D$ functions learned with \cref{alg:gat-minimax}. Left images: $p_0$ data (red points) and initial states of the $D$ functions. Right images: $p_T^*$ data (red points) and the final states of the $D$ functions when the algorithm converges.}
%	\label{fig:minimax2D}
%\end{figure}

\begin{figure}[h!]
	\centering
	\scriptsize
	\begin{minipage}{.26\textwidth}
		\centering
		\includegraphics[width=\linewidth]{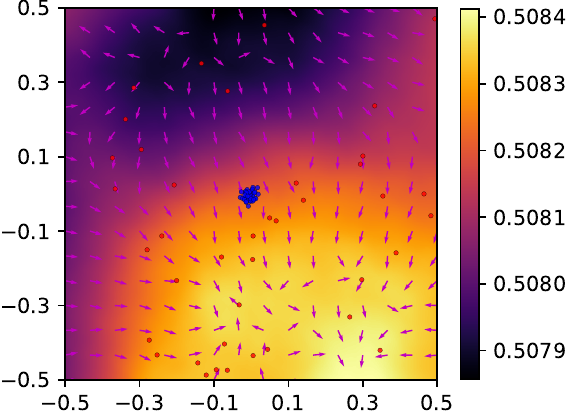}
		\\ (1a)
		%		\caption*{jfkdsf}
	\end{minipage}%
	\begin{minipage}{0.24\textwidth}
		\centering
		\includegraphics[width=\linewidth]{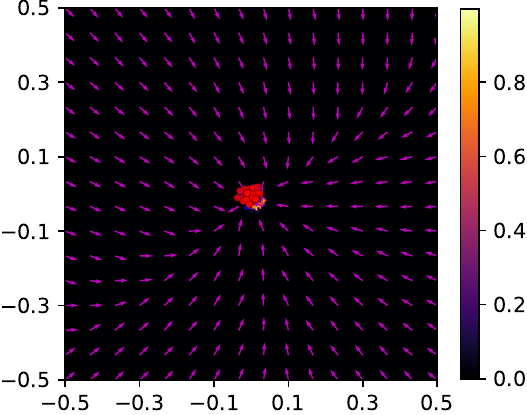}
		\\ (1b)
	\end{minipage}
	\vrule
	\hfill
	\begin{minipage}{.26\textwidth}
		\centering
		\includegraphics[width=\linewidth]{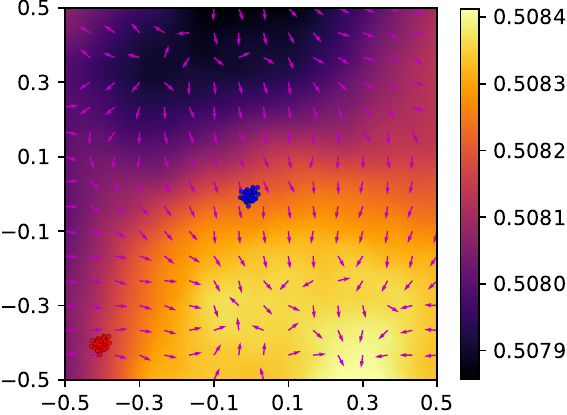}
		\\ (2a)
	\end{minipage}%
	\begin{minipage}{0.23\textwidth}
		\centering
		\includegraphics[width=\linewidth]{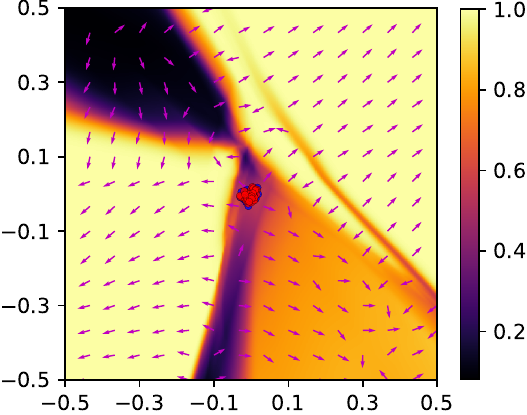}
		\\ (2b)
	\end{minipage}
	\caption{Plots of contours and (normalized) gradient vector fields of the $D$ functions learned with different $p_0$ data. Left and right panel respectively show the initial state (1a and 2a) and final state (1b and 2b) of $D$ when $p_0$ data is respectively uniformly distributed (red points in 1a) and concerntrated in the lower left corner (red points in 2a). $p_\textrm{data}$ is a Gaussian distribution centered at $(0,0)$ (blue points). }
	\label{fig:minimax2D}
\end{figure}

\newpage
\section{Experimental Setups}
\subsubsection{Model Architecture.}
On CIFAR-10 we  use the standard ResNet50~\cite{he2016deep} architecture with ReLU activation for the $D$ model. On CelebA-HQ 256, AFHQ-CAT 256, and LSUN-Church 256 we use a customized architecture (\cref{tab:model-arch}) adapted from~\cite{choi2020stargan}.
\setlength{\tabcolsep}{4pt}
\begin{table}[h!]
	\begin{center}
		\caption{Network architecture for the $D$ model used in CelebA-HQ 256, AFHQ-CAT 256, and LSUN-Church 256.}
		\label{tab:model-arch}
		\resizebox{0.44\textwidth}{!}{%
			\begin{tabular}{@{}lll@{}}
				\toprule
				Layer           & Resample & Output shape              \\ \midrule
				Conv$1\times 1$ & -        & $256\times 256\times 64$  \\
				ResBlock        & AvgPool  & $128\times 128\times 128$ \\
				ResBlock        & AvgPool  & $64\times 64\times 256$   \\
				ResBlock        & AvgPool  & $32\times 32\times 512$   \\
				ResBlock        & AvgPool  & $16\times 16\times 512$   \\
				ResBlock        & AvgPool  & $8\times 8\times 512$     \\
				ResBlock        & AvgPool  & $4\times 4\times 512$     \\
				LeakyReLU       & -        & $4\times 4\times 512$     \\
				Conv$4\times 4$ & -        & $1\times 1\times 512$     \\
				LeakyReLU       & -        & $1\times 1\times 512$     \\
				Reshape         & -        & $512$                       \\
				Linear          & -        & $1$                        \\ \bottomrule
			\end{tabular}%
		}
	\end{center}	
	
\end{table}
\setlength{\tabcolsep}{1.4pt}

\subsubsection{Datasets.}
We evaluate our method on CIFAR-10~\cite{krizhevsky2009learning} (50K training samples), CelebA-HQ 256~\cite{karras2017progressive} (30K training samples), AFHQ-CAT~\cite{choi2020stargan} dataset (5153 training samples), and LSUN-Church~\cite{yu2015lsun} (126227 training samples). AFHQ~\cite{choi2020stargan} is a recently introduced benchmark dataset for image-to-image translation. 
% We include this dataset mainly for the purpose of demonstrating our method's application to image-to-image translation (translation between human faces and and cat faces). 

\subsubsection{Evaluation Metrics.}
We use Inception Score (IS)~\cite{salimans2016improved} and FID score~\cite{heusel2017gans}  to evaluate the quality of generated samples. We follow \cite{karras2020training} and compute the FID score between 50k generated samples and all training samples (IS is also calculated on the generated 50K samples). 
We use the original code from \cite{salimans2016improved} and \cite{heusel2017gans} to calculate the scores. For OOD detection, we use area under the ROC curve (AUROC) as the evaluation metric. 
% Use proper official implementation of FID and IS \url{https://github.com/bioinf-jku/TTUR}, \url{https://github.com/openai/improved-gan/tree/master/inception_score}

\subsubsection{Training.} We use Algorithm 2 to train the models. The training hyperparameters for each task can be found in \cref{tab:training-hyperparameters}. For the $256 \times 256$ tasks, we pretrain the $D$ model on the ImageNet classification task. To mitigate overfitting, we perform random resized cropping, random horizontal flipping on $p_\textrm{data}$ samples.
The performance (FID score) of the model is monitored during training and the best-performing model to used to report the final FID score.

The CIFAR-10 worst-case OOD detection model is trained using in- and out-distribution adversarial training~\cite{augustin2020adversarial}, where in-distribution AT uses a $l^2$-ball of radius 0.25 and PGD attacks of steps 10 and step-size 0.1, and out-distribution AT uses a $l^2$-ball of radius 0.5 and PGD attacks of steps 10 and step-size 0.1. Following ~\cite{augustin2020adversarial}, we use a batch size of 128 and use the recommended AutoAugment policy from~\cite{cubuk2019autoaugment}. The model is trained for 400 epochs using a SGD optimizer with a fixed learning rate of 0.1.
% Following Augustin~\etal~\cite{augustin2020adversarial}, we use a batch size of 128 and perform early stopping based on model performance on the test set.
%A separate  validation set split from the 80 million tiny images is used for selecting the model for final evaluation. 
\setlength{\tabcolsep}{4pt}
\begin{table}[h!]
	\begin{center}
		\caption{Training hyperparameters. We use $\beta_1=0.0, \beta_2=0.99$  for the Adam optimizer. 
		}
		\label{tab:training-hyperparameters}
		\begin{tabular}{@{}lllll@{}}
			\toprule
			& CIFAR-10 & CelebA-HQ 256 & AFHQ-CAT 256 & LSUN-Church 256 \\ \midrule
			Batch size             & 32       & 40            & 40           & 32              \\
			Training iterations    & 172K     & 218K          & 225K         & 215K            \\
			Optimizer              & Adam      & Adam          & Adam         & Adam            \\
			Learning rate          & 5e-4     & 5e-5          & 5e-5         & 5e-5            \\
			$K$ & 0,...,25       & 0,...,40            & 0,...,25           & 0,...,35         \\
			Epochs per $K$ & 5 & 5 & 50 & 1\\
			PGD attack step-size   & 0.1      & 2.0           & 2.0          & 2.0             \\
			$R_1$ regularization   & 0.01     & 30            & 100          & 100             \\ \bottomrule
		\end{tabular}
	\end{center}
\end{table}
\setlength{\tabcolsep}{1.4pt}

%\begin{table}[h!]
%	\begin{center}
	%	\caption{The training schedule for each task}
	%	\label{tab:training-schedule}
	%
	%\begin{tabular}{@{}ll@{}}
	%	\toprule
	%	Task            & Training schedule                                                                                                                      \\ \midrule
	%	CIFAR-10        & \begin{tabular}[c]{@{}l@{}}$K = 0, 1,\dots, 25$; \\ train 5 epochs for each $K$, and then continue training  with $K=25$\end{tabular}  \\ \midrule
	%	CelebA-HQ 256   & \begin{tabular}[c]{@{}l@{}}$K = 0, 1,\dots, 35$; \\ train 5 epochs for each $K$, and then continue training  with $K=35$\end{tabular}  \\ \midrule
	%	AFHQ-CAT 256    & \begin{tabular}[c]{@{}l@{}}$K = 0, 1,\dots, 25$; \\ train 50 epochs for each $K$, and then continue training  with $K=25$\end{tabular} \\ \midrule
	%	LSUN-Church 256 & \begin{tabular}[c]{@{}l@{}}$K = 0, 1,\dots, 35$; \\ train 1 epoch for each $K$, and then continue training  with $K=35$\end{tabular}   \\ \bottomrule
	%\end{tabular}
	%	\end{center}
%\end{table}

\noindent\textbf{Sample Generation.}
The generated samples for FID and IS evaluation are produced by performing PGD attacks on 50K  samples  randomly drawn  from the $p_0$ dataset. The settings for the $p_0$ dataset and the PGD attack can be found in Table~\ref{tab:gen-setting}.

\setlength{\tabcolsep}{4pt}
\begin{table}[h!]
	\begin{center}
		\caption{Sample generation setting}
		\label{tab:gen-setting}
		
		\begin{tabular}{@{}llll@{}}
			\toprule
			Task          & $p_0$ dataset                                & PGD step size & PGD steps \\ \midrule
			CIFAR-10      & 80 million tiny images \cite{torralba200880} & 0.2           & 32        \\
			CelebA-HQ 256 & ImageNet \cite{deng2009imagenet}             & 8.0           & 20        \\
			AFHQ-CAT 256  & ImageNet \cite{deng2009imagenet}             & 8.0           & 14        \\ 
			LSUN-Church 256  & ImageNet \cite{deng2009imagenet}             & 8.0           & 17        \\ 
			\bottomrule
		\end{tabular}
	\end{center}
\end{table}
\setlength{\tabcolsep}{1.4pt}

%\newpage
\section{Extended Experiment Results}
\subsection{Training and Test Time Sampling Efficiency}
\cref{tab:steps} shows that our method has competitive training and test time sampling efficiency to state-of-the-art EBMs. Although VAEBM typically requires much fewer update steps than our method, its per-step efficiency is much worse (\cref{tab:generation-speed}), suggesting that its VAE component has considerable computational complexity. We also observe that the quality of our generated samples is not sensitive to the number of sampling steps as long as the overall perturbation (\#step $\times$ step-size) remains the same (\cref{tab:fid_change}). This allows us to use a much larger step size than the one used during training to speedup test time sampling in real applications. 
% While in EBMs the FID score affected by the number of steps~\cite{nijkamp2019learning}

\setlength{\tabcolsep}{4pt}
\begin{table}[h!]
	\begin{center}
		\caption{The number of update steps in the PGD attack (our method) and Langevin dynamics (other methods). ``PCD'' refers to using a persistent sampling chain.}
		\label{tab:steps}
		\begin{tabular}{@{}lllll@{}}
			\toprule
			& Ours  & VAEBM~\cite{xiao2021vaebm}        & CF-EBM~\cite{zhao2021learning} & JEM~\cite{Grathwohl2020Your}           \\ \midrule
			CIFAR-10 (train)            & 25 & 6 (PCD)  & 50  & 20 (PCD)  \\
			CIFAR-10 (test)            &  32 &  16 & 50  &  100 \\
			CelebA-HQ 256 (train)        & 40 & 6 (PCD)  & 90   & N/A             \\
			CelebA-HQ 256 (test)        &  20 &  24 &  90  & N/A             \\ \bottomrule
		\end{tabular}%
	\end{center}
\end{table}
\setlength{\tabcolsep}{1.4pt}

\setlength{\tabcolsep}{4pt}
\begin{table}[h!]
	\begin{center}
		\caption{Number of steps and wall-clock time to generate 50 CIFAR-10 samples.  Data of NCSN and VAEBM are from \cite{xiao2021vaebm}.}
		\label{tab:generation-speed}
		%		\resizebox{0.5\textwidth}{!}{%
			\begin{tabular}{@{}llll@{}}
				\toprule
				Model & Steps & Wall-clock time & GPU device \\ \midrule
				NCSN~\cite{song2019generative} & 1000 & 107.9 seconds & RTX Titan \\
				VAEBM~\cite{xiao2021vaebm} & 16 & 8.79 seconds &  RTX Titan\\
				Ours & 32 & 2.34 seconds & RTX 2080 Ti \\ \bottomrule
			\end{tabular}%
			%		}
	\end{center}
	
	% (16.31 TFLOPS in FP32),  (13.45 TFLOPS in FP32)
\end{table}
\setlength{\tabcolsep}{1.4pt}

\setlength{\tabcolsep}{4pt}
\begin{table}[h!]
	\begin{center}
		\caption{FID scores of samples generated using different combinations of number of steps and step-size.}
		\label{tab:fid_change}
		%		\resizebox{0.7\textwidth}{!}{%
			%		\begin{tabular}{@{}llll@{}}
				%			\toprule
				%			CIFAR-10      & 13.07 / $64\times 0.1$ & 13.21 / $32\times 0.2$ & 13.49 / $16\times 0.4$ \\ \midrule
				%			CelebA-HQ 256 & 19.19 / $40\times 4.0$       & 18.97 / $20\times 8.0$ &  19.19 / $10\times 16.0$    \\ \bottomrule
				%		\end{tabular}% 
			\begin{tabular}{@{}lll@{}}
				\toprule
				\multicolumn{1}{l}{}           & Number of steps $\times$ step-size & FID   \\ \midrule
				\multirow{3}{*}{CIFAR-10}      & $ 64\times0.1 $                 & 13.07 \\
				& $ 32\times0.2 $                 & 13.21 \\
				& $ 16\times0.4  $                & 13.49 \\ \midrule
				\multirow{3}{*}{CelebA-HQ 256} & $ 40\times4.0 $                 & 19.19 \\
				& $ 20\times8.0 $                 & 18.97 \\
				& $ 10\times16.0 $                & 19.19 \\ \bottomrule
			\end{tabular}%
			%	}
	\end{center}
\end{table}
\setlength{\tabcolsep}{1.4pt}

%\newpage
\subsection{Extend Results on Worst-Case Out-Of-Distribution Detection}
\cref{tab:ood256} shows that under a PGD adversary with $l^2$ radius 7.0 our model exhibits strong out-distribution robustness. (Note that according to \cite{augustin2020adversarial}, a perturbation of $7.0$ is already large enough to make undefended models (e.g., OE~\cite{hendrycks2018deep}) fail completely at the OOD detection task).  
%When we further increase the perturbation limit to 100, the AUC scores decrease to near 0, suggesting that obfuscated gradients did not occur.

\setlength{\tabcolsep}{4pt}
\begin{table}[h!]
	\begin{center}
		\caption{OOD detection results on $256\times 256$ datasets. Each entry shows the AUC score on clean OOD samples (left value) and AUC score on adversarial OOD samples (right value).
			Adversarial OOD samples are computed by  maximizing the model output in a $l^2$-ball of radius 7.0  around OOD samples via Auto-PGD~\cite{Croce2020ReliableEO} with 100 steps and 5 random restarts. Results are computed using 1024 in-distribution samples and 1024 out-distribution samples.
		}
		\label{tab:ood256}                           
		
		%\resizebox{\textwidth}{!}{%
			\begin{tabular}{@{}lccc@{}}
				\toprule
				\multirow{2}{*}{OOD dataset} & \multicolumn{3}{c}{In-distribution dataset}          \\ \cmidrule(l){2-4} 
				& CelebA-HQ 256    & AFHQ-CAT 256    & LSUN-Church 256 \\ \midrule
				\multicolumn{4}{l}{Standard OOD detection}                                          \\ \midrule
				Uniform noise                & 1.0              & 1.0             & 0.9476          \\
				SVHN                         & 0.9967           & 0.9944          & 0.9668          \\
				CIFAR-10                     & 0.9978           & 0.9930          & 0.9081          \\
				ImageNet validation set      & 0.9986           & 0.9971          & 0.9409          \\
				AFHQ-CAT 256                 & 0.9984           & N/A             & 0.9691          \\
				CelebA-HQ 256                & N/A              & 0.9900          & 0.9794          \\
				LSUN-Church 256                & 0.9999           & 0.9997          & N/A             \\ \midrule
				\multicolumn{4}{l}{Worst-case OOD detection}                                        \\ \midrule
				Uniform noise                & 1.0              & 1.0             & 0.9330          \\
				SVHN                         & 0.9928           & 0.9880          & 0.9566          \\
				CIFAR-10                     & 0.9952           & 0.9859          & 0.8857          \\
				ImageNet validation set      & 0.9973           & 0.9937          & 0.9270          \\
				AFHQ-CAT 256                 & 0.9958           & N/A             & 0.9587          \\
				CelebA-HQ 256                & N/A              & 0.9773          & 0.9714          \\
				LSUN-Church 256                &   0.9998 &  0.9991 & N/A             \\ \bottomrule
			\end{tabular}%
			%}
	\end{center}
\end{table}
\setlength{\tabcolsep}{1.4pt}

%\newpage
\subsection{Extended Results on Generation}
Additional results are summarized below:
\begin{itemize}
	
	\item \textbf{Uncurated generation samples.}
	\cref{fig:cifar10-samples},
	\cref{fig:celebahq-samples-uncurated}, \cref{fig:cat-samples-uncurated}, and \cref{fig:church-samples-uncurated} show the uncurated generated samples on CIFAR-10, CelebA-HQ 256, AFHQ-CAT 256, and LSUN-Church 256. Note that we have used the same seed images (\cref{fig:seed}) to generated these results. We find that some generated images contain artifacts. By first applying Gaussian smoothing ($\sigma=10$) to the source  images ($p_0$ data), 
	we are able to obtain more visually pleasing results (\cref{fig:celebahq-samples-uncurated-blur}). The generated samples contain less artifacts,  but have a slightly worse FID. 
	% The smoothing seems to be playing a similar role as the ``truncation trick'' used by other generative models to generate better-looking results (with reduced diversity) by using noise initialized in the high density area of the latent space. 
	The smoothing filters out high frequency components, and
	seems to be playing a similar role as reduced-temperature sampling~\cite{vahdat2020nvae,xiao2021vaebm} and the ``truncation trick''~\cite{brock2018large}, where better-looking results (typically with reduced diversity) can be generated from latent noise sampled from the high density area of the latent space.

	\item \textbf{Nearest Neighbor Analysis.}
	\cref{fig:cifar10-nn}, 
	\cref{fig:celebahq-nn}, \cref{fig:cat-nn}, and \cref{fig:church-nn}  show the pixel space and inception feature space nearest neighbors of the generated samples on CIFAR-10, CelebA-HQ 256, AFHQ-CAT 256, and LSUN-Church 256. Note that none of the nearest neighbors resemble the generated samples, suggesting that the models have not memorized the training data.

	\item \textbf{Interpolation.}
	\cref{fig:celebahq-interpolation}, \cref{fig:cat-interpolation}, and \cref{fig:church-interpolation} show the interpolation results on CelebA-HQ 256, AFHQ-CAT 256, and LSUN-Church 256. The interpolation works reasonable well even on AFHQ-CAT where only about 5000 training images are available.

	\item \textbf{Intermediate Generation Results.}
	\cref{fig:celebahq-longrun}, \cref{fig:cat-longrun}, and \cref{fig:church-longrun} show the intermediate generation results. It can be seen that the model is capable of transforming natural images into valid images of the target data distribution. In addition, when the number of PGD attack steps is too large, the generated samples become saturated, which 
	%In addition, some of these samples look similar to each other, suggesting that some  PGD attacks have converged to the same point. 
	suggests that the model, like many EBMs trained with short-run MCMC, do not have a valid steady-state that reflects the distribution of target data.
	\item \textbf{Compositional Visual Generation.}
	\cref{fig:compose}  shows that our model can be composed like regular EBMs~\cite{du2020compositional}.
	\item \textbf{Denosing and Inpainting.}
	\cref{fig:celebahq-denoising-inpainting} and \cref{fig:cat-denoising-inpainting} show uncurated denoising and inpainting results on CelebA-HQ 256 and AFHQ-CAT 256.
	
	\item \textbf{Image Translation.}
	\cref{fig:trans} shows uncurated image translation results on CelebA-HQ 256 and AFHQ-CAT 256.

\end{itemize}

\begin{figure}[h!]
	\centering
	\begin{minipage}[b]{0.46\textwidth}
		\centering
		\includegraphics[width=\linewidth]{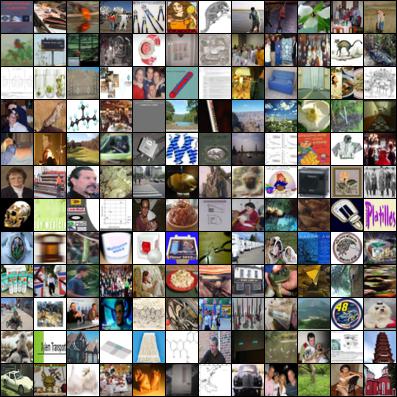}
		\\ Source images
	\end{minipage}
	\hfill
	\begin{minipage}[b]{0.46\textwidth}
		\centering
		\includegraphics[width=\linewidth]{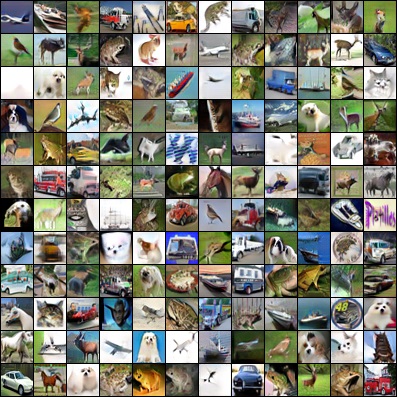}
		\\ Generated images
	\end{minipage}
	\caption{Uncurated CIFAR-10 generated samples.}
	\label{fig:cifar10-samples}
\end{figure}

\begin{figure}[h!]
	\centering
	\begin{minipage}[b]{0.46\textwidth}
		\centering
		\includegraphics[width=\linewidth]{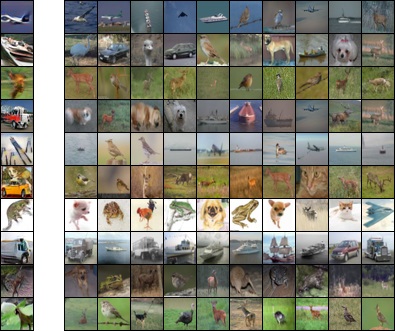}
		\\ Generated samples (left panel) and pixel space nearest neighbors (right panel).
	\end{minipage}
	\hfill
	\begin{minipage}[b]{.46\textwidth}
		\centering
		\includegraphics[width=\linewidth]{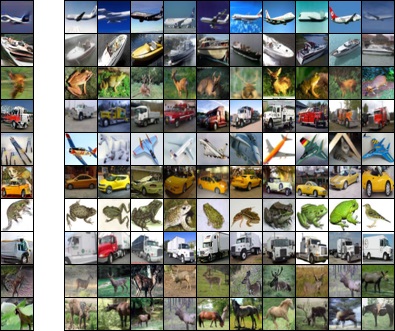}
		\\ Generated samples (left panel) and Inception feature space nearest neighbors (right panel).
	\end{minipage}%
	\caption{Nearest neighbors of generated samples on CIFAR-10.}
	\label{fig:cifar10-nn}
\end{figure}

\begin{figure}[h!]
	\centering
	\includegraphics[width=0.95\linewidth]{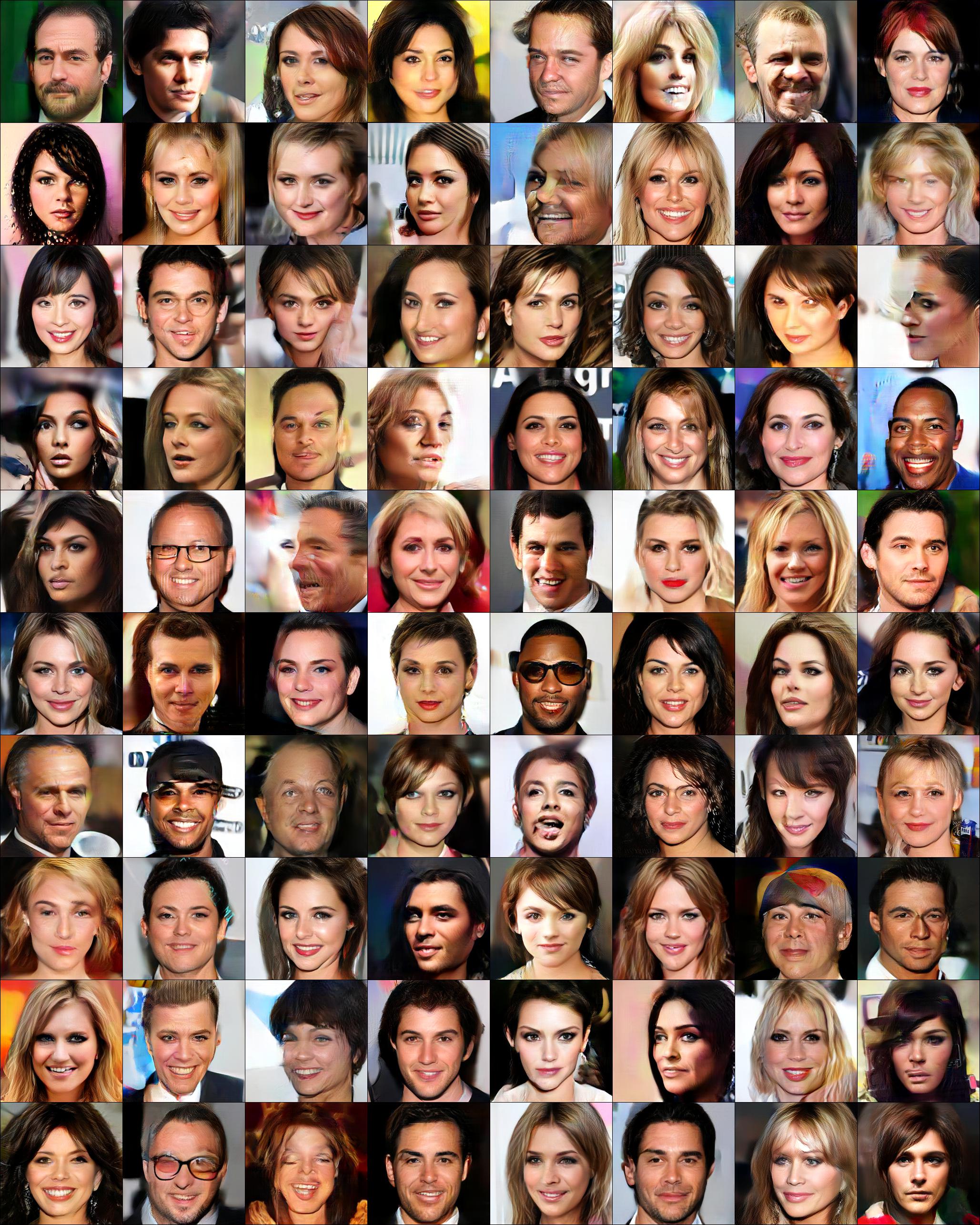}
	\caption{Uncurated generated samples on CelebAHQ-256. Source images are in \cref{fig:seed}.}
	\label{fig:celebahq-samples-uncurated}
\end{figure}

\begin{figure}[h!]
	\centering
	\includegraphics[width=0.95\linewidth]{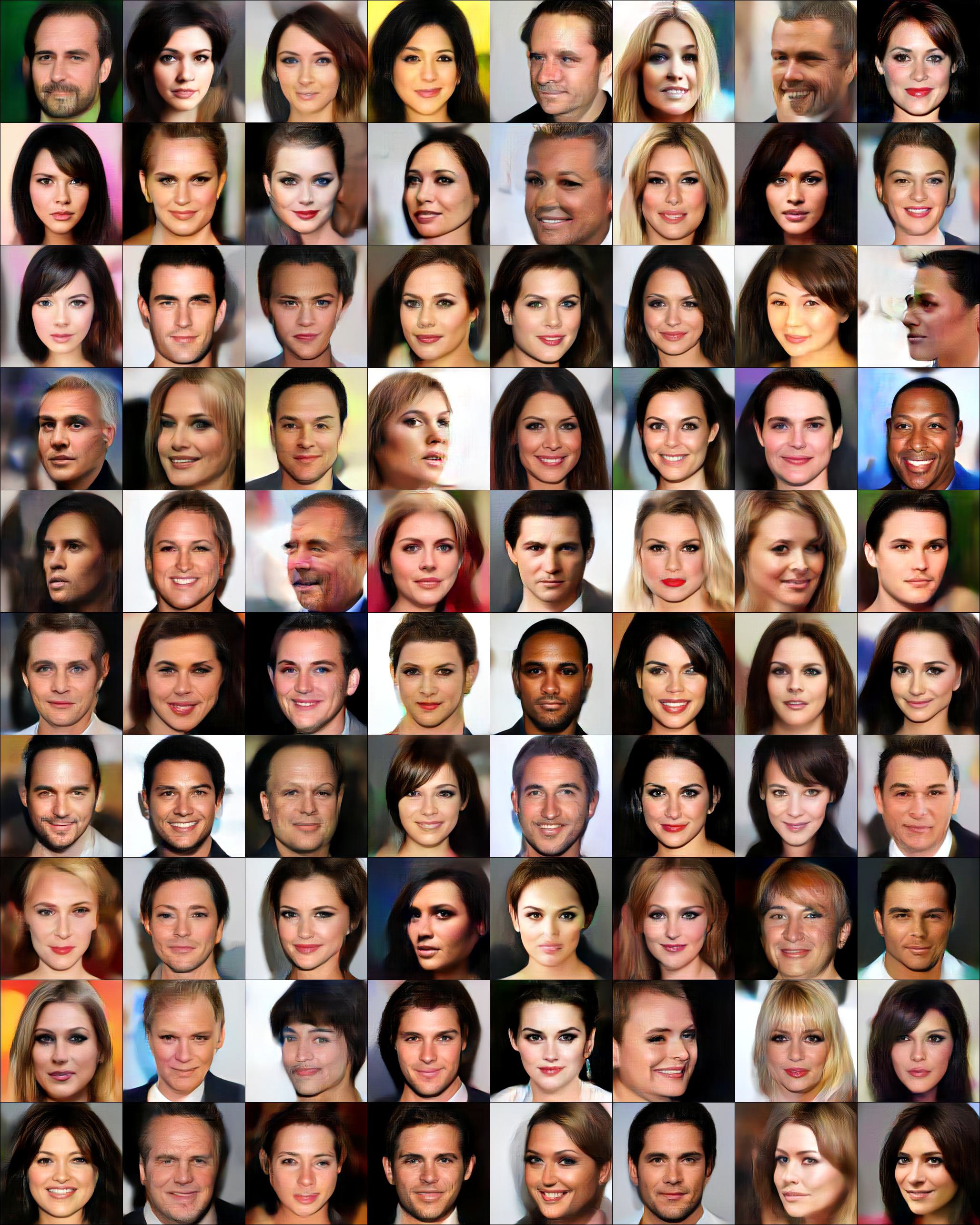}
	\caption{Uncurated generated samples on CelebAHQ-256. The source images used to generate these samples are obtained by applying Gaussian blur ($ \sigma=10 $) to the images in \cref{fig:seed}.}
	\label{fig:celebahq-samples-uncurated-blur}
\end{figure}

\begin{figure}[h!]
	\centering
	\includegraphics[width=0.95\linewidth]{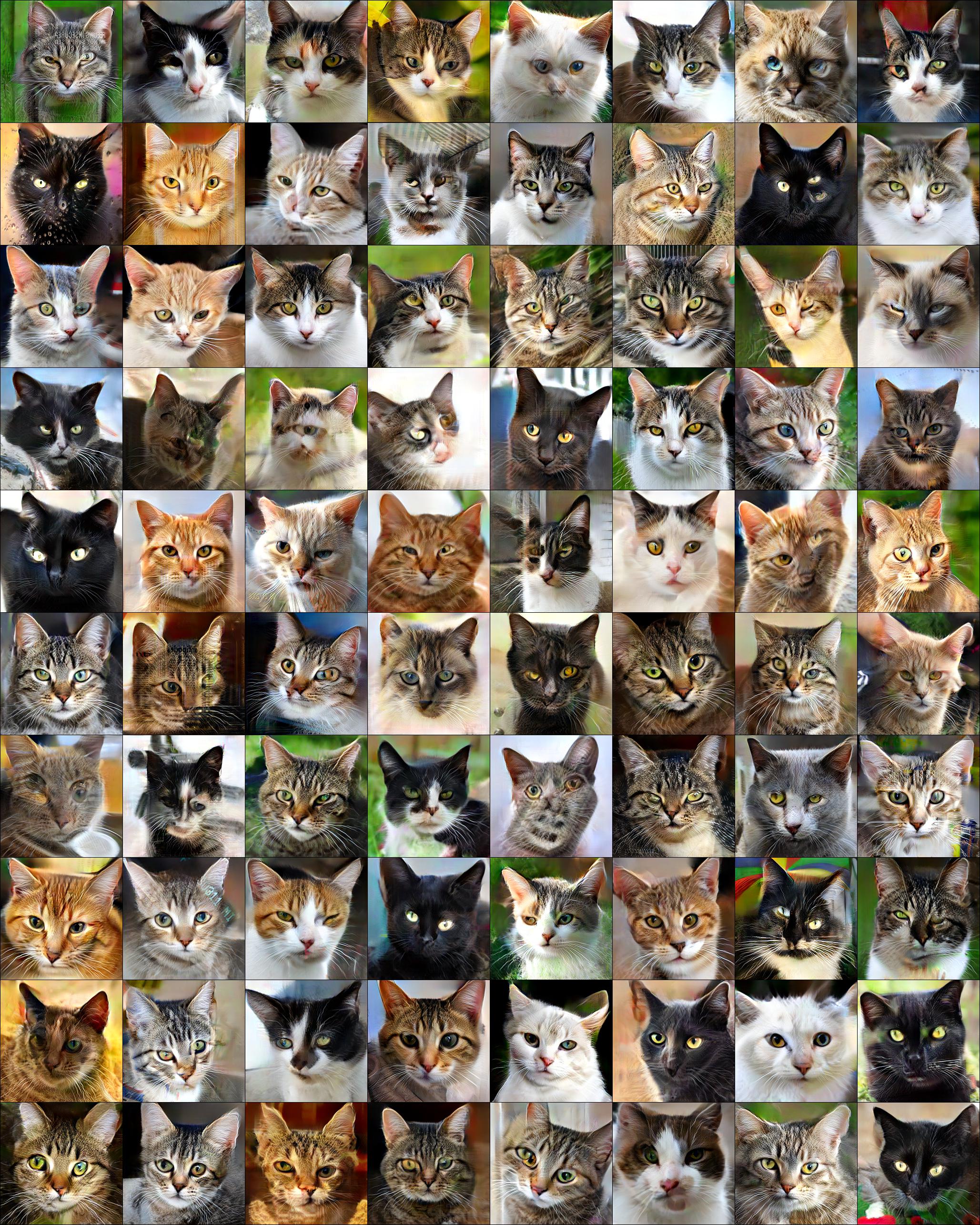}
	\caption{Uncurated generated samples on AFHQ-CAT 256. Source images are in \cref{fig:seed}.}
	\label{fig:cat-samples-uncurated}
\end{figure}

\begin{figure}[h!]
	\centering
	\includegraphics[width=0.95\linewidth]{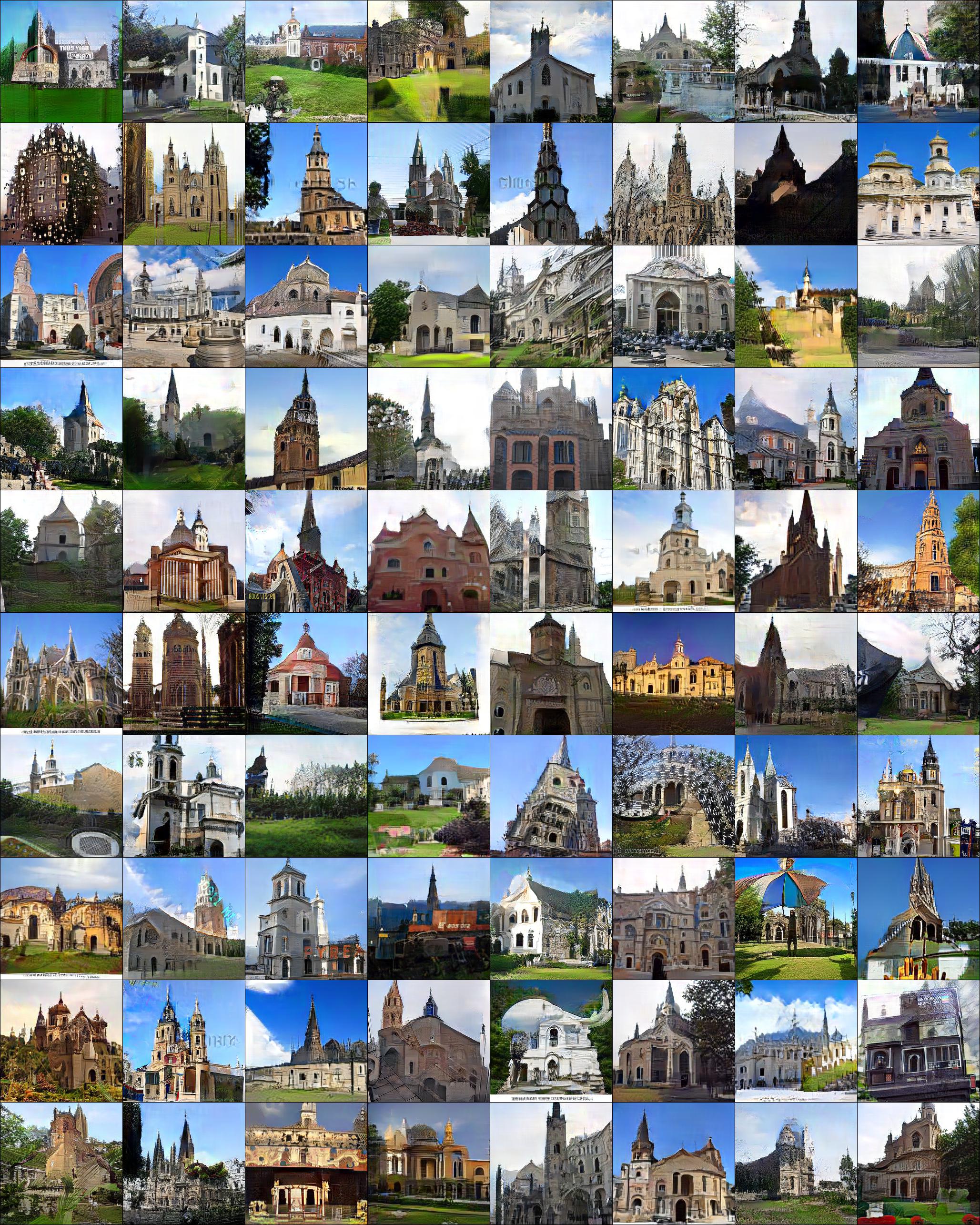}
	\caption{Uncurated generated samples on LSUN-Church 256. Source images are in \cref{fig:seed}.}
	\label{fig:church-samples-uncurated}
\end{figure}

\begin{figure}[h!]
	\centering
	\begin{minipage}[b]{0.46\textwidth}
		\centering
		\includegraphics[width=\linewidth]{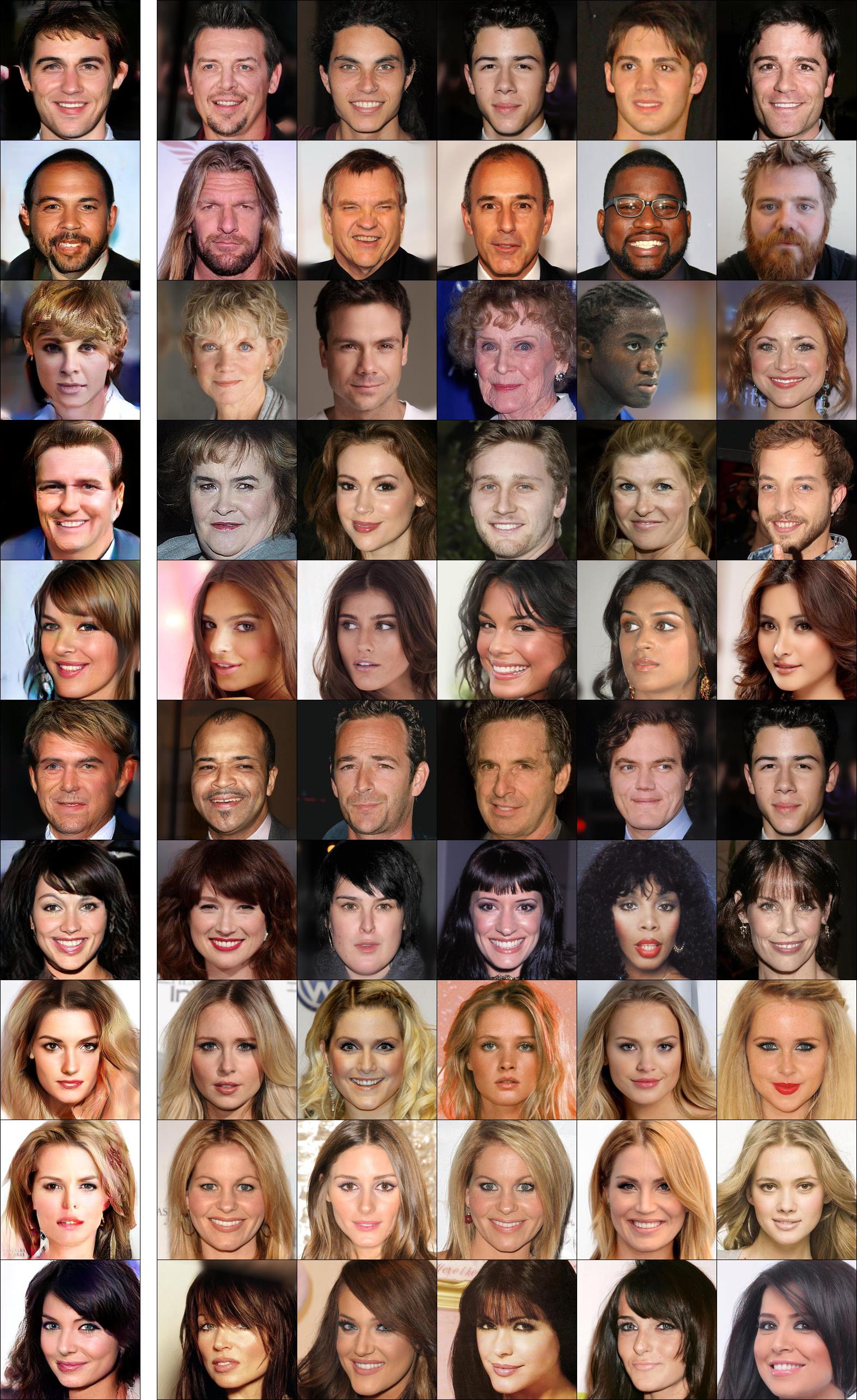}
		\\ Generated samples (left panel) and pixel space nearest neighbors (right panel
	\end{minipage}
	\hfill
	\begin{minipage}[b]{.46\textwidth}
		\centering
		\includegraphics[width=\linewidth]{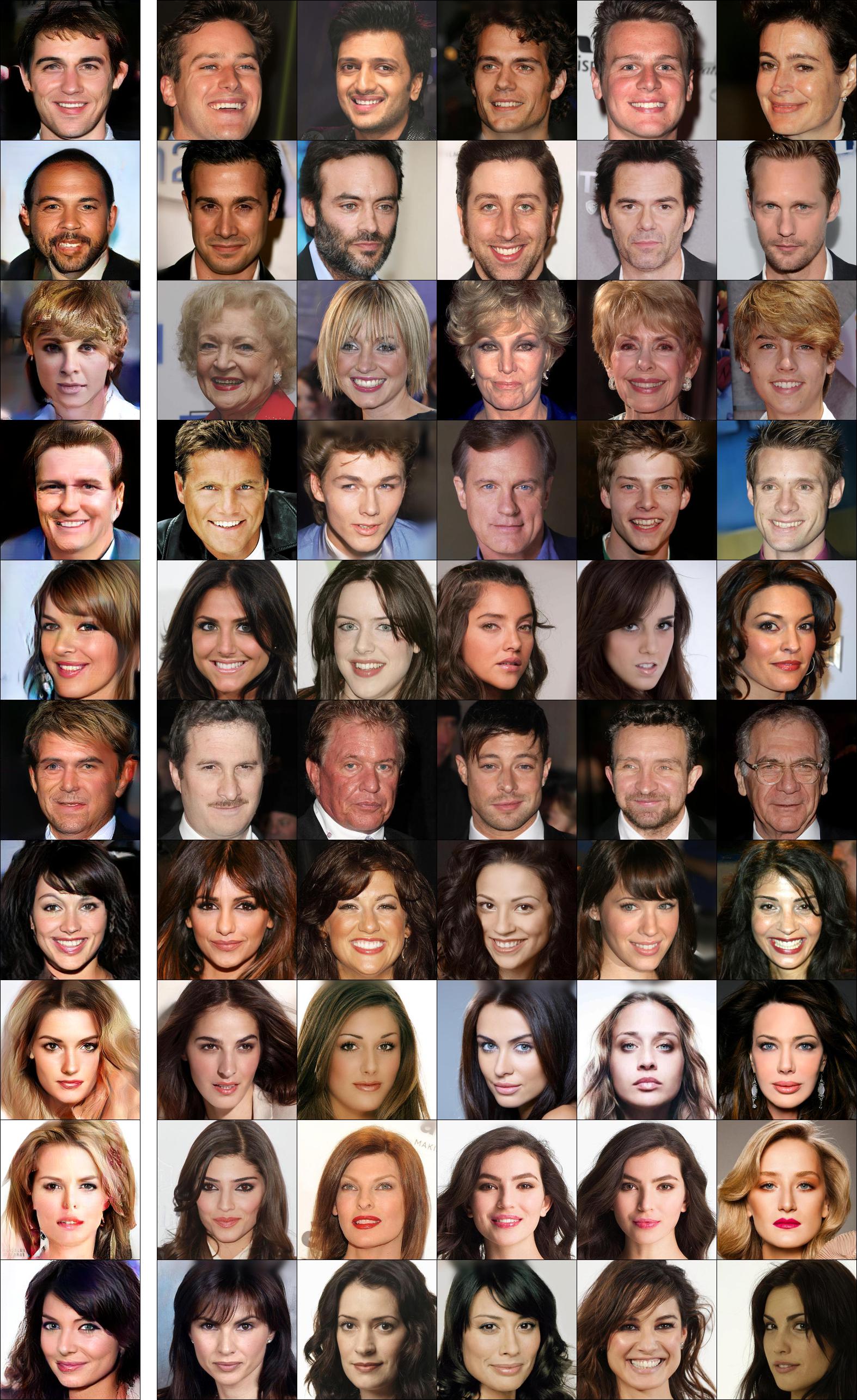}
		\\ Generated samples (left panel) and Inception feature space nearest neighbors (right panel).
	\end{minipage}%
	\caption{Nearest neighbors of generated samples on CelebA-HQ 256.}
	\label{fig:celebahq-nn}
\end{figure}

\begin{figure}[h!]
	\centering
	\begin{minipage}[b]{0.46\textwidth}
		\centering
		\includegraphics[width=\linewidth]{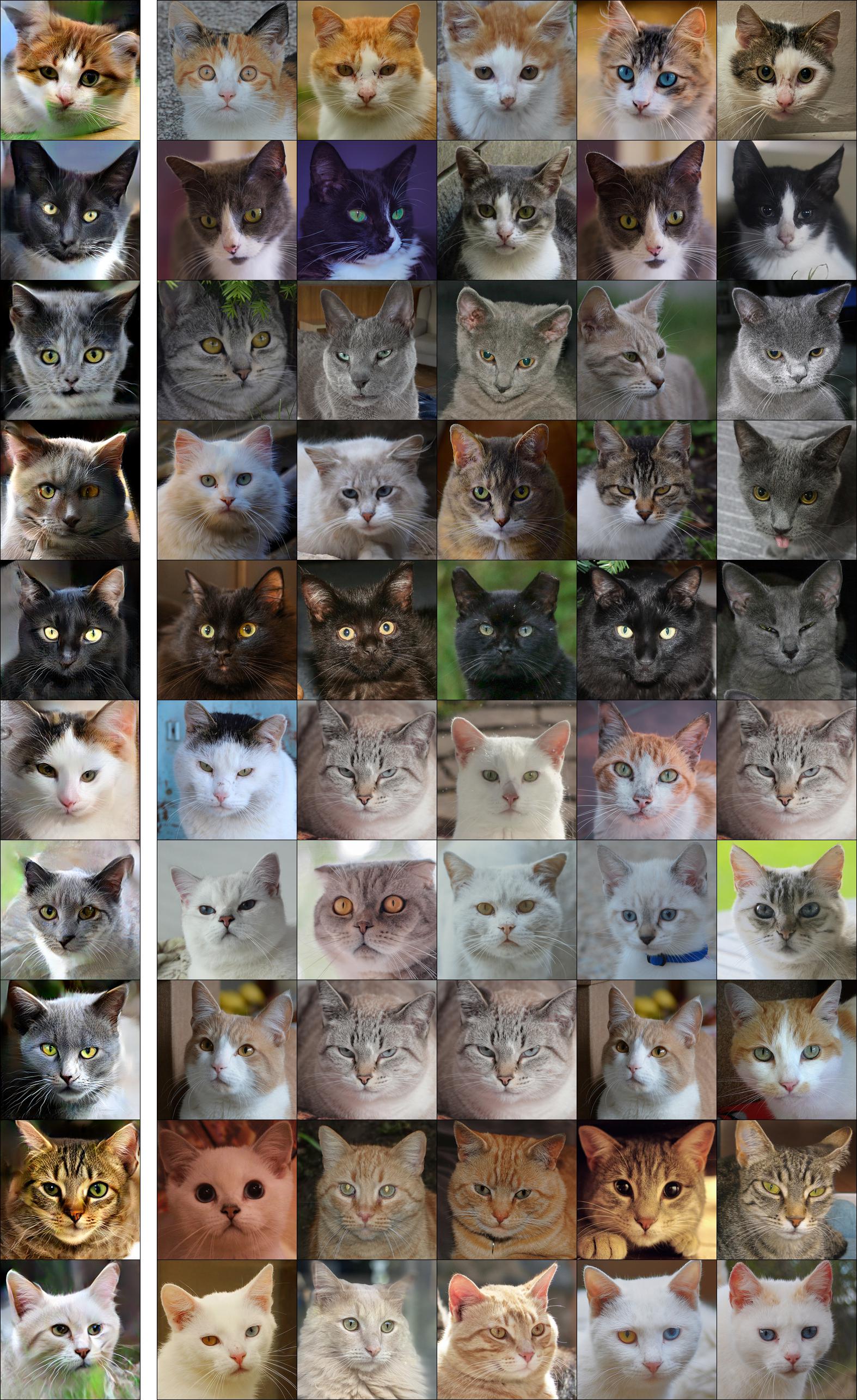}
		\\ Generated samples (left panel) and pixel space nearest neighbors (right panel
	\end{minipage}
	\hfill
	\begin{minipage}[b]{.46\textwidth}
		\centering
		\includegraphics[width=\linewidth]{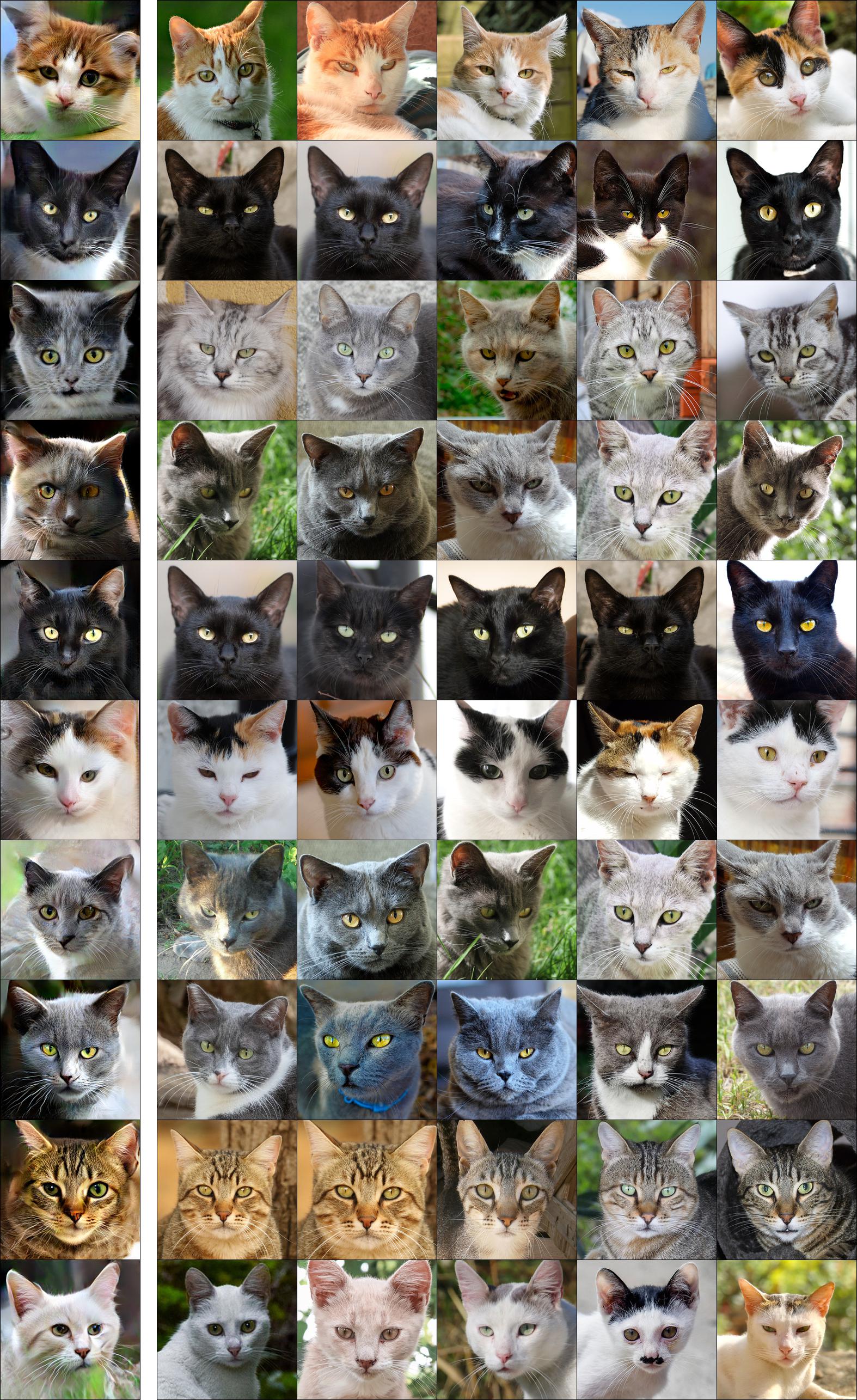}
		\\ Generated samples (left panel) and Inception feature space nearest neighbors (right panel).
	\end{minipage}%
	\caption{Nearest neighbors of generated samples on AFHQ-CAT 256.}
	\label{fig:cat-nn}
\end{figure}

\begin{figure}[h!]
	\centering
	\begin{minipage}[b]{0.46\textwidth}
		\centering
		\includegraphics[width=\linewidth]{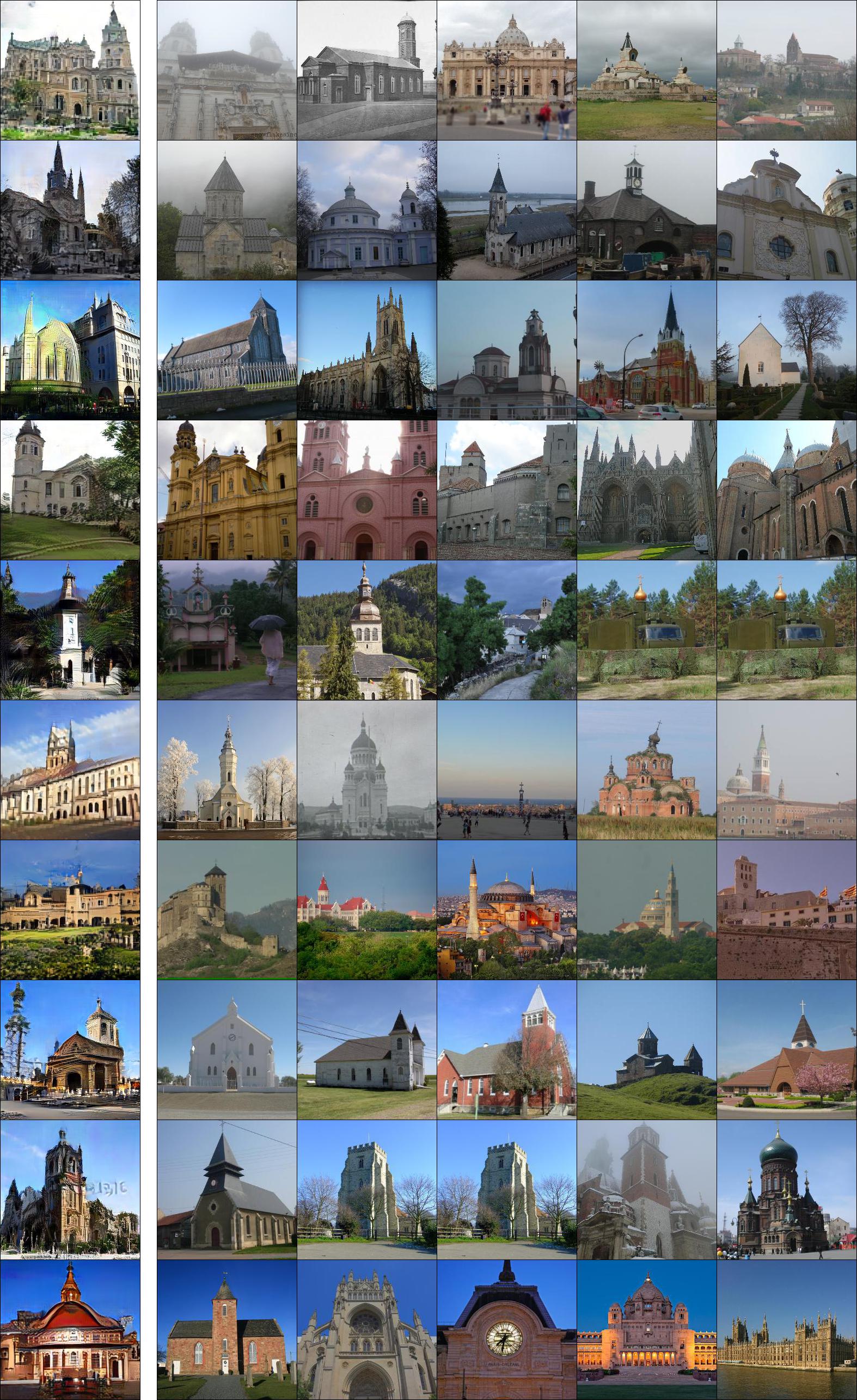}
		\\ Generated samples (left panel) and pixel space nearest neighbors (right panel
	\end{minipage}
	\hfill
	\begin{minipage}[b]{.46\textwidth}
		\centering
		\includegraphics[width=\linewidth]{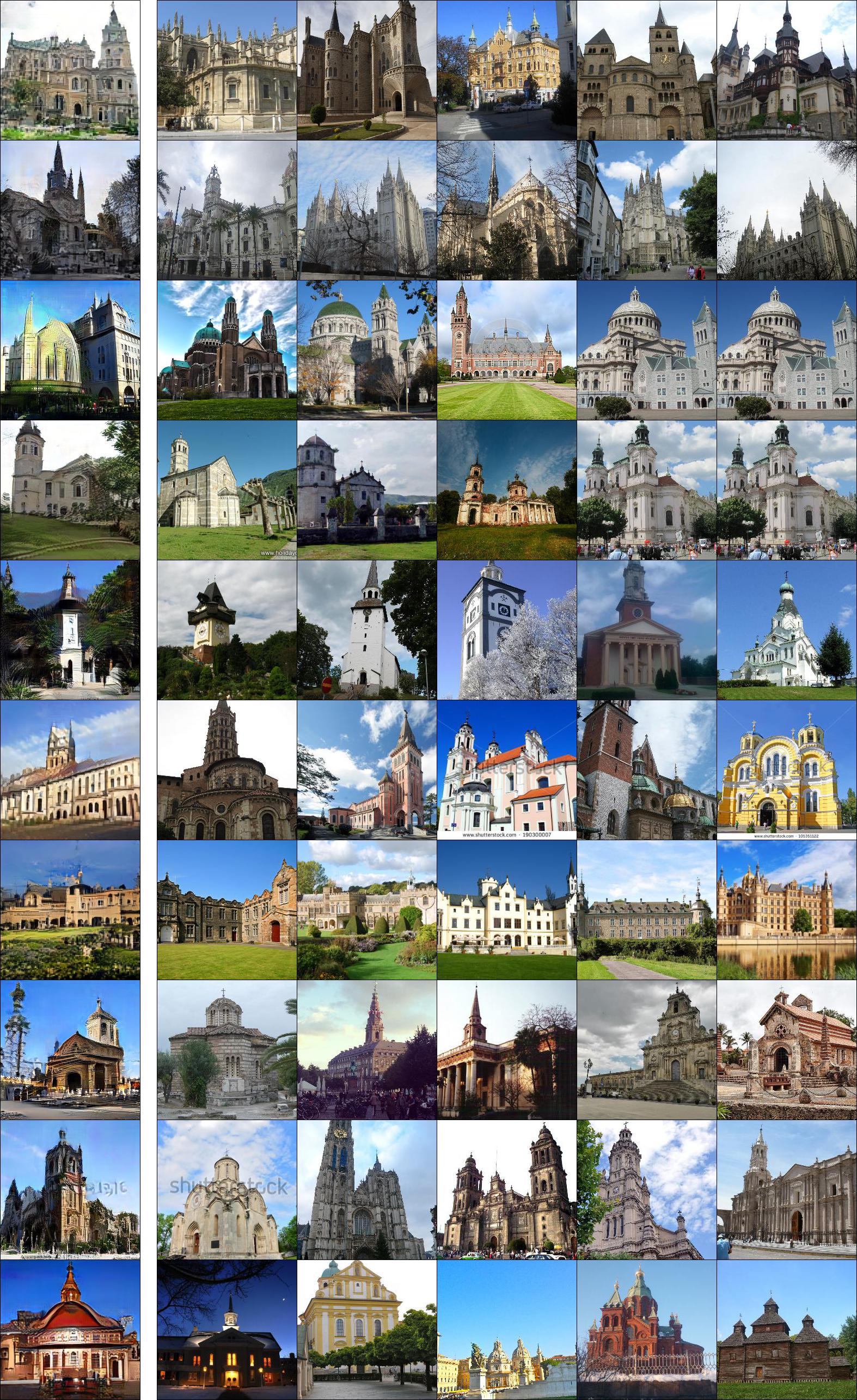}
		\\ Generated samples (left panel) and Inception feature space nearest neighbors (right panel).
	\end{minipage}%
	\caption{Nearest neighbors of generated samples on LSUN-Church 256.}
	\label{fig:church-nn}
\end{figure}

\begin{figure}[h!]
	\centering
	\includegraphics[width=\linewidth]{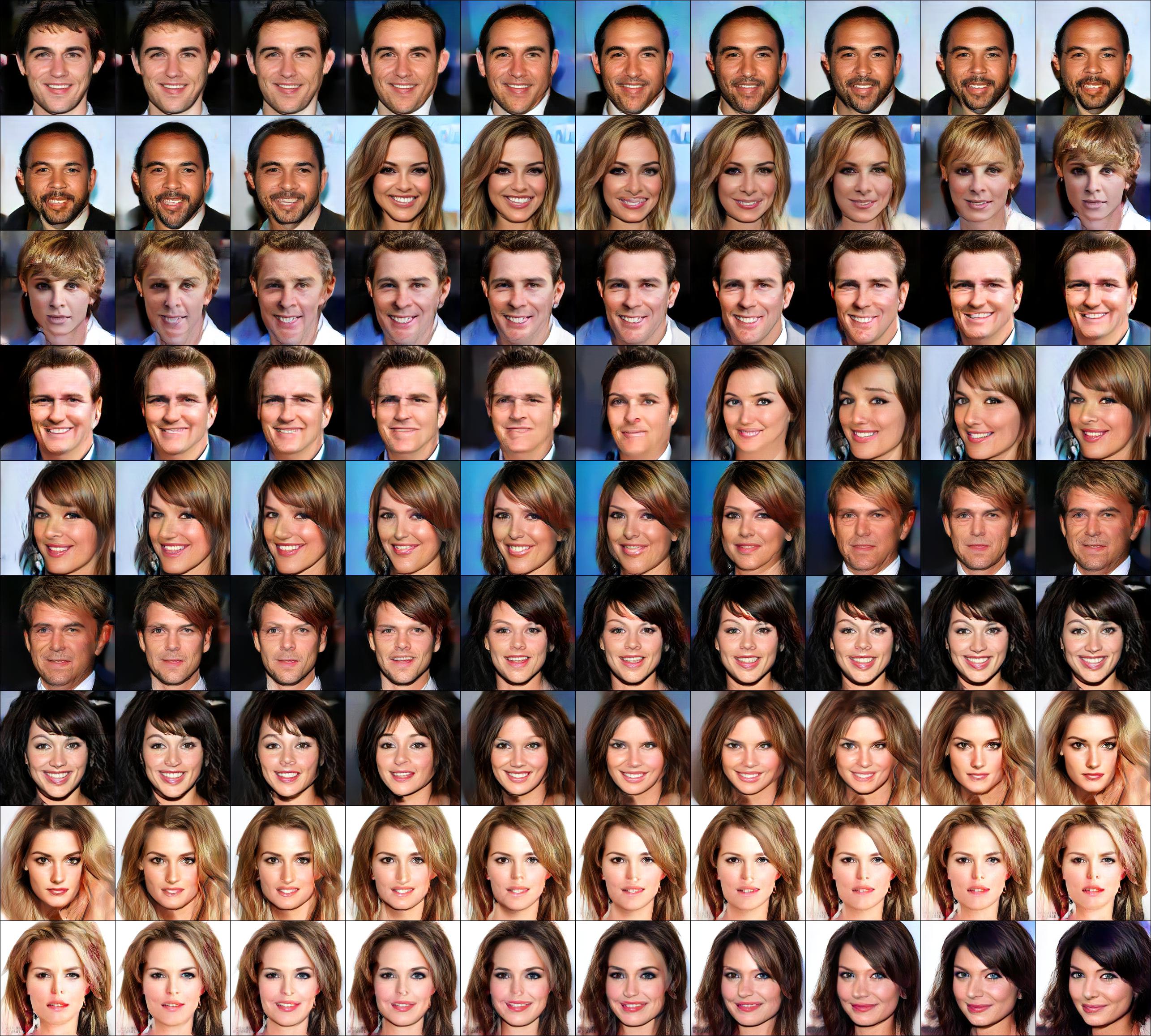}
	\caption{Interpolation results on CelebA-HQ 256. Intermediate images are generated by performing PGD attacks on linear interpolations between the source images used to generate the leftmost and rightmost samples.}
	\label{fig:celebahq-interpolation}
\end{figure}

\begin{figure}[h!]
	\centering
	\includegraphics[width=\linewidth]{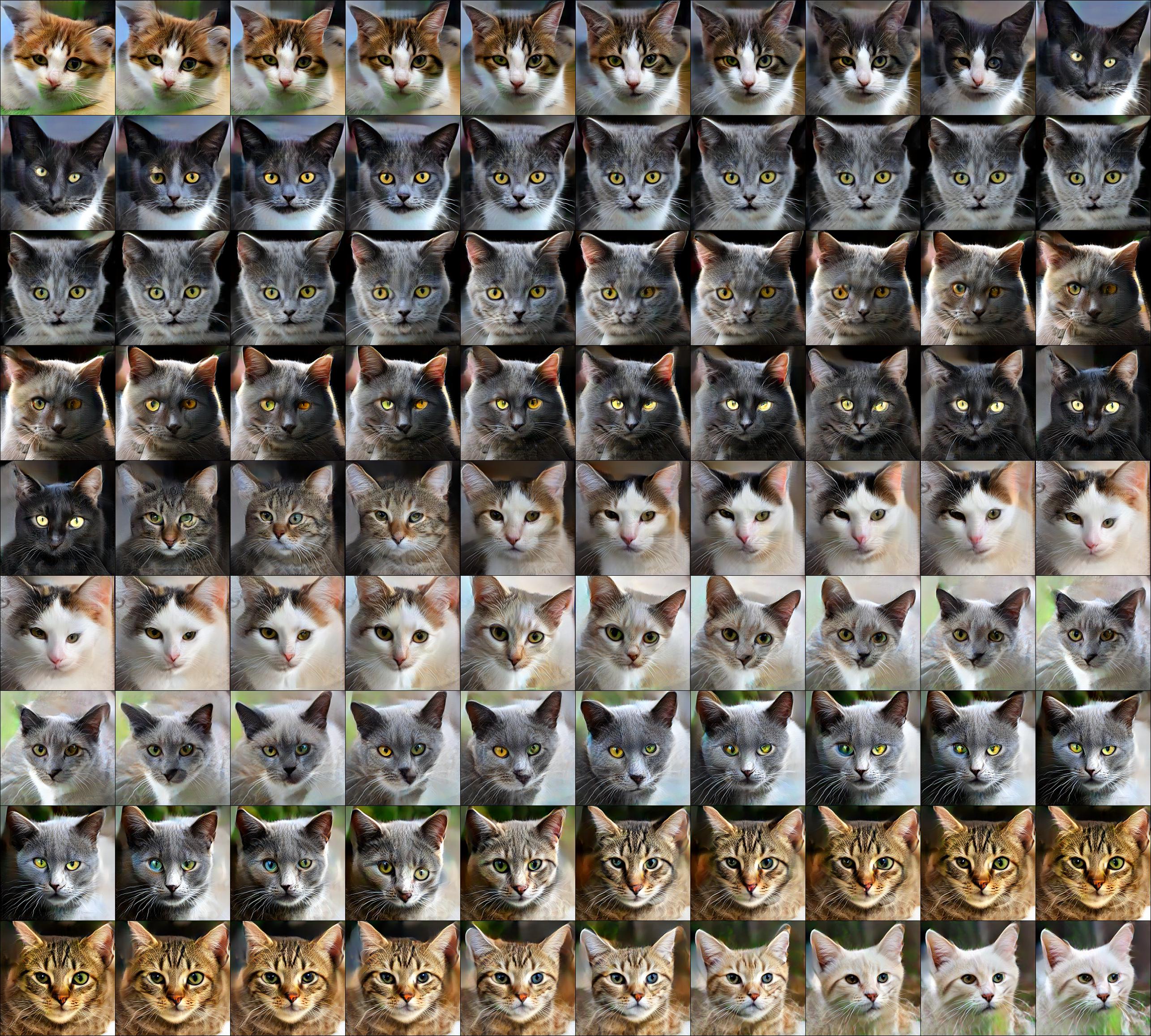}
	\caption{Interpolation results on  AFHQ-CAT 256. Intermediate images are generated by performing PGD attacks on linear interpolations between the source images used to generate the leftmost and rightmost samples.}
	\label{fig:cat-interpolation}
\end{figure}

\begin{figure}[h!]
	\centering
	\includegraphics[width=\linewidth]{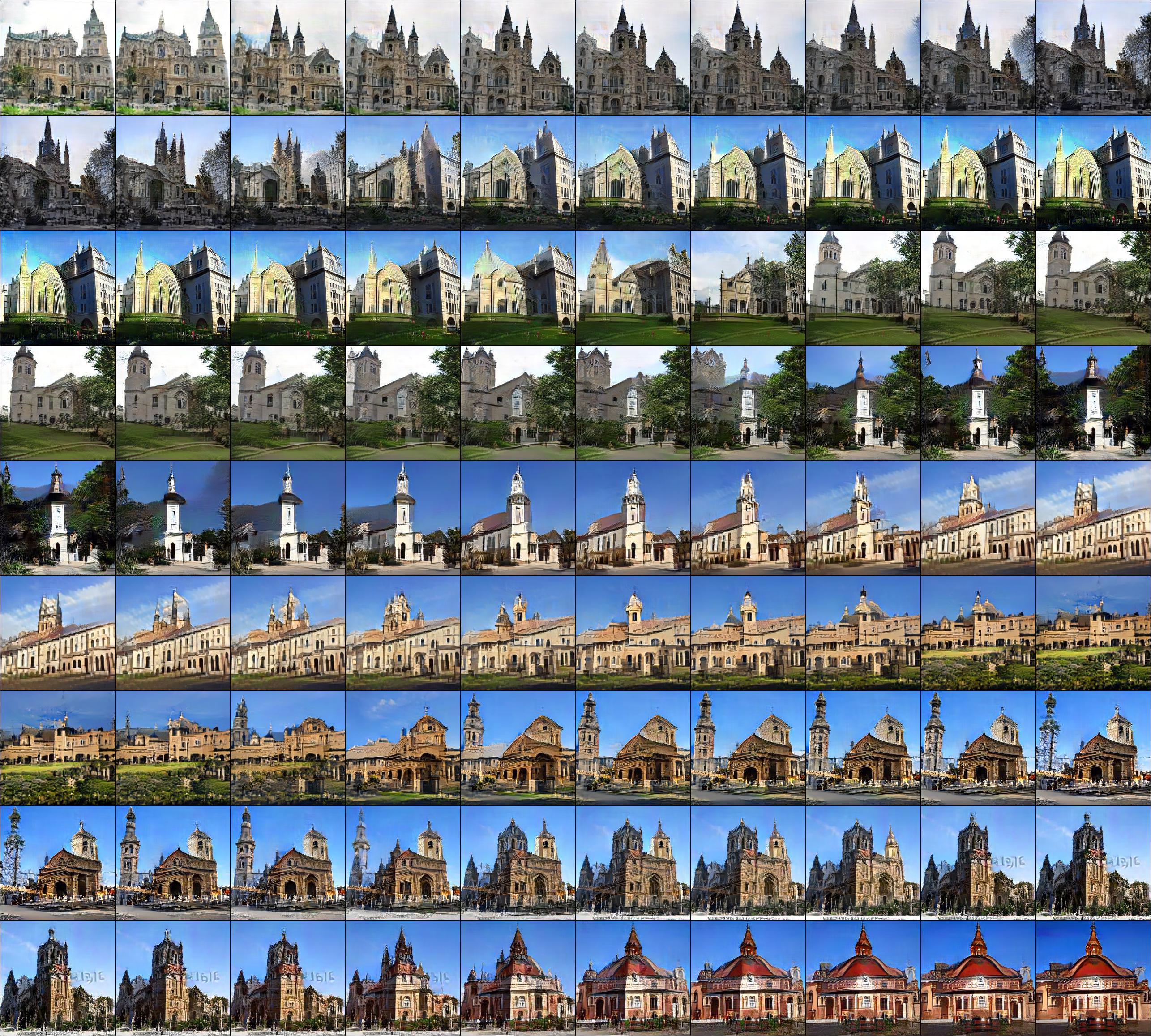}
	\caption{Interpolation results on  LSUN-Church 256. Intermediate images are generated by performing PGD attacks on linear interpolations between the source images used to generate the leftmost and rightmost samples.}
	\label{fig:church-interpolation}
\end{figure}

%\begin{figure}[h!]
%	\centering
%	\begin{minipage}[b]{0.46\textwidth}
	%		\centering
	%		\includegraphics[width=\linewidth]{figs/appendix/celebahq_longrun.jpg}
	%		\\CelebA-HQ 256 samples generated by long-run PGD attacks.  The attack steps for column 1-7 are [0,   6,  13,  19,  26,  33, 200].
	%	\end{minipage}
%	\hfill
%	\begin{minipage}[b]{.46\textwidth}
	%		\centering
	%		\includegraphics[width=\linewidth]{figs/appendix/cat_longrun.jpg}
	%		\\ AFHQ-CAT 256 samples generated by long-run PGD attacks. The attack steps for column 1-7 are [0,   6,  13,  19,  26,  33, 200].
	%	\end{minipage}%
%	\begin{minipage}[b]{.46\textwidth}
	%	\centering
	%	\includegraphics[width=\linewidth]{figs/appendix/church_longrun.jpg}
	%	\\ AFHQ-CAT 256 samples generated by long-run PGD attacks. The attack steps for column 1-7 are [0,   6,  13,  19,  26,  33, 200].
	%\end{minipage}%
	%	\caption{Longrun result.}
	%	\label{fig:longrun}
	%\end{figure}
	
	\begin{figure}[h!]
		\centering
		\includegraphics[width=\linewidth]{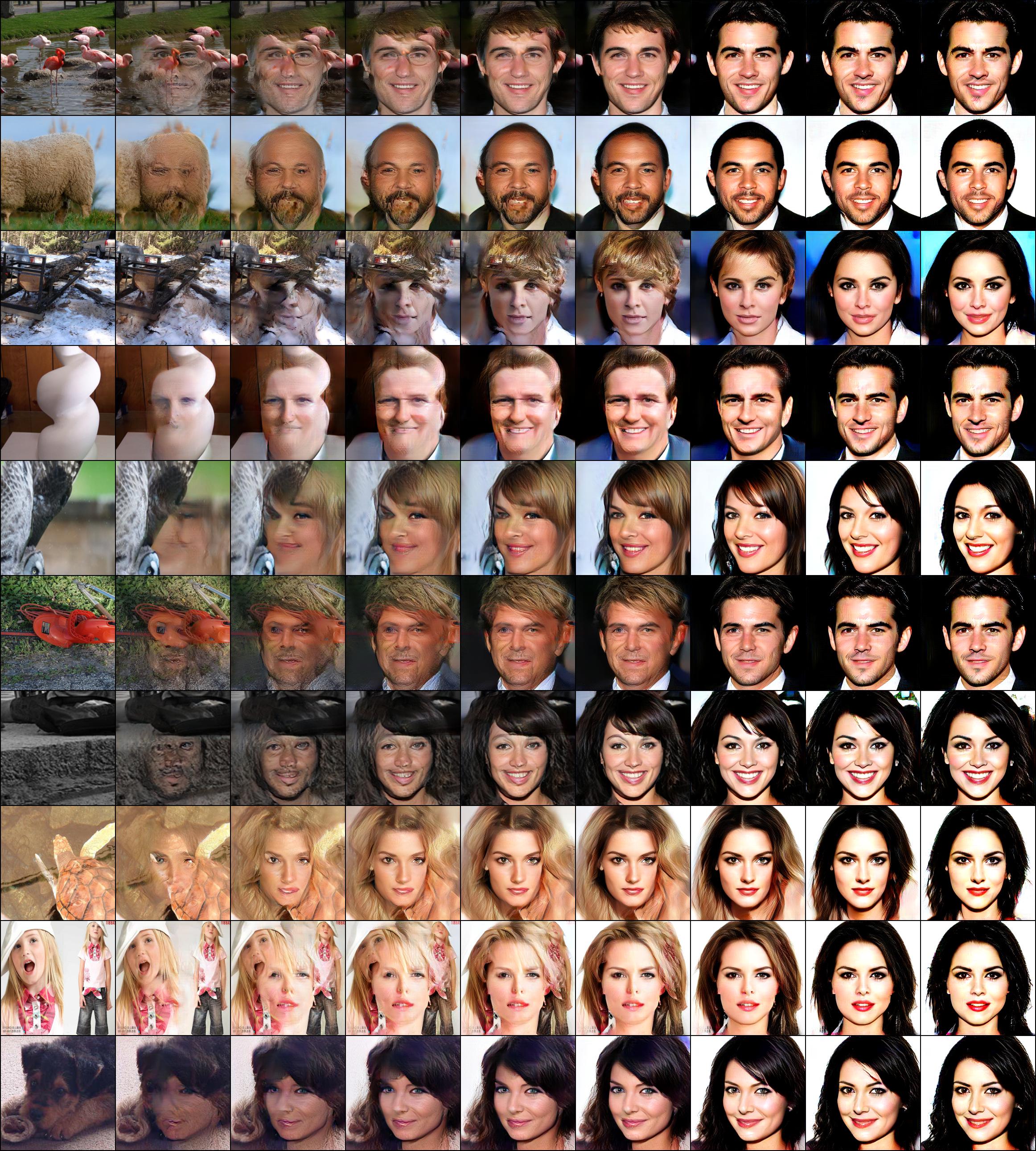}
		\caption{CelebA-HQ 256 intermediate generation results.  The PGD attack steps for column 1-9 are [ 0,  4,  8, 12, 16, 20, 30, 40, 50] (steps 20 has the best FID score).}
		\label{fig:celebahq-longrun}
	\end{figure}
	
	\begin{figure}[h!]
		\centering
		\includegraphics[width=\linewidth]{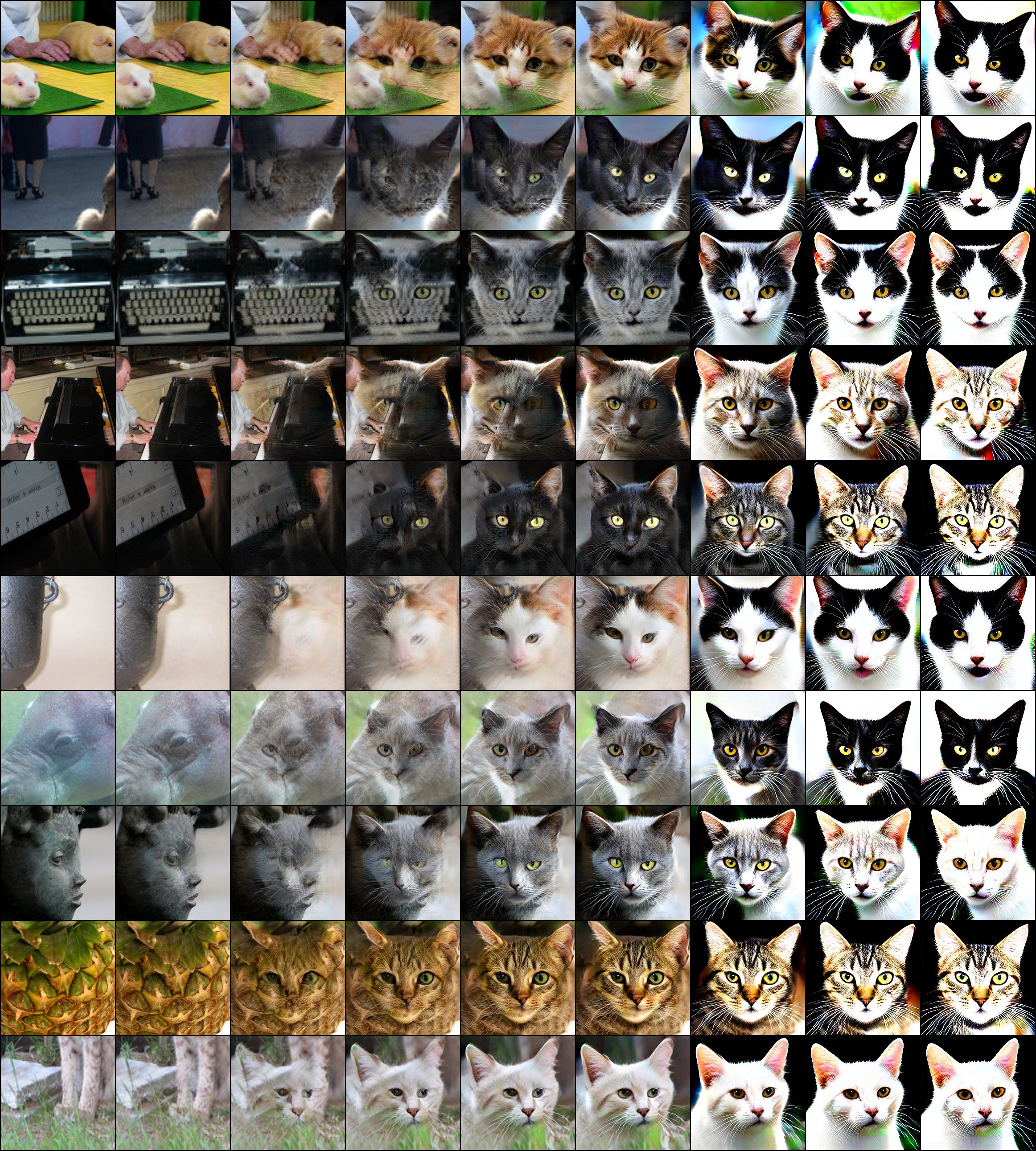}
		\caption{AFHQ-CAT 256 intermediate generation results.  The PGD attack steps for column 1-9 are [0,  2,  5,  8, 11, 14, 30, 50, 100]  (steps 14 has the best FID score).}
		\label{fig:cat-longrun}
	\end{figure}
	
	\begin{figure}[h!]
		\centering
		\includegraphics[width=\linewidth]{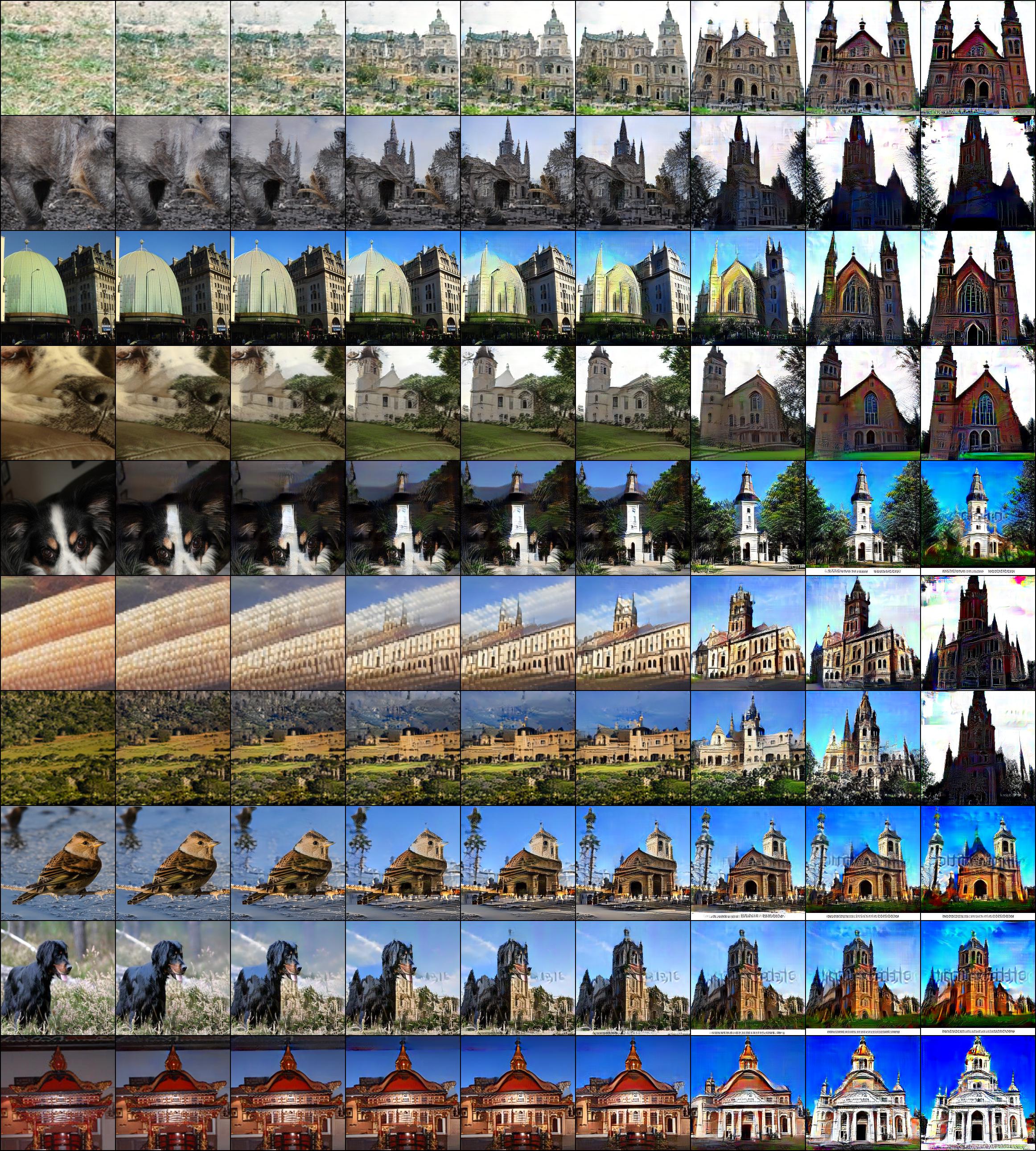}
		\caption{LSUN-Church 256 intermediate generation results.  The PGD attack steps for column 1-9 are [ 0,  3,  6, 10, 13, 17, 30, 50, 100]  (steps 17 has the best FID score).}
		\label{fig:church-longrun}
	\end{figure}
	
	\begin{figure}[h!]
		\centering
		\includegraphics[width=0.8\linewidth]{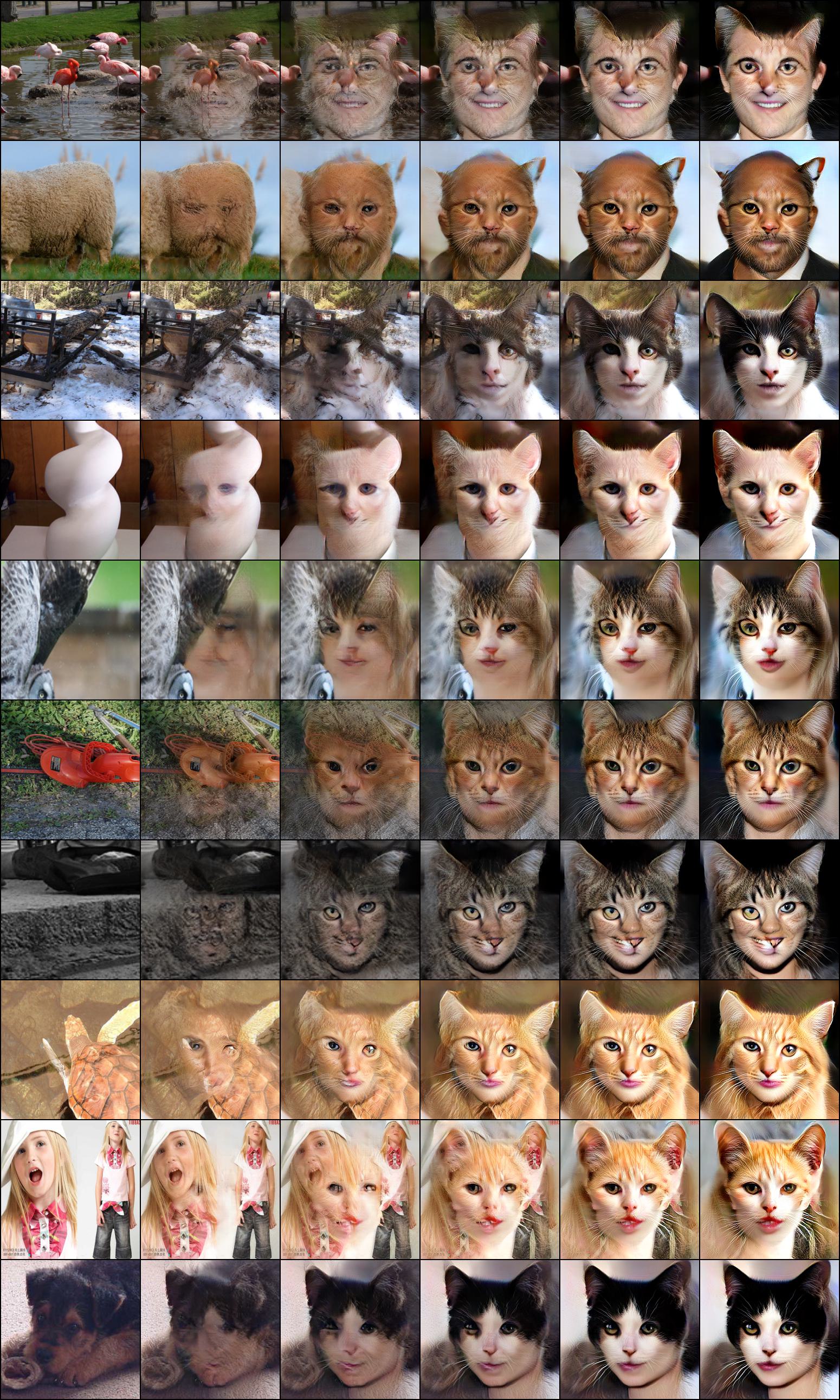}
		\caption{Concept conjunction~\cite{du2020compositional} using the CelebA-HQ model and AFHQ-CAT model. The generated samples have both human face features and cat face features.}
		\label{fig:compose}
	\end{figure}

	%
	%\begin{figure}[h!]
	%	\centering
	%	\begin{minipage}[b]{.51\textwidth}
		%		\centering
		%		\includegraphics[width=\linewidth]{figs/appendix/celebahq_longrun.jpg}
		%		\\CelebA-HQ 256 samples generated by long-run PGD attacks.  The attack steps for column 1-7 are [0,   6,  13,  19,  26,  33, 200].
		%	\strut\end{minipage}%
	%%\hfill\allowbreak%
	%\centering
	%	\begin{minipage}[b]{0.51\textwidth}
		%		\includegraphics[width=\linewidth]{figs/appendix/cat_longrun.jpg}
		%				\\ AFHQ-CAT 256 samples generated by long-run PGD attacks. The attack steps for column 1-7 are [0,   6,  13,  19,  26,  33, 200].
		%%		\includegraphics[width=0.46\linewidth]{figs/appendix/church_longrun.jpg}
		%%		\\ AFHQ-CAT 256 samples generated by long-run PGD attacks. The attack steps for column 1-7 are [0,   6,  13,  19,  26,  33, 200].
		%	\strut\end{minipage}%
	%	\caption{Longrun result.}
	%	\label{fig:longrun}
	%\end{figure}
	
	\begin{figure}[h!]
		\centering
		\includegraphics[width=\linewidth]{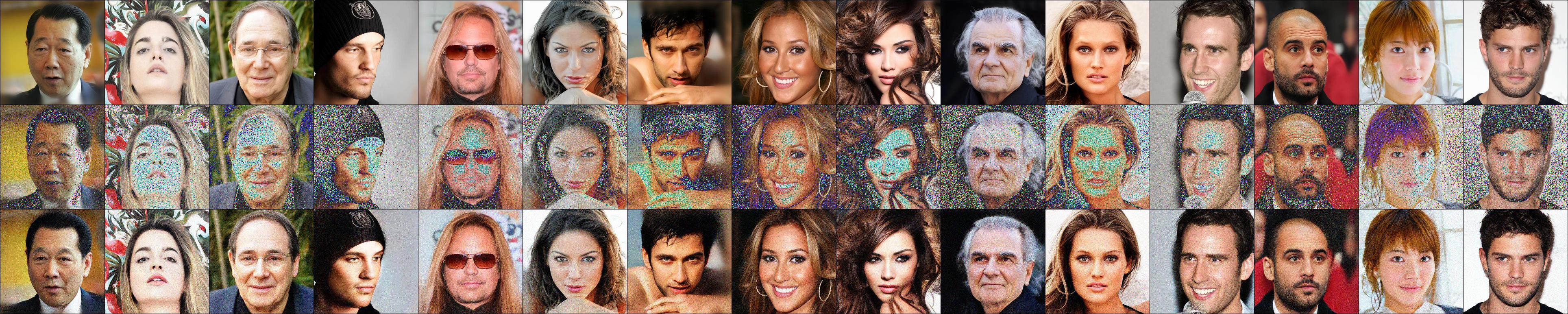}
		\\{Original images (1st row), images with additive Gaussian noise of standard deviation of 0.1 (2nd row), and recovered images (last row).}	\includegraphics[width=\linewidth]{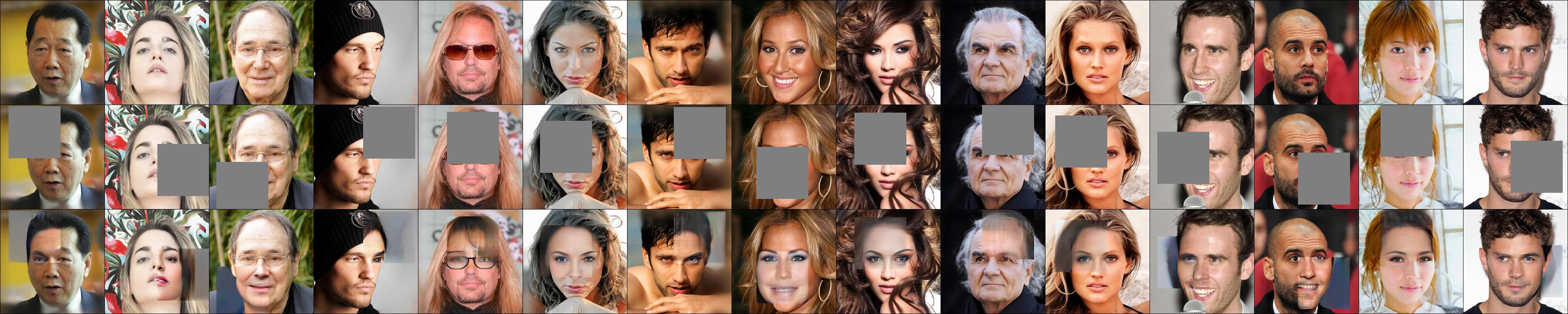}
		\\{Original image (1st row), occluded images (2nd row), and recovered images (last row).}
		\caption{Uncurated denoising and inpainting results on CelebA-HQ 256.}
		\label{fig:celebahq-denoising-inpainting}
	\end{figure}
	
	%
	%\begin{figure}[h!]
	%		\centering
	%		\includegraphics[width=\linewidth]{figs/appendix/ffhq_denoising.png}
	%		\\{First row: original images. Second row: images with additive Gaussian noise of standard deviation of 0.1. Last row: recovered images.}
	%		\centering
	%		\includegraphics[width=\linewidth]{figs/appendix/ffhq_inpainting.jpg}
	%		\\{First row: occluded images. Second row: recovered images.}
	%	\caption{Uncurated denoising and inpainting results on FFHQ dataset. The source images are from FFHQ dataset~\cite{karras2019style}.}
	%	\label{fig:ffhq-denoising-inpainting}
	%\end{figure}

	\begin{figure}[h!]
		\centering
		\includegraphics[width=\linewidth]{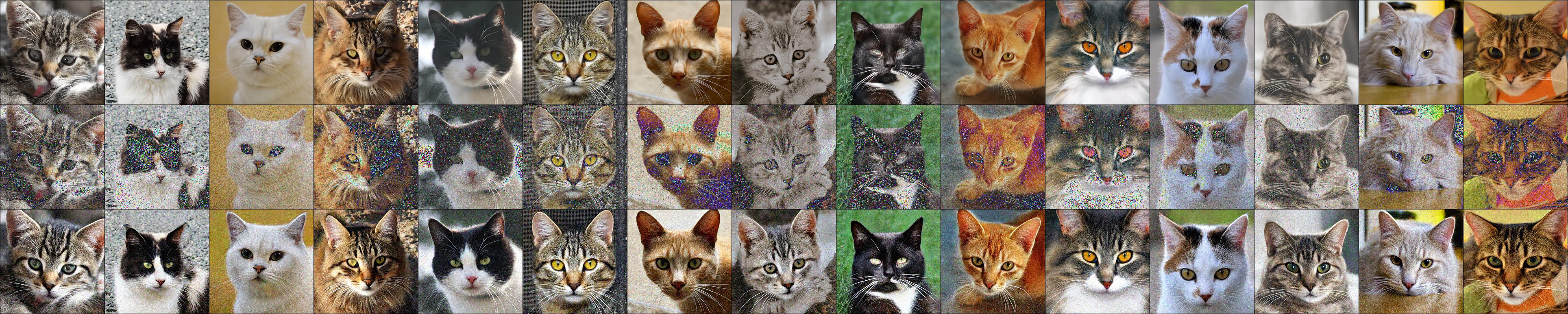}
		\\{Original images (1st row), images with additive Gaussian noise of standard deviation of 0.1 (2nd row), and recovered images (last row).}
		\includegraphics[width=\linewidth]{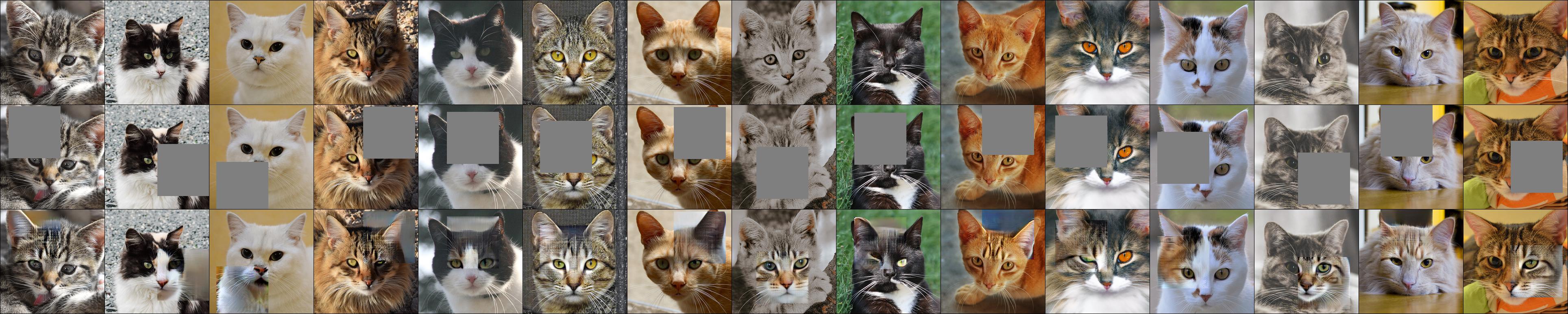}
		\\{Original image (1st row), occluded images (2nd row), and recovered images (last row).}
		\caption{Uncurated denoising and inpainting results on AFHQ-CAT 256.}
		\label{fig:cat-denoising-inpainting}
	\end{figure}

	\begin{figure}[h!]
		\centering
		\begin{minipage}[b]{0.46\textwidth}
			\centering
			\includegraphics[width=\linewidth]{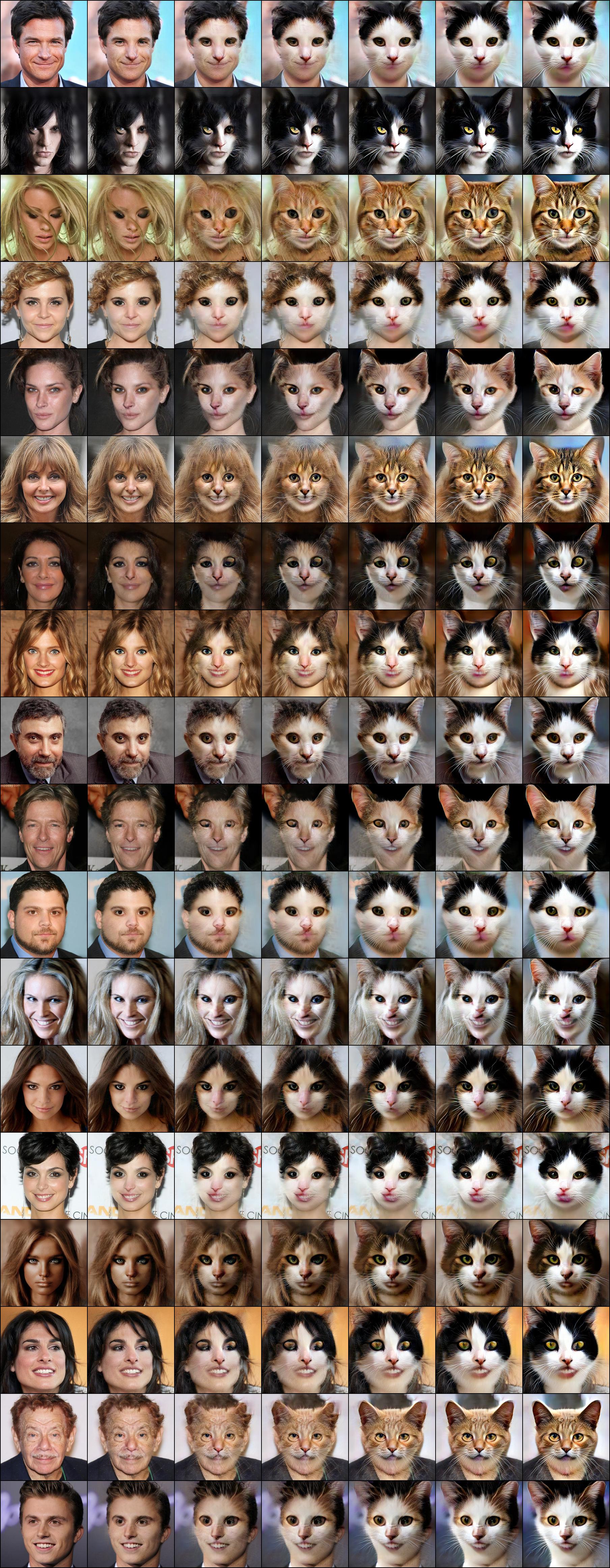}
			\\ Uncurated image translation results on CelebA-HQ 256.
		\end{minipage}
		\hfill
		\begin{minipage}[b]{.46\textwidth}
			\centering
			\includegraphics[width=\linewidth]{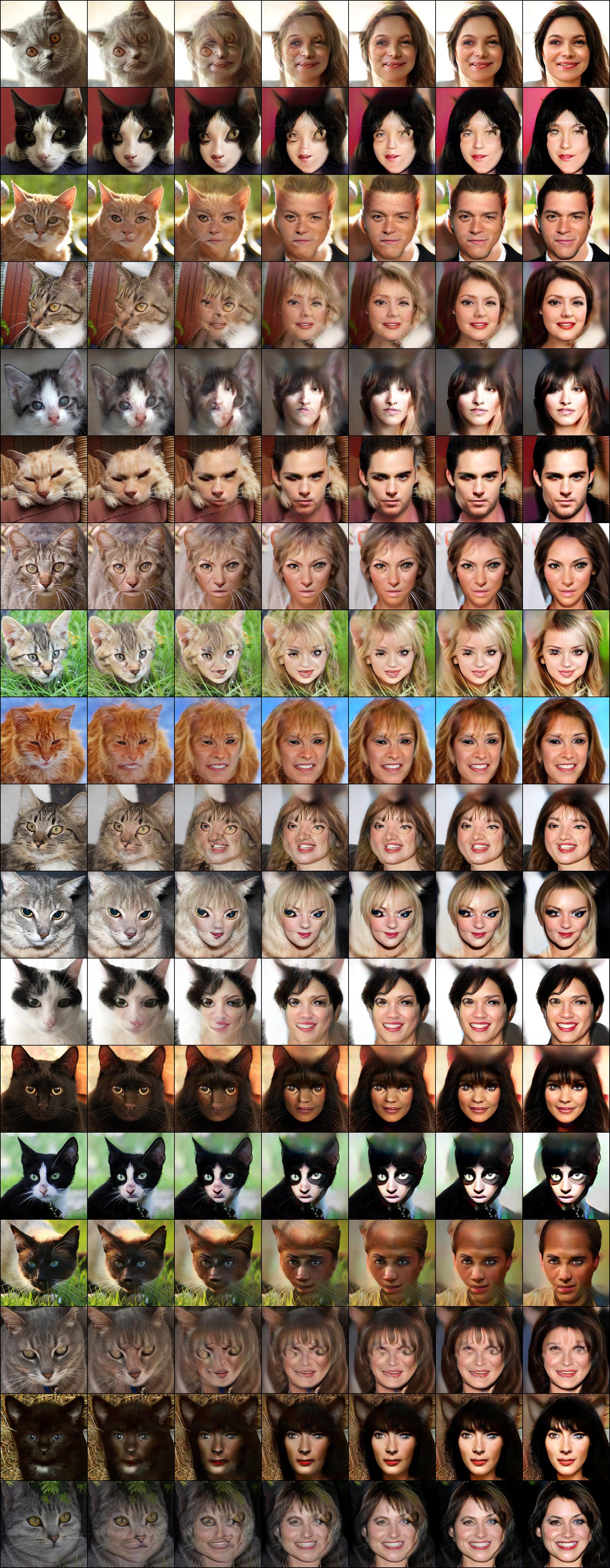}
			\\ Uncurated image translation results on  AFHQ-CAT 256.
		\end{minipage}%
		\caption{Uncurated image translation samples.}
		\label{fig:trans}
	\end{figure}

	\begin{figure}[h!]
		\centering
		\includegraphics[width=0.95\linewidth]{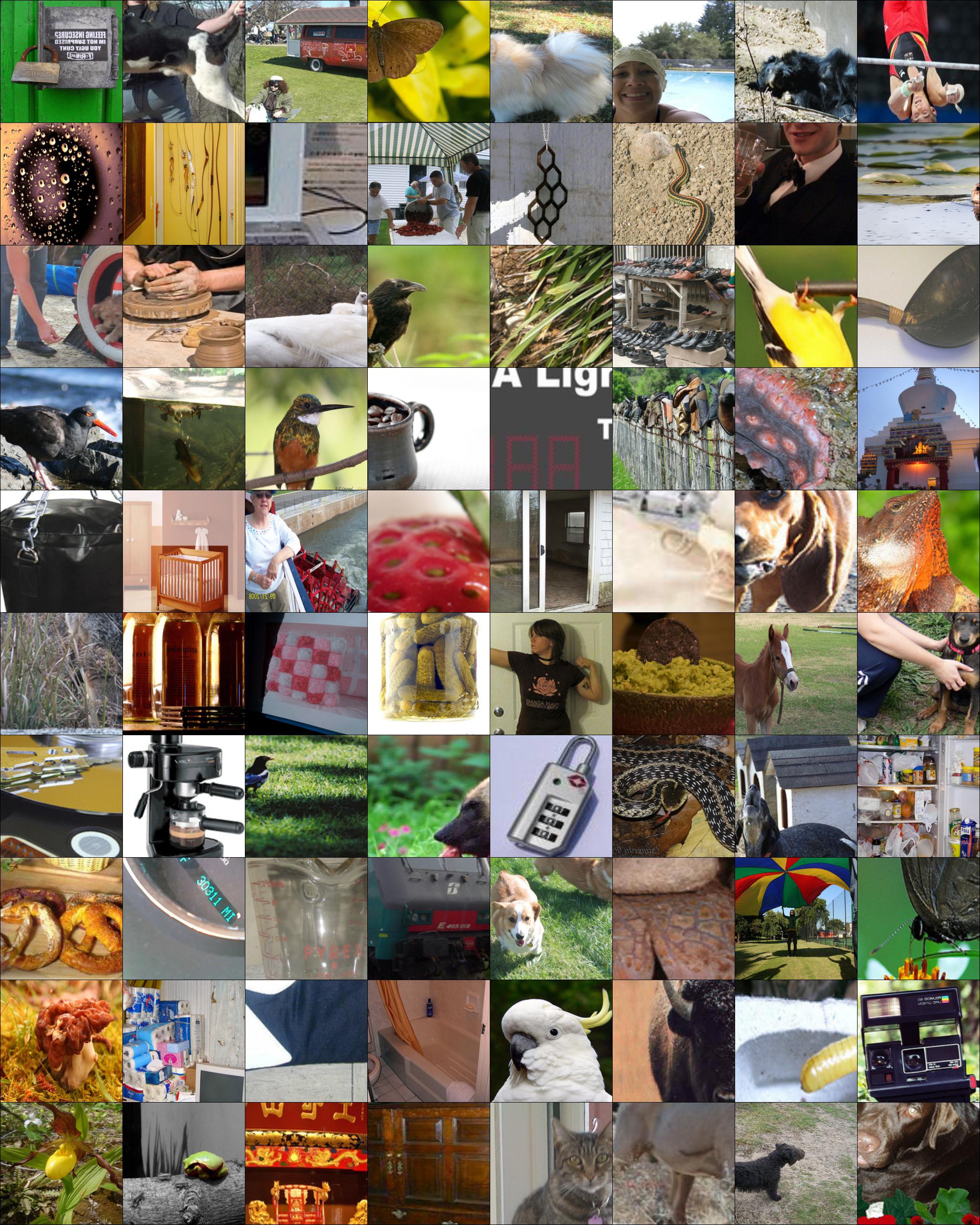}
		\caption{Seed images used to generate samples in \cref{fig:celebahq-samples-uncurated}, \cref{fig:celebahq-samples-uncurated-blur}, \cref{fig:cat-samples-uncurated}, and \cref{fig:church-samples-uncurated}.}
		\label{fig:seed}
	\end{figure}

\fi

\end{document}